\documentclass[runningheads]{llncs}
\usepackage{graphicx}
\usepackage{comment}
\usepackage{amsmath,amssymb} 
\usepackage{color}

\usepackage{booktabs}
\usepackage{subcaption}
\usepackage{xcolor}
\usepackage{multirow}
\usepackage{tabularx}
\usepackage{colortbl}
\usepackage{tikz}
\usepackage{pgfplots}

\newcommand{\PNP}{P\emph{n}P}
\newcommand{\eg}{\emph{e.g.}}
\newcommand{\ie}{\emph{i.e.}}

\newcommand{\gt}{\text{gt}}
\newcommand{\pred}{\text{pred}}

\DeclareMathOperator*{\argmin}{\arg\!\min}

\begin{document}
\pagestyle{headings}
\mainmatter
\def\ECCVSubNumber{2501}  

\title{Geometric Correspondence Fields:\\Learned Differentiable Rendering for\\3D Pose Refinement in the Wild} 

\titlerunning{Geometric Correspondence Fields}
%
\author{Alexander Grabner\inst{1,2} \and Yaming Wang\inst{2} \and Peizhao Zhang\inst{2} \and Peihong Guo\inst{2} \and\\ Tong Xiao\inst{2} \and Peter Vajda\inst{2} \and Peter M. Roth\inst{1} \and Vincent Lepetit\inst{1}}
\authorrunning{A. Grabner et al.}
%
\institute{\textsuperscript{1} Graz University of Technology, Austria \qquad \textsuperscript{2} Facebook Inc., USA\\ \email{\{alexander.grabner, pmroth, lepetit\}@icg.tugraz.at}\\ \email{\{wym, stzpz, phg, xiaot, vajdap\}@fb.com}}
\maketitle

\begin{abstract}
We present a novel 3D pose refinement approach based on differentiable rendering for objects of arbitrary categories in the wild. In contrast to previous methods, we make two main contributions: First, instead of comparing real-world images and synthetic renderings in the RGB or mask space, we compare them in a feature space optimized for 3D pose refinement. Second, we introduce a novel differentiable renderer that learns to approximate the rasterization backward pass from data instead of relying on a hand-crafted algorithm. For this purpose, we predict deep cross-domain correspondences between RGB images and 3D model renderings in the form of what we call geometric correspondence fields. These correspondence fields serve as pixel-level gradients which are analytically propagated backward through the rendering pipeline to perform a gradient-based optimization directly on the 3D pose. In this way, we precisely align 3D models to objects in RGB images which results in significantly improved 3D pose estimates. We evaluate our approach on the challenging Pix3D dataset and achieve up to 55\% relative improvement compared to state-of-the-art refinement methods in multiple metrics.
\end{abstract}

\section{Introduction}
\label{sec:intro}

\begin{figure}[t]
\centering
\begin{subfigure}{0.22\linewidth}
	\begin{center}
		\includegraphics[width=\linewidth]{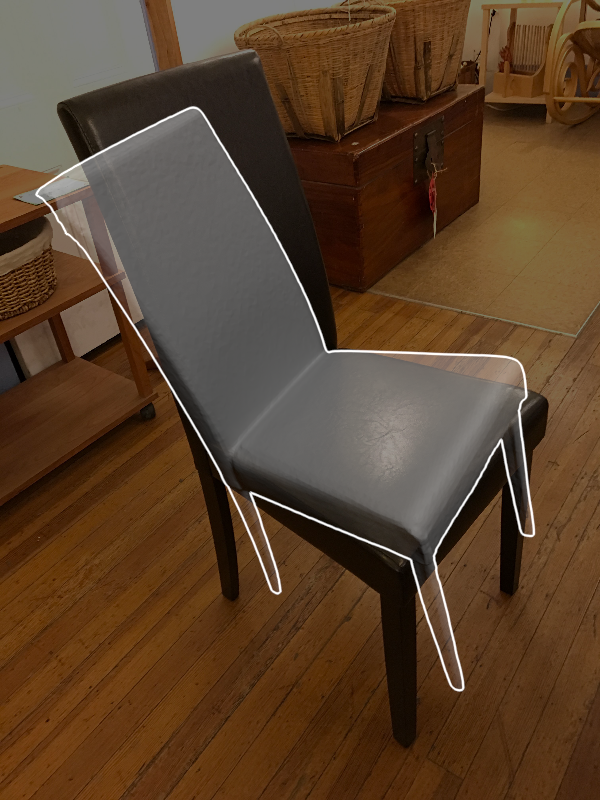}
	\end{center}
\end{subfigure}\qquad\begin{subfigure}{0.22\linewidth}
	\begin{center}
		\includegraphics[width=\linewidth]{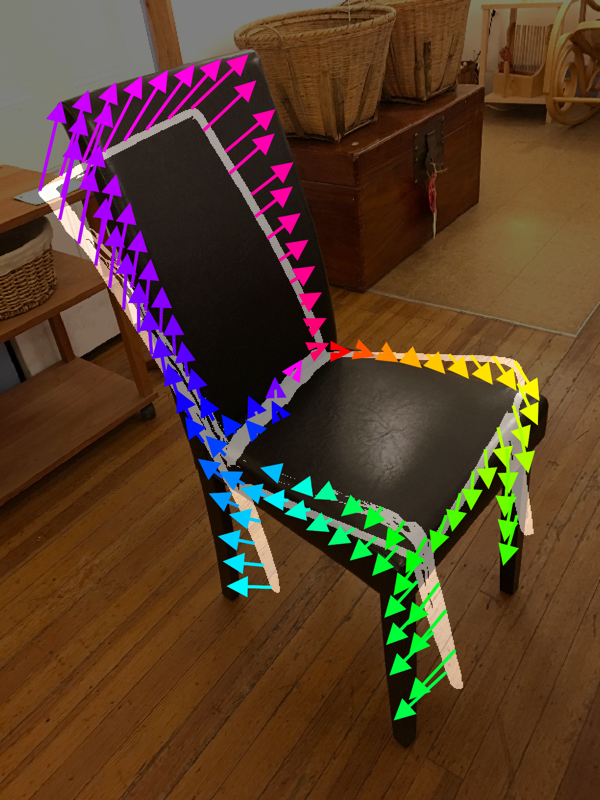}
	\end{center}
\end{subfigure}\qquad\begin{subfigure}{0.22\linewidth}
	\begin{center}
		\includegraphics[width=\linewidth]{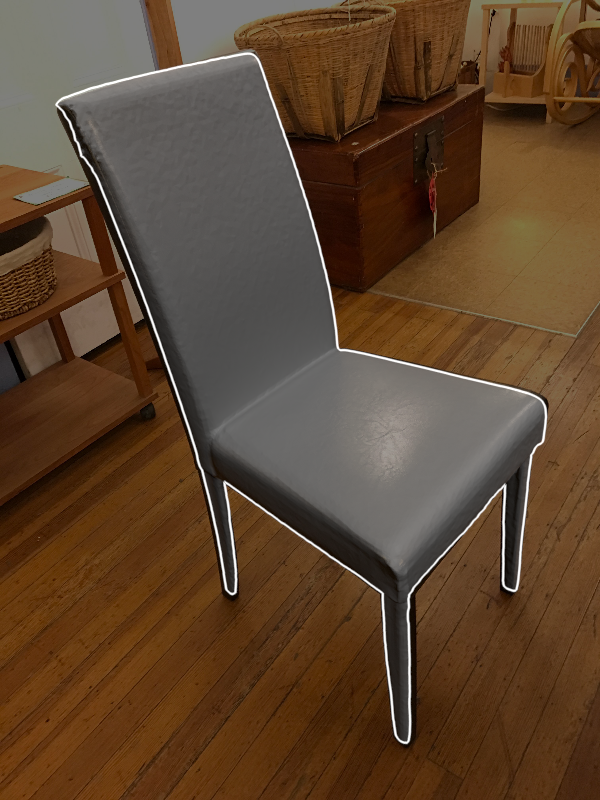}
	\end{center}
\end{subfigure}
	\caption{Given an initial 3D pose predicted by a feed-forward network (\textit{left}), we predict deep cross-domain correspondences between real-world RGB images and synthetic 3D model renderings in the form of geometric correspondence fields (\textit{middle}) that enable us to refine the 3D pose in a differentiable rendering framework (\textit{right}).}
	\label{fig:teaser}
\end{figure}

Recently, there have been significant advances in single image 3D object pose estimation thanks to deep learning~\cite{Grabner2018a,Mousavian20163d,Tulsiani2015viewpoints}. However, the accuracy achieved by today's feed-forward networks is not sufficient for many applications like augmented reality or robotics~\cite{Grabner2019a,Wang2018fine}. As shown in Figure~\ref{fig:teaser}, feed-forward networks can robustly estimate the coarse high-level 3D rotation and 3D translation of objects in the wild (\textit{left}) but fail to predict fine-grained 3D poses (\textit{right})~\cite{Xiao2019pose}.

To improve the accuracy of predicted 3D poses, refinement methods aim at aligning 3D models to objects in RGB images. In this context, many methods train a feed-forward network that directly predicts 3D pose updates given the input image and a 3D model rendering under the current 3D pose estimate~\cite{Li2018deepim,Rad2017iccv,Zakharov2019dpod}. In contrast, more recent methods use differentiable rendering~\cite{Loper2014opendr} to explicitly optimize an objective function conditioned on the input image and renderer inputs like the 3D pose~\cite{Kato2018renderer,Palazzi2018end}. These methods yield more accurate 3D pose updates because they exploit prior knowledge about the rendering pipeline.

However, existing approaches based on differentiable rendering have significant shortcomings because they rely on comparisons in the RGB or mask space. First, methods which compare real-world images and synthetic renderings in the RGB space require photo-realistic 3D model renderings~\cite{Loper2014opendr}. Generating such renderings is difficult because objects in the real world are subject to complex scene lighting, unknown reflection properties, and cluttered backgrounds. Moreover, many 3D models only provide geometry but no textures or materials which makes photo-realistic rendering impossible~\cite{Sun2018pix3d}. Second, methods which rely on comparisons in the mask space need to predict accurate masks from real-world RGB images~\cite{Kato2018renderer,Palazzi2018end}. Generating such masks is difficult even using state-of-the-art approaches like Mask R-CNN~\cite{He2017mask}. Additionally, masks discard valuable object shape information which makes 2D-3D alignment ambiguous. As a consequence, the methods described above are not robust in practice. Finally, computing gradients for the non-differentiable rasterization operation in rendering is still an open research problem and existing approaches rely on hand-crafted approximations for this task~\cite{Henderson18bmvc,Kato2018renderer,Loper2014opendr}.

To overcome these limitations, we compare RGB images and 3D model renderings in a feature space optimized for 3D pose refinement and learn to approximate the rasterization backward pass in differentiable rendering from data. In particular, we introduce a novel network architecture that jointly performs both tasks. Our network maps real-world images and synthetic renderings to a common feature space and predicts deep cross-domain correspondences in the form of \emph{geometric correspondence fields} (see Figure~\ref{fig:teaser}, \textit{middle}). Geometric correspondence fields hold 2D displacement vectors between corresponding 2D object points in RGB images and 3D model renderings similar to optical flow~\cite{Dosovitskiy2015flownet}. These predicted 2D displacement vectors serve as pixel-level gradients that enable us to approximate the rasterization backward pass and compute accurate gradients for renderer inputs like the 3D pose, the 3D model, or the camera intrinsics that minimize an ideal geometric reprojection loss.

Our approach has three main advantages: First, we can leverage depth, normal, and object coordinate~\cite{Brachmann2014learning} renderings which provide 3D pose information more explicitly than RGB and mask renderings~\cite{Li2018deepim}. Second, we avoid task-irrelevant appearance variations in the RGB space and 3D pose ambiguities in the mask space~\cite{Grabner2019b}. Third, we learn to approximate the rasterization backward pass from data instead of relying on a hand-crafted algorithm~\cite{Henderson18bmvc,Kato2018renderer,Loper2014opendr}.

To demonstrate the benefits of our novel 3D pose refinement approach, we evaluate it on the challenging Pix3D~\cite{Sun2018pix3d} dataset. We present quantitative as well as qualitative results and significantly outperform state-of-the-art refinement methods in multiple metrics by up 55\% relative. Finally, we combine our refinement approach with feed-forward 3D pose estimation~\cite{Grabner2019a} and 3D model retrieval~\cite{Grabner2019b} methods to predict fine-grained 3D poses for objects in the wild without providing initial 3D poses or ground truth 3D models at runtime \textbf{given only a single RGB image}. To summarize, our main contributions are:
\begin{itemize}
	\item We introduce the first refinement method based on differentiable rendering that does not compare real-world images and synthetic renderings in the RGB or mask space but in a feature space optimized for the task at hand. 
	\item We present a novel differentiable renderer that learns to approximate the rasterization backward pass instead of relying on a hand-crafted algorithm.
\end{itemize}
\section{Related Work}
\label{sec:relatedwork}

In the following, we discuss prior work on differentiable rendering, 3D pose estimation, and 3D pose refinement.

\subsection{Differentiable Rendering}

Differentiable rendering~\cite{Loper2014opendr} is a powerful concept that provides inverse graphics capabilities by computing gradients for 3D scene parameters from 2D image observations. This novel technique recently gained popularity for 3D reconstruction~\cite{Kundu20183d,Nguyen2018rendernet}, scene lighting estimation~\cite{Azinovic2019inverse,Li2018differentiable}, and texture prediction~\cite{Kanazawa2018learning,Yao20183d}.

However, rendering is a non-differentiable process due to the rasterization operation~\cite{Kato2018renderer}. Thus, differentiable rendering approaches either try to mimic rasterization with differentiable operations~\cite{Liu2019softras,Palazzi2018end} or use conventional rasterization and approximate its backward pass~\cite{Genova2018unsupervised,Henderson18bmvc,Wu2016learning}.

In this work, we also approximate the rasterization backward pass but, in contrast to existing methods, do not rely on hand-crafted approximations. Instead, we train a network that performs the approximation. This idea is not only applicable for 3D pose estimation but also for other tasks like 3D reconstruction, human pose estimation, or the prediction of camera intrinsics in the future.

\subsection{3D Pose Estimation}
Modern 3D pose estimation approaches build on deep feed-forward networks and can be divided into two groups: Direct and correspondence-based methods. 

Direct methods predict 3D pose parameters as raw network outputs. They use classification~\cite{Tulsiani2015pose,Tulsiani2015viewpoints}, regression~\cite{Massa2016crafting,Xiang2016objectnet3d}, or hybrid variants of both \cite{Mahendran2018mixed,Xiao2019pose} to estimate 3D rotation and 3D translation~\cite{Li2018unified,Mottaghi2015coarse,Mousavian20163d} in an end-to-end manner. Recent approaches additionally integrate these techniques into detection pipelines to deal with multiple objects in a single image~\cite{Kehl2017ssd,Kundu20183d,Wang2018fine,Xiang2018posecnn}.

In contrast, correspondence-based methods predict keypoint locations and recover 3D poses from 2D-3D correspondences using \PNP~algorithms~\cite{Peng2019pvnet,Rad2017iccv} or trained shape models~\cite{pavlakos17object3d}. In this context, different methods predict sparse object-specific keypoints~\cite{pavlakos17object3d,Peng2019pvnet,pepik20153d}, sparse virtual control points~\cite{Grabner2018a,Rad2017iccv,Tekin2018real}, or dense unsupervised 2D-3D correspondences~\cite{Brachmann2014learning,Brachmann2016uncertainty,Grabner2019a,Jafari2018ipose,Wang2019normalized}.

In this work, we use the correspondence-based feed-forward approach presented in~\cite{Grabner2019a} to predict initial 3D poses for refinement.

\subsection{3D Pose Refinement}

3D pose refinement methods are based on the assumption that the projection of an object's 3D model aligns with the object's appearance in the image given the correct 3D pose. Thus, they compare renderings under the current 3D pose to the input image to get feedback on the prediction. 

A simple approach to refine 3D poses is to generate many small perturbations and evaluate their accuracy using a scoring function~\cite{Liu2019deep,Zabulis20143d}. However, this is computationally expensive and the design of the scoring function is unclear. Therefore, other approaches try to predict iterative 3D pose updates with deep networks instead~\cite{Li2018deepim,Manhardt2018deep,Rad2017iccv,Zakharov2019dpod}. In practice, though, the performance of these methods is limited because they cannot generalize to 3D poses or 3D models that have not been seen during training~\cite{Rad2017iccv}.

Recent approaches based on differentiable rendering overcome these limitations~\cite{Kato2018renderer,Loper2014opendr,Palazzi2018end}. Compared to the methods described above, they analytically propagate error signals backward through the rendering pipeline to compute more accurate 3D pose updates. In this way, they exploit knowledge about the 3D scene geometry and the projection pipeline for the optimization. 

In contrast to existing differentiable rendering approaches that rely on comparisons in the RGB~\cite{Loper2014opendr} or mask~\cite{Kato2018renderer,Palazzi2018end} space, we compare RGB images and 3D model renderings in a feature space that is optimized for 3D pose refinement.

\section{Learned 3D Pose Refinement}
\label{sec:method}

Given a single RGB image, a 3D model, and an initial 3D pose, we compute iterative updates to refine the 3D pose, as shown in Figure~\ref{fig:system}. For this purpose, we first introduce the objective function that we optimize at runtime (see Sec.~\ref{sec:objective}). We then explain how we compare the input RGB image to renderings under the current 3D pose in a feature space optimized for refinement (see Sec.~\ref{sec:feature-space}), predict pixel-level gradients that minimize an ideal geometric reprojection loss in the form of geometric correspondence fields (see Sec.~\ref{sec:gcf}), and propagate gradients backward through the rendering pipeline to perform a gradient-based optimization directly on the 3D pose (see Sec.~\ref{sec:learned-rendering}).

\newcommand{\xdasharrow}[2][->]{
\tikz[baseline=-\the\dimexpr\fontdimen22\textfont2\relax]{
\node[anchor=south,font=\scriptsize, inner ysep=1.5pt,outer xsep=2.2pt](x){#2};
\draw[dashed,#1](x.south west)--(x.south east);
}
}
\newcommand{\xnormalarrow}[2][->]{
\tikz[baseline=-\the\dimexpr\fontdimen22\textfont2\relax]{
\node[anchor=south,font=\scriptsize, inner ysep=1.5pt,outer xsep=2.2pt](x){#2};
\draw[#1](x.south west)--(x.south east);
}
}
\begin{figure}[t]
	\begin{center}
		\includegraphics[width=0.93\linewidth]{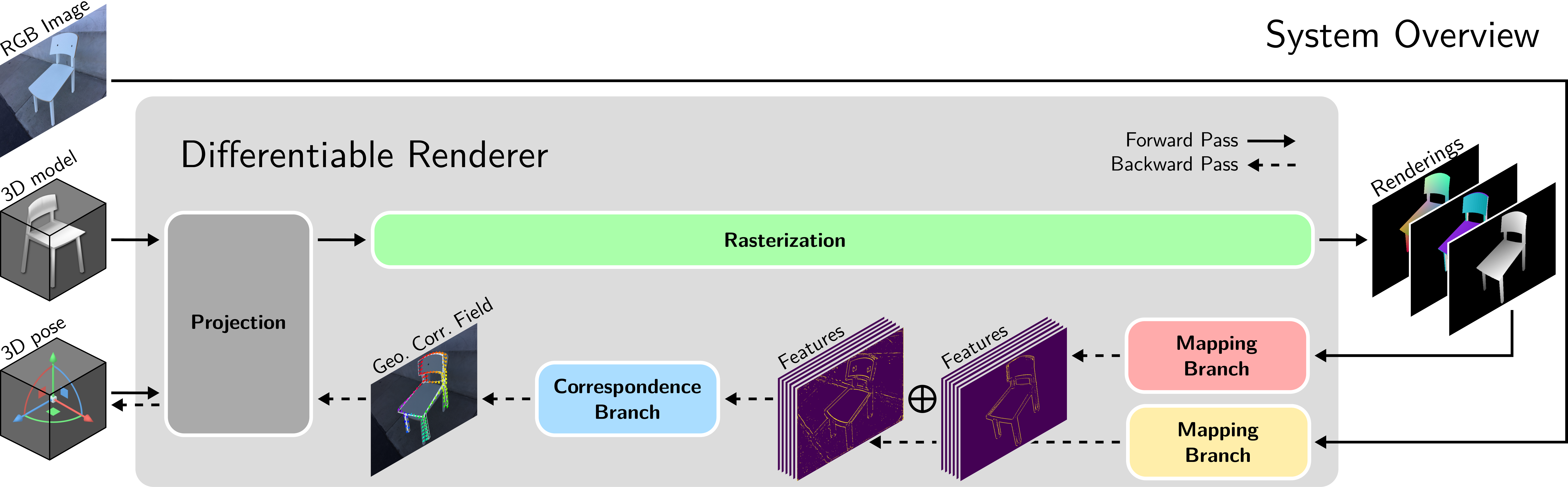}
	\end{center}
	\caption{Overview of our system. In the forward pass ($\xnormalarrow[->,>=latex]{~~~~}$), we generate 3D model renderings under the current 3D pose. In the backward pass ($\xdasharrow[<-,>=latex]{~~~~}$), we map the RGB image and our renderings to a common feature space and predict a geometric correspondence field that enables us to approximate the rasterization backward pass and compute gradients for the 3D pose that minimize an ideal geometric reprojection loss.}
	\label{fig:system}
\end{figure}

\subsection{Runtime Object Function}
\label{sec:objective}

Our approach to refine the 3D pose of an object is based on the numeric optimization of an objective function at runtime. In particular, we seek to minimize an ideal geometric reprojection loss
\begin{equation}
e(\mathcal{P}) = \frac{1}{2} \sum_{i} \Vert \text{proj}(\mathbf{M}_i,\mathcal{P}_\text{gt}) - \text{proj}(\mathbf{M}_i,\mathcal{P}) \Vert^2_2
\label{eq:loss}
\end{equation}
for all provided 3D model vertices $\mathbf{M}_i$. In this case, $\text{proj}(\cdot)$ performs the projection from 3D space to the 2D image plane, $\mathcal{P}$ denotes the 3D pose parameters, and $\mathcal{P}_\text{gt}$ is the ground truth 3D pose. Hence, it is clear that $\argmin e(\mathcal{P}) = \mathcal{P}_\text{gt}$. 

To efficiently minimize $e(\mathcal{P})$ using a gradient-based optimization starting from an initial 3D pose, we compute gradients for the 3D pose using the Jacobian of $e(\mathcal{P})$ with respect to $\mathcal{P}$. Applying the chain rule yields the expression
\begin{equation}
\resizebox{0.93 \textwidth}{!}{$
\left(\frac{\partial e(\mathcal{P})}{\partial \mathcal{P}}\right)(\mathcal{P}_\text{curr}) = \sum_i \left[\frac{\partial \text{proj}(\mathbf{M}_i,\mathcal{P})}{\partial \mathcal{P}}\right]^T \Big[ \text{proj}(\mathbf{M}_i, \mathcal{P}_\text{gt}) - \text{proj}(\mathbf{M}_i, \mathcal{P}_\text{curr})\Big] \> ,$}
\label{eq:jacobian}
\end{equation}
where $\mathcal{P}_\text{curr}$ is the current 3D pose estimate and the point where the Jacobian is evaluated. In this case, the term $\left[\frac{\partial \text{proj}(\mathbf{M}_i,\mathcal{P})}{\partial \mathcal{P}}\right]^T$ can be computed analytically because it is simply a sequence of differentiable operations. In contrast, the term $\big[\text{proj}(\mathbf{M}_i, \mathcal{P}_\text{gt}) - \text{proj}(\mathbf{M}_i, \mathcal{P}_\text{curr})\big]$ cannot be computed analytically because the 3D model vertices projected under the ground truth 3D pose, \ie, $\text{proj}(\mathbf{M}_i, \mathcal{P}_\text{gt})$, are unknown at runtime and can only be observed indirectly via the input image. 

However, for visible vertices, this term can be calculated given a geometric correspondence field (see Sec.~\ref{sec:learned-rendering}). Thus, we introduce a novel network architecture that learns to predict geometric correspondence fields given an RGB image and 3D model renderings under the current 3D pose estimate in the following. Moreover, we embed this network in a differentiable rendering framework to approximate the rasterization backward pass and compute gradients for renderer inputs like the 3D pose of an object in an end-to-end manner (see Figure~\ref{fig:system}).

\subsection{Refinement Feature Space}
\label{sec:feature-space}

The first step in our approach is to render the provided 3D model under the current 3D pose using the forward pass of our differentiable renderer (see Figure~\ref{fig:system}). In particular, we generate depth, normal, and object coordinate~\cite{Brachmann2014learning} renderings. These representations provide 3D pose and 3D shape information more explicitly than RGB or mask renderings which makes them particularly useful for 3D pose refinement~\cite{Grabner2019b}. By concatenating the different renderings along the channel dimension, we leverage complementary information from different representations in the backward pass rather than relying on a single type of rendering~\cite{Loper2014opendr}.

Next, we begin the backward pass of our differentiable renderer by mapping the input RGB image and our multi-representation renderings to a common feature space. For this task, we use two different network branches that bridge the domain gap between the real and rendered images (see Figure~\ref{fig:system}). Our mapping branches use a custom architecture based on task-specific design choices:

First, we want to predict local cross-domain correspondences under the assumption that the initial 3D pose is close to the ground truth 3D pose. Thus, we do not require features with global context but features with local discriminability. For this reason, we use small fully convolutional networks which are fast, memory-efficient, and learn low-level features that generalize well across different objects. Because the low-level structures appearing across different objects are similar, we do not require a different network for each object~\cite{Rad2017iccv} but address objects of all categories with a single class-agnostic network for each domain. 

Second, we want to predict correspondences with maximum spatial accuracy. Thus, we do not use pooling or downsampling but maintain the spatial resolution throughout the network. In this configuration, consecutive convolutions provide sufficient receptive field to learn advanced shape features which are superior to simple edge intensities~\cite{Kehl2017ssd}, while higher layers benefit from full spatial parameter sharing during training which increases generalization. As a consequence, the effective minibatch size during training is higher than the number of images per minibatch because only a subset of all image pixels contributes to each computed feature. In addition, the resulting high spatial resolution feature space provides an optimal foundation for computing spatially accurate correspondences.

For the implementation of our mapping branches, we use fully convolutional networks consisting of an initial $7\times7$ Conv-BN-ReLU block, followed by three residual blocks~\cite{He2016deep,He2016identity}, and a $1\times1$ Conv-BN-ReLU block for dimensionality reduction. This architecture enforces local discriminability and high spatial resolution and maps RGB images and multi-representation renderings to $W\times H\times64$ feature maps, where $W$ and $H$ are the spatial dimensions of the input image.

\subsection{Geometric Correspondence Fields}
\label{sec:gcf}

After mapping RGB images and 3D model renderings to a common feature space, we compare their feature maps and predict cross-domain correspondences. For this purpose, we concatenate their feature maps and use another fully convolutional network branch to predict correspondences, as shown in Figure~\ref{fig:system}.

In particular, we regress per-pixel correspondence vectors in the form of geometric correspondence fields (see Figure~\ref{fig:teaser}, \textit{middle}). Geometric correspondence fields hold 2D displacement vectors between corresponding 2D object points in real-world RGB images and synthetic 3D model renderings similar to optical flow~\cite{Dosovitskiy2015flownet}. These displacement vectors represent the projected relative 2D motion of individual 2D object points that is required to minimize the reprojection error and refine the 3D pose. A geometric correspondence field has the same spatial resolution as the respective input RGB image and two channels, \ie, $W\times H\times2$. 

If an object's 3D model and 3D pose are known, we can render the ground truth geometric correspondence field for an arbitrary 3D pose. For this task, we first compute the 2D displacement $\nabla \mathbf{m}_i = \text{proj}(\mathbf{M}_i,\mathcal{P}_\text{gt}) - \text{proj}(\mathbf{M}_i,\mathcal{P}_\text{curr})$ between the projection under the ground truth 3D pose $\mathcal{P}_\text{gt}$ and the current 3D pose $\mathcal{P}_\text{curr}$ for each 3D model vertex $\mathbf{M}_i$. We then generate a ground truth geometric correspondence field $G(\mathcal{P}_\text{curr},\mathcal{P}_\text{gt})$ by rendering the 3D model using a shader that interpolates the per-vertex 2D displacements $\nabla \mathbf{m}_i$ across the projected triangle surfaces using barycentric coordinates.

In our scenario, predicting correspondences using a network has two advantages compared to traditional correspondence matching~\cite{Hartley2003multiple}. First, predicting correspondences with convolutional kernels is significantly faster than exhaustive feature matching during both training and testing~\cite{Zagoruyko2015learning}. This is especially important in the case of dense correspondences. Second, training explicit correspondences can easily result in degenerated feature spaces and requires tedious regularization and hard negative sample mining~\cite{Choy2016universal}.

For the implementation of our correspondence branch, we use three consecutive $7\times7$ Conv-BN-ReLU blocks followed by a final $7\times7$ convolution which reduces the channel dimensionality to two. For this network, a large receptive field is crucial to cover correspondences with high spatial displacement.

However, in many cases local correspondence prediction is ambiguous. For example, many objects are untextured and have homogeneous surfaces, \eg, the backrest and the seating surface of the chair in Figure~\ref{fig:teaser}, which cause unreliable correspondence predictions. To address this problem, we additionally employ a geometric attention module which restricts the correspondence prediction to visible object regions with significant geometric discontinuities, as outlined in white underneath the 2D displacement vectors in Figure~\ref{fig:teaser}. We identify these regions by finding local variations in our renderings. 

In particular, we detect rendering-specific intensity changes larger than a certain threshold within a local $5\times5$ window to construct a geometric attention mask $w^{att}$. For each pixel of $w^{att}$, we compute the geometric attention weight
\begin{equation}
w^{att}_{x,y} = \underset{u,v \in W}{\max}\bigg( \delta^R\Big( R(\mathcal{P}_\text{curr})_{x,y},R(\mathcal{P}_\text{curr})_{x-u,y-v}\Big)\bigg) > t^R \> .
\end{equation}
In this case, $R(\mathcal{P}_\text{curr})$ is a concatenation of depth, normal, and object coordinate renderings under the current 3D pose $\mathcal{P}_\text{curr}$, $(x,y)$ is a pixel location, and $(u,v)$ are pixel offsets within the window $W$. The comparison function $\delta^R(\cdot)$ and the threshold $t^R$ are different for each type of rendering. For depth renderings, we compute the absolute difference between normalized depth values and use a threshold of $0.1$. For normal renderings, we compute the angle between normals and use a threshold of $15^\circ$. For object coordinate renderings, we compute the Euclidean distance between 3D points and use a threshold of $0.1$. If any of these thresholds applies, the corresponding pixel $(x,y)$ in our geometric attention mask $w^{att}$ becomes active. Because we already generated these renderings before, our geometric attention mechanism requires almost no additional computations and is available during both training and testing.

\textbf{Training.} During training of our system, we optimize the learnable part of our differentiable renderer, \ie, a joint network $f(\cdot)$ consisting of our two mapping branches and our correspondence branch with parameters $\theta$ (see Figure~\ref{fig:system}). Formally, we minimize the error between predicted $f(\cdot)$ and ground truth $G(\cdot)$ geometric correspondence fields as
\begin{equation}
\underset{\theta}{\text{min}} \sum_{x,y} w^{att}_{x,y} \Vert f(I, R(\mathcal{P}_\text{curr}); \theta)_{x,y} - G(\mathcal{P}_\text{curr},\mathcal{P}_\text{gt})_{x,y} \Vert^2_2 \> .
\label{eq:train}
\end{equation}
In this case, $w^{att}$ is a geometric attention mask, $I$ is an RGB image, $R(\mathcal{P}_\text{curr})$ is a concatenation of depth, normal, and object coordinate renderings generated under a random 3D pose $\mathcal{P}_\text{curr}$ produced by perturbing the ground truth 3D pose $\mathcal{P}_\text{gt}$, $G(\mathcal{P}_\text{curr},\mathcal{P}_\text{gt})$ is the ground truth geometric correspondence field, and $(x,y)$ is a pixel location. In particular, we first generate a random 3D pose $\mathcal{P}_\text{curr}$ around the ground truth 3D pose $\mathcal{P}_\text{gt}$ for each training sample in each iteration. For this purpose, we sample 3D pose perturbations from normal distributions and apply them to $\mathcal{P}_\text{gt}$ to generate $\mathcal{P}_\text{curr}$. For 3D rotations, we use absolute perturbations with $\sigma=5^\circ$. For 3D translations, we use relative perturbations with $\sigma=0.1$. We then render the ground truth geometric correspondence field $G(\mathcal{P}_\text{curr},\mathcal{P}_\text{gt})$ between the perturbed 3D pose $\mathcal{P}_\text{curr}$ and the ground truth 3D pose $\mathcal{P}_\text{gt}$ as described above, generate concatenated depth, normal, and object coordinate renderings $R(\mathcal{P}_\text{curr})$ under the perturbed 3D pose $\mathcal{P}_\text{curr}$, and compute the geometric attention mask $w^{att}$. Finally, we predict a geometric correspondence field using our network $f(I, R(\mathcal{P}_\text{curr}); \theta)$ given the RGB image $I$ and the renderings $R(\mathcal{P}_\text{curr})$, and optimize for the network parameters $\theta$. 

In this way, we train a network that performs three tasks: First, it maps RGB images and multi-representation 3D model renderings to a common feature space. Second, it compares features in this space. Third, it predicts geometric correspondence fields which serve as pixel-level gradients that enable us to approximate the rasterization backward pass of our differentiable renderer.

\subsection{Learned Differentiable Rendering}
\label{sec:learned-rendering}

\begin{figure}
\centering
\begin{subfigure}{0.22\linewidth}
	\begin{center}
		\includegraphics[width=\linewidth]{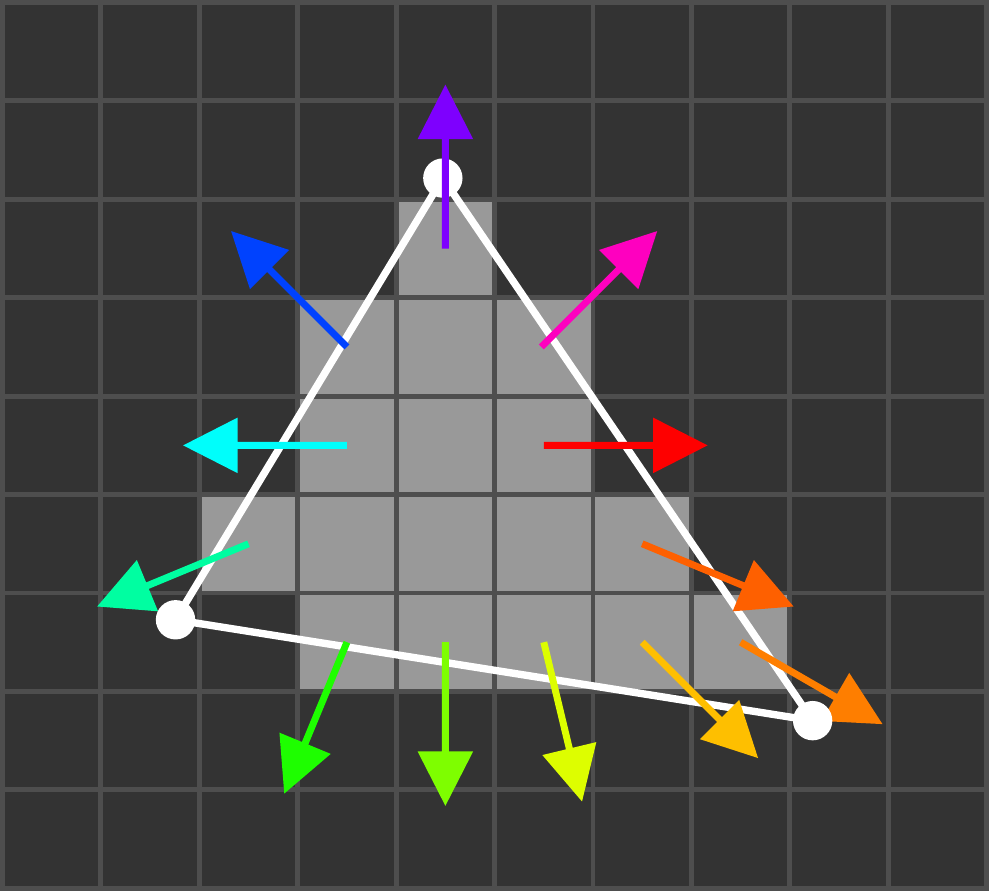}
	\end{center}
\end{subfigure}\qquad\begin{subfigure}{0.22\linewidth}
	\begin{center}
		\includegraphics[width=\linewidth]{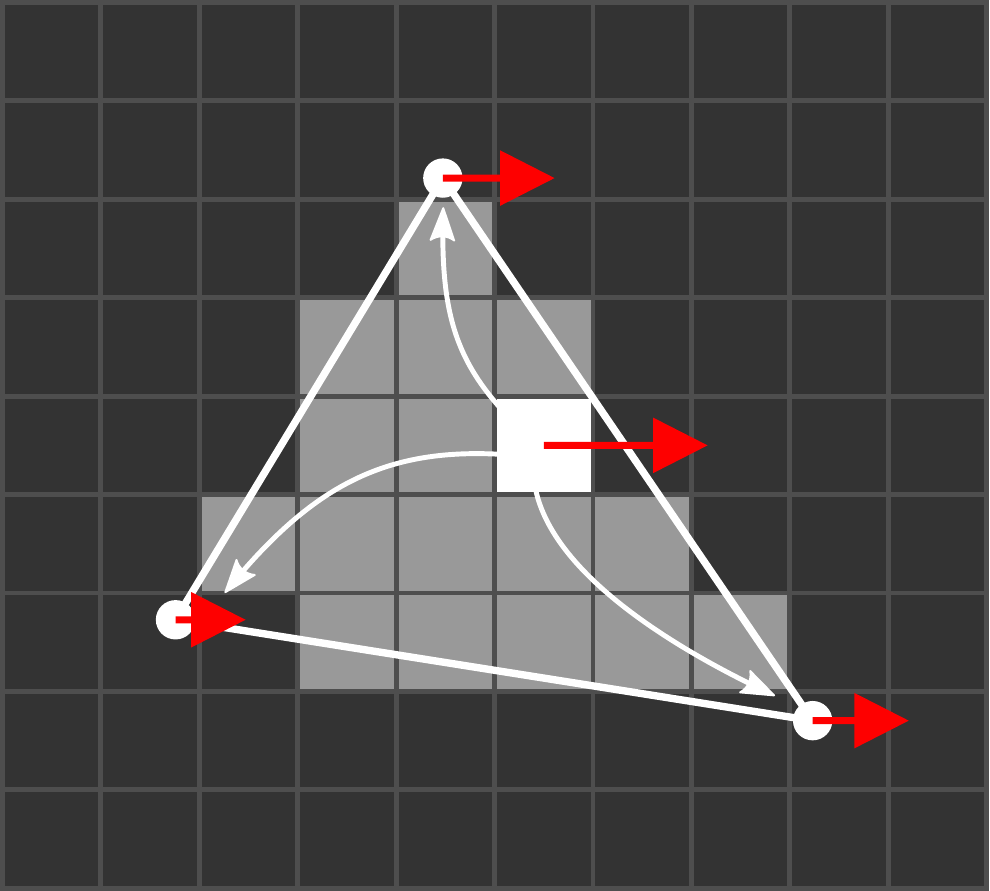}
	\end{center}
\end{subfigure}\qquad\begin{subfigure}{0.22\linewidth}
	\begin{center}
		\includegraphics[width=\linewidth]{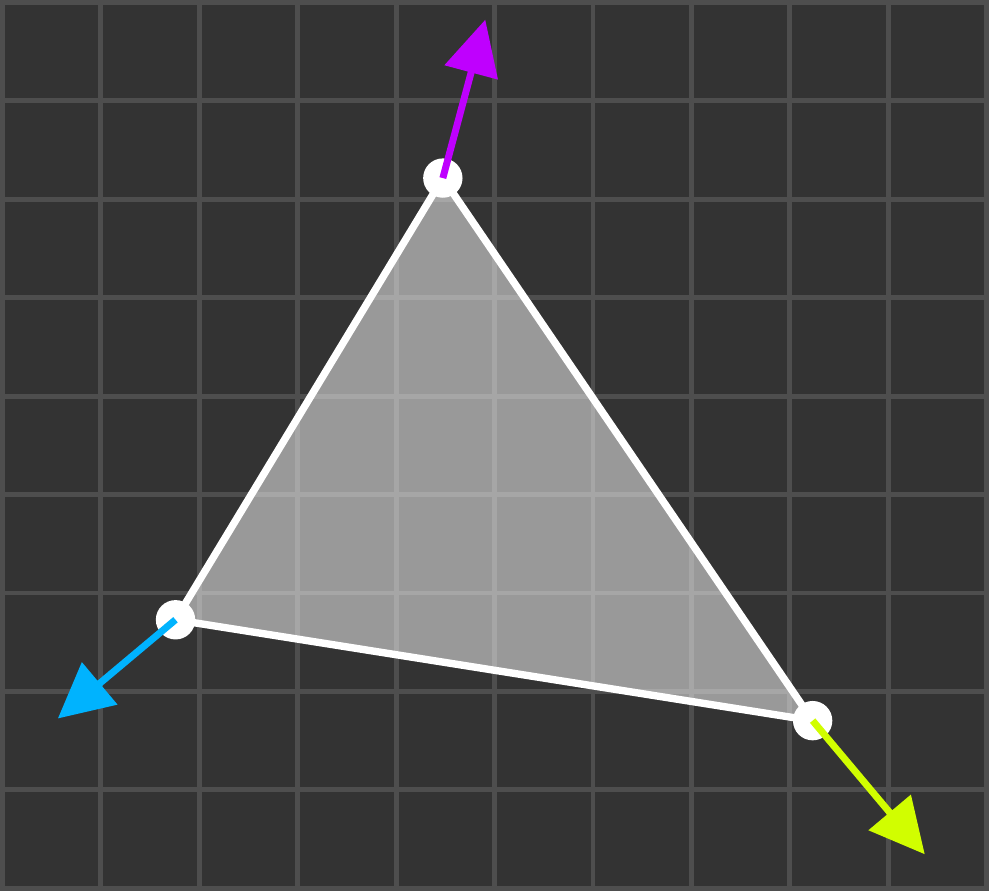}
	\end{center}
\end{subfigure}
\caption{To approximate the rasterization backward pass, we predict a geometric correspondence field (\textit{left}), disperse the predicted 2D displacement of each pixel among the vertices of its corresponding visible triangle (\textit{middle}), and normalize the contributions of all pixels. In this way, we obtain gradients for projected 3D model vertices (\textit{right}).}
\label{fig:triangle}
\end{figure}

In the classic rendering pipeline, the only non-differentiable operation is the rasterization~\cite{Loper2014opendr} that determines which pixels of a rendering have to be filled, solves the visibility of projected triangles, and fills the pixels using a shading computation. This discrete operation raises one main challenge: Its gradient is zero, which prevents gradient flow~\cite{Kato2018renderer}. However, we must flow non-zero gradients from pixels to projected 3D model vertices to perform differentiable rendering.

We solve this problem using geometric correspondence fields. Instead of actually differentiating a loss in the image space and relying on hand-crafted comparisons between pixel intensities to approximate the gradient flow from pixels to projected 3D model vertices~\cite{Henderson18bmvc,Kato2018renderer}, we first use a network to predict per-pixel 2D displacement vectors in the form of a geometric correspondence field, as shown in Figure~\ref{fig:system}. We then compute gradients for projected 3D model vertices by simply accumulating the predicted 2D displacement vectors using our knowledge of the projected 3D model geometry, as illustrated in Figure~\ref{fig:triangle}. 

Formally, we compute the gradient of a projected 3D model vertex $\mathbf{m}_i$ as
\begin{equation}
\begin{gathered}
\nabla \mathbf{m}_i = \frac{1}{\underset{x,y}\sum{w^{att\phantom{,}}_{x,y} w^{bar,i}_{x,y}}} \sum_{x,y} w^{att\phantom{,}}_{x,y} w^{bar,i}_{x,y} \: f(I, R(\mathcal{P}_\text{curr}); \theta)_{x,y}\\
\forall x, y : \mathbf{m}_i \in \bigtriangleup_{\texttt{IndexMap}_{x,y}} \> .
\end{gathered}
\end{equation}
In this case, $f(I, R(\mathcal{P}_\text{curr}); \theta)$ is a geometric correspondence field predicted by our network $f(\cdot)$ with frozen parameters $\theta$ given an RGB image $I$ and concatenated 3D model renderings $R(\mathcal{P}_\text{curr})$ under the current 3D pose estimate $\mathcal{P}_\text{curr}$, $w^{att\phantom{,}}_{x,y}$ is a geometric attention weight, $w^{bar,i}_{x,y}$ is a barycentric weight for $\mathbf{m}_i$, and $(x,y)$ is a pixel position. We accumulate predicted 2D displacement vectors for all positions $(x,y)$ for which $\mathbf{m}_i$ is a vertex of the triangle $\bigtriangleup_{\texttt{IndexMap}_{x,y}}$ visible at $(x,y)$. For this task, we generate an \texttt{IndexMap} which stores the index of the visible triangle for each pixel during the forward pass of our differentiable renderer.

\textbf{Inference.} Our computed $\nabla \mathbf{m}_i$ approximate the second term in Eq.~(\ref{eq:jacobian}) that cannot be computed analytically. In this way, our approach combines local per-pixel 2D displacement vectors into per-vertex gradients and further computes accurate global 3D pose gradients considering the 3D model geometry and the rendering pipeline. Our experiments show that this approach generalizes better to unseen data than predicting 3D pose updates with a network~\cite{Li2018deepim,Zakharov2019dpod}.

During inference of our system, we perform iterative updates to refine $\mathcal{P}_\text{curr}$. In each iteration, we compute a 3D pose gradient by evaluating our refinement loop presented in Figure~\ref{fig:system}. For our implementation, we use the Adam optimizer~\cite{Kingma2014adam} with a small learning rate and perform multiple updates to account for noisy correspondences and achieve the best accuracy.

\section{Experimental Results}
\label{sec:experiments}

\definecolor{light_red}{rgb}{0.8, 1, 0.8}
\definecolor{light_blue}{rgb}{0.8, 0.9, 1}

To demonstrate the benefits of our 3D pose refinement approach, we evaluate it on the challenging Pix3D~\cite{Sun2018pix3d} dataset which provides in-the-wild images for objects of different categories. In particular, we quantitatively and qualitatively compare our approach to state-of-the-art refinement methods in Sec.~\ref{sec:sota}, perform an ablation study in Sec.~\ref{sec:ablation}, and combine our refinement approach with feed-forward 3D pose estimation~\cite{Grabner2019a} and 3D model retrieval~\cite{Grabner2019b} methods to predict fine-grained 3D poses without providing initial 3D poses or ground truth 3D models in Sec.~\ref{sec:retrieval}. We follow the evaluation protocol of~\cite{Grabner2019a} and report the median ($MedErr$) of rotation, translation, pose, and projection distances. Details on evaluation setup, datasets, and metrics as well as extensive results and further experiments are provided in our \textbf{supplementary material}.

\subsection{Comparison to the State of the Art}
\label{sec:sota}

We first quantitatively compare our approach to state-of-the-art refinement methods. For this purpose, we perform 3D pose refinement on top of an initial 3D pose estimation baseline. In particular, we predict initial 3D poses using the feed-forward approach presented in~\cite{Grabner2019a} which is the state of the art for single image 3D pose estimation on the Pix3D dataset (\textit{Baseline}). We compare our refinement approach to traditional image-based refinement without differentiable rendering~\cite{Zakharov2019dpod} (\textit{Image Refinement}) and mask-based refinement with differentiable rendering~\cite{Kato2018renderer} (\textit{Mask Refinement}). 

RGB-based refinement with differentiable rendering~\cite{Loper2014opendr} is not possible in our setup because all available 3D models lack textures and materials. This approach even fails if we compare grey-scale images and renderings because the image intensities at corresponding locations do not match. As a consequence, 2D-3D alignment using a photo-metric loss is impossible.

For \textit{Image Refinement}, we use grey-scale instead of RGB renderings because all available 3D models lack textures and materials. In addition, we do not perform a single full update~\cite{Zakharov2019dpod} but perform up to 1000 iterative updates with a small learning rate of $\eta = 0.05$ using the Adam optimizer~\cite{Kingma2014adam} for all methods.

For \textit{Mask Refinement}, we predict instance masks from the input RGB image using Mask R-CNN~\cite{He2017mask}. To achieve maximum accuracy, we internally predict masks at four times the original spatial resolution proposed in Mask R-CNN and fine-tune a model pre-trained on COCO~\cite{Lin2014microsoft} on Pix3D. 

Table~\ref{table:pix3d} ({\setlength\fboxsep{0.5pt}\colorbox{light_red}{upper part}}) summarizes our results. In this experiment, we provide the ground truth 3D model of the object in the image for refinement. Compared to the baseline, \textit{Image Refinement} only achieves a small improvement in the rotation, translation, and pose metrics. There is almost no improvement in the projection metric ($MedErr_{P}$), as this method does not minimize the reprojection error. Traditional refinement methods are not aware of the rendering pipeline and the underlying 3D scene geometry and can only provide coarse 3D pose updates~\cite{Zakharov2019dpod}. In our in-the-wild scenario, the number of 3D models, possible 3D pose perturbations, and category-level appearance variations is too large to simulate all permutations during training. As a consequence, this method cannot generalize to examples which are far from the ones seen during training and only achieves small improvements.

\begin{table*}[t]
	\centering
	\caption{\setlength\fboxsep{0.5pt} Quantitative 3D pose refinement results on the Pix3D dataset. In the case of provided \colorbox{light_red}{ground truth 3D models}, our refinement significantly outperforms previous refinement methods across all metrics by \textbf{up to 55\%} relative. In the case of automatically \colorbox{light_blue}{retrieved 3D models} (+ Retrieval~\cite{Grabner2019b}), we reduce the 3D pose error ($MedErr_{R,t}$) compared to the state of the art for single image 3D pose estimation on Pix3D (\textit{Baseline}) \textbf{by 55\%} relative \textbf{without using additional inputs}.}
	\label{table:pix3d}
	\resizebox{\columnwidth}{!}{%
		\setlength{\tabcolsep}{4pt}
		\begin{tabular}{lc|c|c|c|c}
			\toprule
			\multicolumn{2}{c}{}&\multicolumn{1}{c}{\bf Rotation}&\multicolumn{1}{c}{\bf Translation}&\multicolumn{1}{c}{\bf Pose}&\multicolumn{1}{c}{\bf Projection}\\
			\cmidrule(lr){3-3}\cmidrule(lr){4-4}\cmidrule(lr){5-5}\cmidrule(lr){6-6}
			\multirow{2}{*}{Method}&\multicolumn{1}{c}{\multirow{2}{*}{Input}}&\multicolumn{1}{c}{$MedErr_R$}&\multicolumn{1}{c}{$MedErr_{t}$}&\multicolumn{1}{c}{$MedErr_{R,t}$}&\multicolumn{1}{c}{$MedErr_{P}$}\\
			&\multicolumn{1}{c}{}&\multicolumn{1}{c}{$\cdot1$}&\multicolumn{1}{c}{$\cdot10^{2}$}&\multicolumn{1}{c}{$\cdot10^{2}$}&\multicolumn{1}{c}{$\cdot10^{2}$}\\
			\midrule
			Baseline~\cite{Grabner2019a}&RGB Image&6.75&6.21&4.76&3.71\\
			\midrule
			Image Refinement~\cite{Zakharov2019dpod}&RGB Image + 3D Model&\cellcolor{light_red}6.46&\cellcolor{light_red}5.43&\cellcolor{light_red}4.31&\cellcolor{light_red}3.67\\
			Mask Refinement~\cite{Kato2018renderer}&RGB Image + 3D Model&\cellcolor{light_red}3.56&\cellcolor{light_red}4.06&\cellcolor{light_red}2.96&\cellcolor{light_red}1.90\\
			Our Refinement&RGB Image + 3D Model&\cellcolor{light_red}\bf2.56&\cellcolor{light_red}\bf1.74&\cellcolor{light_red}\bf1.34&\cellcolor{light_red}\bf1.27\\
			\midrule
			Image Refinement~\cite{Zakharov2019dpod} + Retrieval~\cite{Grabner2019b}&RGB Image&\cellcolor{light_blue}6.47&\cellcolor{light_blue}5.51&\cellcolor{light_blue}4.33&\cellcolor{light_blue}3.74\\
			Mask Refinement~\cite{Kato2018renderer} + Retrieval~\cite{Grabner2019b}&RGB Image&\cellcolor{light_blue}5.47&\cellcolor{light_blue}5.25&\cellcolor{light_blue}4.15&\cellcolor{light_blue}3.12\\
			Our Refinement + Retrieval~\cite{Grabner2019b}&RGB Image&\cellcolor{light_blue}\bf3.79&\cellcolor{light_blue}\bf2.65&\cellcolor{light_blue}\bf2.14&\cellcolor{light_blue}\bf2.18\\
			\bottomrule
	\end{tabular}}
\end{table*}

\begin{figure}[t]
	\centering
	\begin{subfigure}{0.5\linewidth}
		\begin{center}
\pgfplotsset{
	compat=1.11,
	legend image code/.code={
		\draw[mark repeat=1,mark phase=1]
		plot coordinates {
			(0cm,0cm)
			(0.4cm,0cm) 
		};%
	}
}
\begin{tikzpicture}[
scale=.65,
]
\begin{axis}[
width = 7cm,
height = 6cm,
ytick={0,0.225,0.45,0.675,0.9},
yticklabels={0,,,,0.9},
xtick={0,0.025,0.05,0.075,0.1},
xticklabels={0,,,,0.1},
ymin=-0.06,
xmin=-0.005,
ymax=0.9,
xmax=0.1,
xlabel=$Threshold$,
ylabel=$Acc_{R,t}$,
axis x line*=bottom, 
axis y line*=left, 
x label style={at={(axis description cs:0.5,-0.05)},anchor=north},
y label style={at={(axis description cs:-0.025,.54)},anchor=south},
legend style={at={(1,0.08)},anchor=south east,font=\small},
legend cell align={left},
every axis plot/.append style={line width=1.5pt}]

\definecolor{alex_baseline}{HTML}{666666}
\definecolor{alex_image}{HTML}{66FF66}
\definecolor{alex_mask}{HTML}{FF6666}
\definecolor{alex_ours}{HTML}{66AAFF}

\addplot[smooth,alex_baseline] plot coordinates {
	(0.0,0.0)
	(0.005,0.013815674089411586)
	(0.01,0.04598079567808886)
	(0.015,0.10979771422081726)
	(0.02,0.17242749092806428)
	(0.025,0.23872813625555905)
	(0.03,0.3016946181843129)
	(0.035,0.37256109717710273)
	(0.04,0.43970539910477596)
	(0.045,0.5023997992625873)
	(0.05,0.5507660254959469)
	(0.055,0.5930581540772482)
	(0.06,0.6320705361513259)
	(0.065,0.6719870533223959)
	(0.07,0.7005381132204918)
	(0.075,0.7364967524711855)
	(0.08,0.7576534389446485)
	(0.085,0.7702864073844957)
	(0.09,0.7867112847241701)
	(0.095,0.8061050365024593)
	(0.1,0.8134529815016636)
};
\addlegendentry{Baseline~\cite{Grabner2019a}}

\addplot[smooth,alex_image] plot coordinates {
(0.0,0.0)
(0.005,0.016673440232518044)
(0.01,0.0704356327047834)
(0.015,0.13530095427001593)
(0.02,0.21087262151979172)
(0.025,0.29369432495328357)
(0.03,0.3650549432587319)
(0.035,0.4348757902332294)
(0.04,0.5034170259353958)
(0.045,0.5511081136765764)
(0.05,0.5990672935229493)
(0.055,0.6342499489645181)
(0.06,0.6753782326716754)
(0.065,0.7045632200983982)
(0.07,0.7296139885279027)
(0.075,0.7566523482533184)
(0.08,0.7763849207162876)
(0.085,0.7961827259605901)
(0.09,0.8102191169565598)
(0.095,0.823209323022984)
(0.1,0.8357030061865449)
};
\addlegendentry{Image Ref.~\cite{Zakharov2019dpod}\hspace*{-1cm}}

\addplot[smooth,alex_mask] plot coordinates {
(0.0,0.0)
(0.005,0.03036374693416352)
(0.01,0.10058753510422264)
(0.015,0.19640908535604282)
(0.02,0.30345369915790277)
(0.025,0.417301806946216)
(0.03,0.5232384988978528)
(0.035,0.5958203783378806)
(0.04,0.6413439250066209)
(0.045,0.6810259035644539)
(0.05,0.7109776150144604)
(0.055,0.7324546488111904)
(0.06,0.7515221173958369)
(0.065,0.7637601895257267)
(0.07,0.7780958489765843)
(0.075,0.7836419228295369)
(0.08,0.7931529162957487)
(0.085,0.800444557304969)
(0.09,0.8107241054734151)
(0.095,0.8151894876837165)
(0.1,0.8201870053324932)
};
\addlegendentry{Mask Ref.~\cite{Kato2018renderer}}

\addplot[smooth,alex_ours] plot coordinates {
(0.0,0.0)
(0.005,0.13719413261109475)
(0.01,0.3751944248749325)
(0.015,0.5577922954326155)
(0.02,0.6621462243823696)
(0.025,0.7361721280113029)
(0.03,0.77923952199323)
(0.035,0.8151707590911612)
(0.04,0.8346711829790425)
(0.045,0.8430332526523512)
(0.05,0.853642076935414)
(0.055,0.8606297913870676)
(0.06,0.8698652007737362)
(0.065,0.8723012213653764)
(0.07,0.8767119647673988)
(0.075,0.8807779516143655)
(0.08,0.8814210298511036)
(0.085,0.8840289553336689)
(0.09,0.8849213213196948)
(0.095,0.887122587829281)
(0.1,0.8889073198013329)
};
\addlegendentry{Our Ref.}

\end{axis}
\end{tikzpicture}
			\caption{GT 3D models}
			\label{fig:curves-A}
		\end{center}
	\end{subfigure}\begin{subfigure}{0.5\linewidth}
		\begin{center}
\pgfplotsset{
	compat=1.11,
	legend image code/.code={
		\draw[mark repeat=1,mark phase=1]
		plot coordinates {
			(0cm,0cm)
			(0.4cm,0cm) 
		};%
	}
}
\begin{tikzpicture}[
scale=.65,
]
\begin{axis}[
width = 7cm,
height = 6cm,
ytick={0,0.225,0.45,0.675,0.9},
yticklabels={0,,,,0.9},
xtick={0,0.025,0.05,0.075,0.1},
xticklabels={0,,,,0.1},
ymin=-0.06,
xmin=-0.005,
ymax=0.9,
xmax=0.1,
xlabel=$Threshold$,
ylabel=$Acc_{R,t}$,
axis x line*=bottom, 
axis y line*=left, 
x label style={at={(axis description cs:0.5,-0.05)},anchor=north},
y label style={at={(axis description cs:-0.025,.54)},anchor=south},
legend style={at={(1,0.08)},anchor=south east,font=\small},
legend cell align={left},
every axis plot/.append style={line width=1.5pt}]

\definecolor{alex_baseline}{HTML}{666666}
\definecolor{alex_image}{HTML}{66FF66}
\definecolor{alex_mask}{HTML}{FF6666}
\definecolor{alex_ours}{HTML}{66AAFF}

\addplot[smooth,alex_baseline] plot coordinates {
(0.0,0.0)
(0.005,0.013815674089411586)
(0.01,0.04598079567808886)
(0.015,0.10979771422081726)
(0.02,0.17242749092806428)
(0.025,0.23872813625555905)
(0.03,0.3016946181843129)
(0.035,0.37256109717710273)
(0.04,0.43970539910477596)
(0.045,0.5023997992625873)
(0.05,0.5507660254959469)
(0.055,0.5930581540772482)
(0.06,0.6320705361513259)
(0.065,0.6719870533223959)
(0.07,0.7005381132204918)
(0.075,0.7364967524711855)
(0.08,0.7576534389446485)
(0.085,0.7702864073844957)
(0.09,0.7867112847241701)
(0.095,0.8061050365024593)
(0.1,0.8134529815016636)
};
\addlegendentry{Baseline~\cite{Grabner2019a}}

\addplot[smooth,alex_image] plot coordinates {
(0.0,0.0)
(0.005,0.015429054081791833)
(0.01,0.0683551845580151)
(0.015,0.13621961658629733)
(0.02,0.20778254130435125)
(0.025,0.28682631659399616)
(0.03,0.36212650495565385)
(0.035,0.438935066381805)
(0.04,0.5042717267900967)
(0.045,0.551455475738598)
(0.05,0.5921677778194745)
(0.055,0.6337659071331764)
(0.06,0.6731967172911996)
(0.065,0.7031340605712183)
(0.07,0.7222386413759364)
(0.075,0.7517502994103096)
(0.08,0.7694788598116675)
(0.085,0.7844322479422138)
(0.09,0.798600660604825)
(0.095,0.8128962149146376)
(0.1,0.8273836471612472)
};
\addlegendentry{Image Ref.~\cite{Zakharov2019dpod}\hspace*{-1cm}}

\addplot[smooth,alex_mask] plot coordinates {
(0.0,0.0)
(0.005,0.02631489926139654)
(0.01,0.07953058812519861)
(0.015,0.16237441338211592)
(0.02,0.24815494374447705)
(0.025,0.3434130795783405)
(0.03,0.4314737090215386)
(0.035,0.49263211204533236)
(0.04,0.5305966440081593)
(0.045,0.5719095355945243)
(0.05,0.6041988059094295)
(0.055,0.6232088646815852)
(0.06,0.6452372256867029)
(0.065,0.6687835043263775)
(0.07,0.6869988775632891)
(0.075,0.7018258489030778)
(0.08,0.7170020694033549)
(0.085,0.7289537415638854)
(0.09,0.7380482004925627)
(0.095,0.7446585452224663)
(0.1,0.7506030676655479)
};
\addlegendentry{Mask Ref.~\cite{Kato2018renderer}}

\addplot[smooth,alex_ours] plot coordinates {
	(0.0,0.0)
	(0.005,0.09972051867524656)
	(0.01,0.28266718811195024)
	(0.015,0.435072931842173)
	(0.02,0.5066063953638074)
	(0.025,0.5732304900132729)
	(0.03,0.6216835872555428)
	(0.035,0.6622187734315621)
	(0.04,0.6916993115357556)
	(0.045,0.7096626260191675)
	(0.05,0.7258695574564956)
	(0.055,0.7394629585353742)
	(0.06,0.756604744689978)
	(0.065,0.7662250157727474)
	(0.07,0.7765364586993886)
	(0.075,0.7917825700836776)
	(0.08,0.7974060267949928)
	(0.085,0.8093675745200783)
	(0.09,0.8161060012224444)
	(0.095,0.8218204276861503)
	(0.1,0.8255782172868794)
};
\addlegendentry{Our Ref.}

\end{axis}
\end{tikzpicture}
			\caption{Retrieved 3D models}
			\label{fig:curves-B}
		\end{center}
	\end{subfigure}
	\caption{Evaluation on 3D pose accuracy at different thresholds. We significantly outperform other methods on strict thresholds using both GT and retrieved 3D models.}
	\label{fig:curves}
\end{figure}
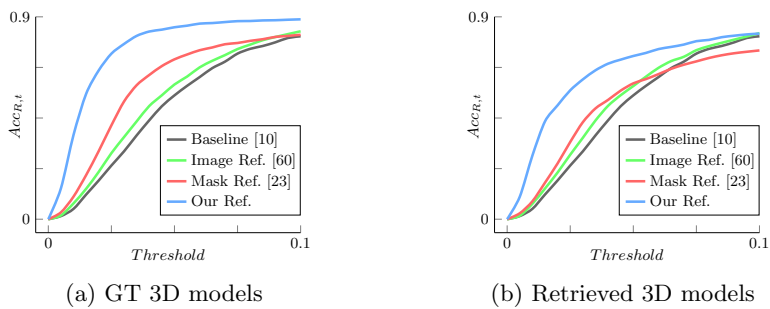

Additionally, we observe that after the first couple of refinement steps, the predicted updates are not accurate enough to refine the 3D pose but start to jitter without further improving the 3D pose. Moreover, for many objects, the prediction fails and the iterative updates cause the 3D pose to drift off. Empirically, we obtain the best overall results for this method using only 20 iterations. For all other methods based on differentiable rendering, we achieve the best accuracy after the full 1000 iterations. A detailed analysis on this issue is presented in our supplementary material.

\begin{figure}[t]
	\setlength{\tabcolsep}{1pt}
	\setlength{\fboxsep}{-1pt}
	\setlength{\fboxrule}{1pt}
	\definecolor{boxgreen}{rgb}{0.3, 1.0, 0.3}
	\definecolor{boxred}{rgb}{1.0, 0.3, 0.3}
	\newcommand{\colImgN}[1]{{\includegraphics[width=0.078\linewidth]{#1}}}
	\newcommand{\colImgR}[1]{{\color{boxred}\fbox{\colImgN{#1}}}}
	\newcommand{\colImgG}[1]{{\color{boxgreen}\fbox{\colImgN{#1}}}}
	\centering
	\begin{tabular}{cccccccccccc}
		\colImgN{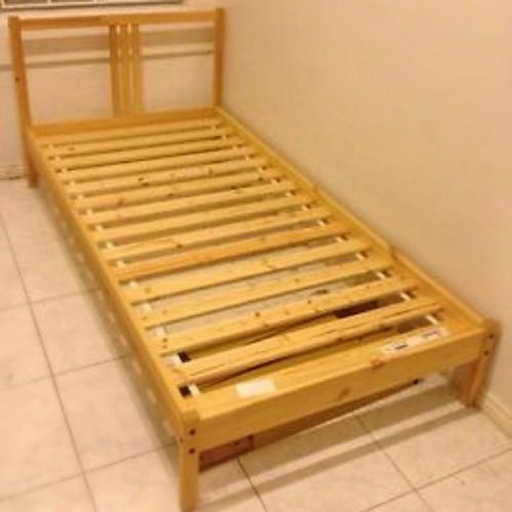}& \colImgN{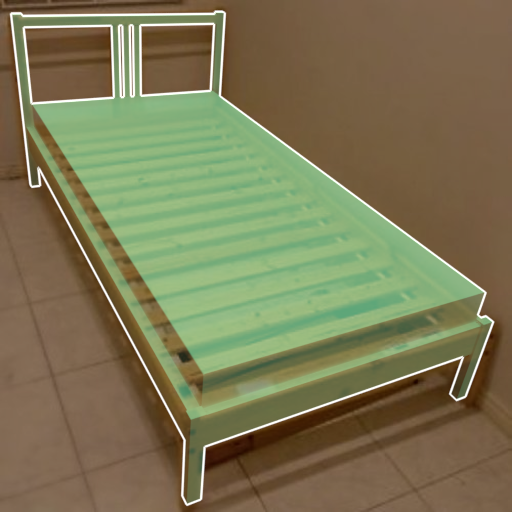}& \colImgN{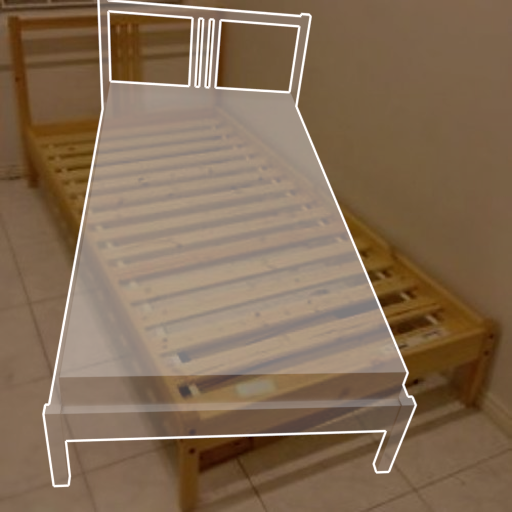}& \colImgN{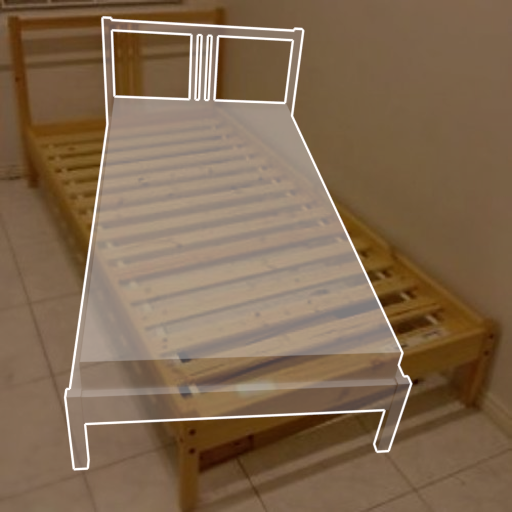}& \colImgN{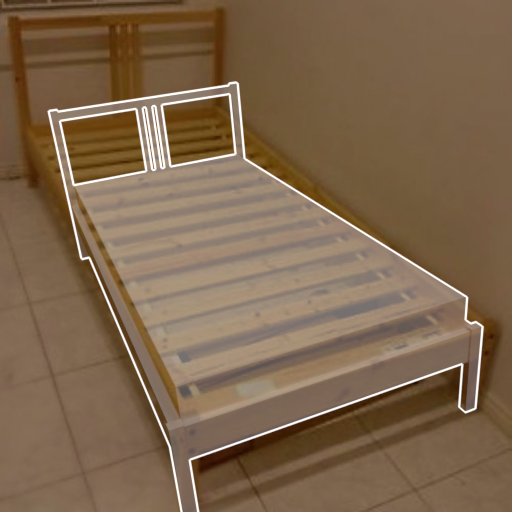}& \colImgN{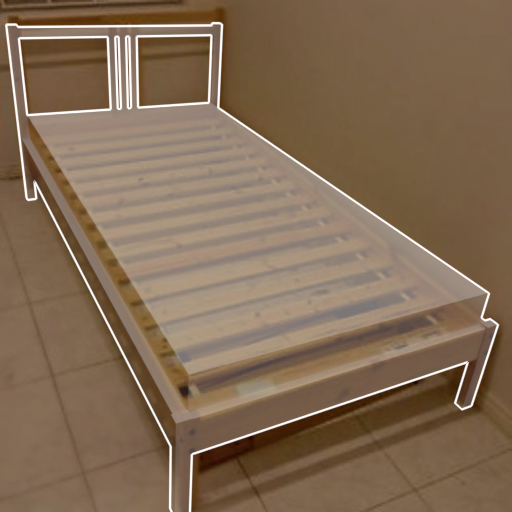}& \colImgN{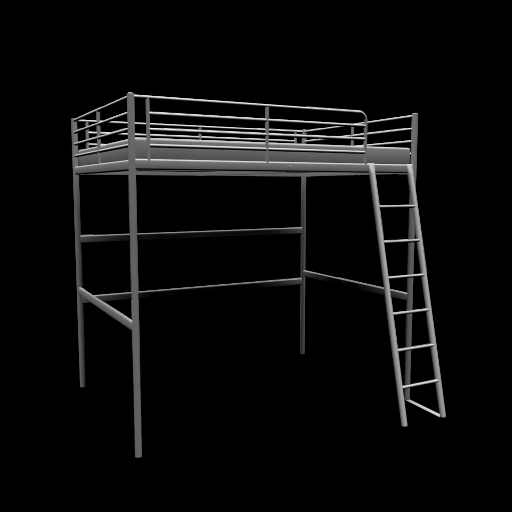}& \colImgN{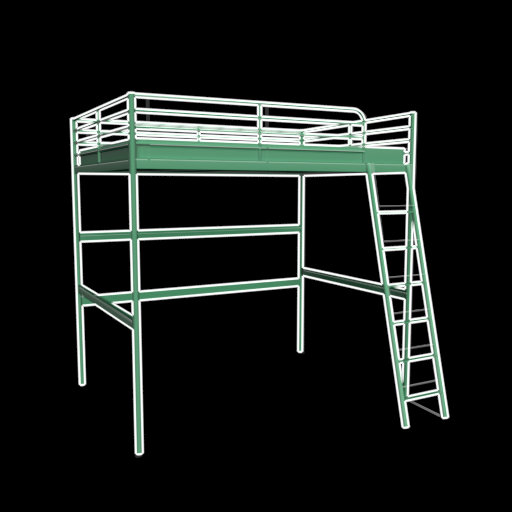}& \colImgN{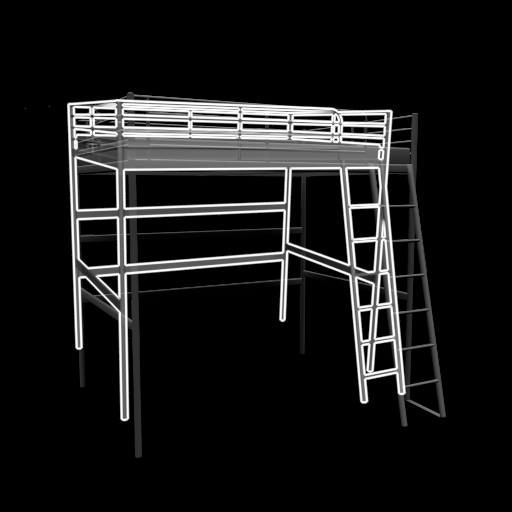}& \colImgN{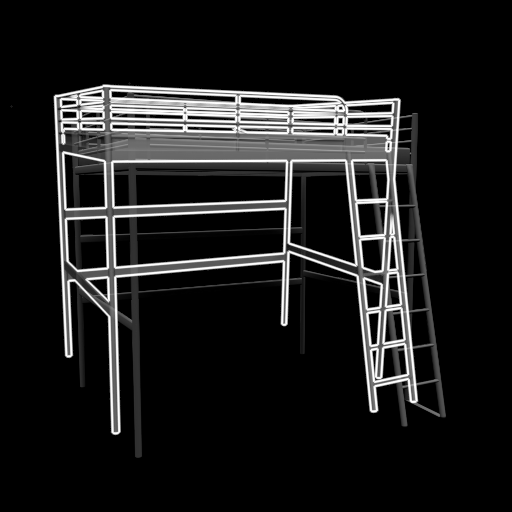}& \colImgN{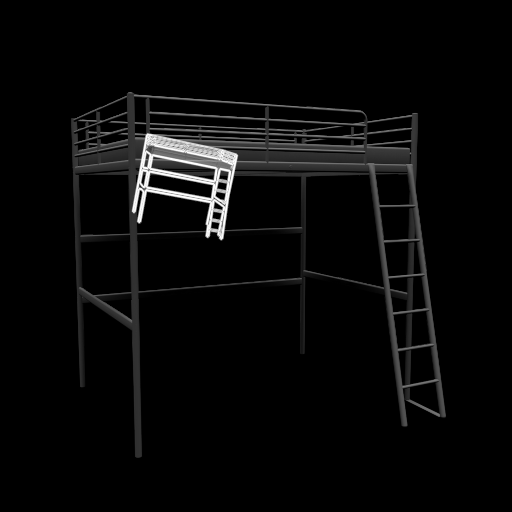}& \colImgN{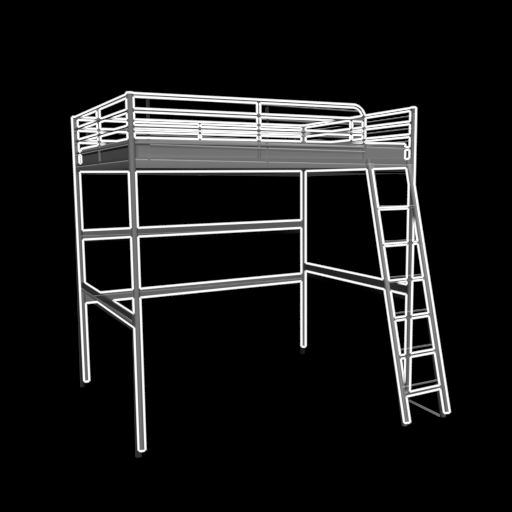}\\[-1pt]
		\colImgN{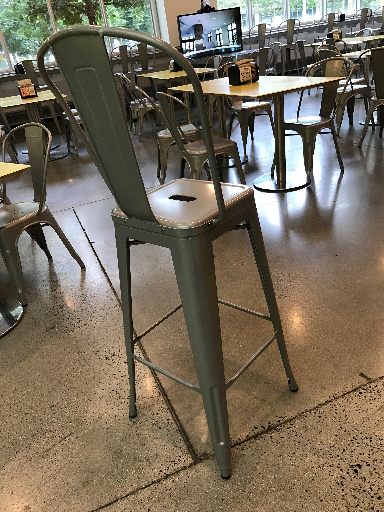}& \colImgN{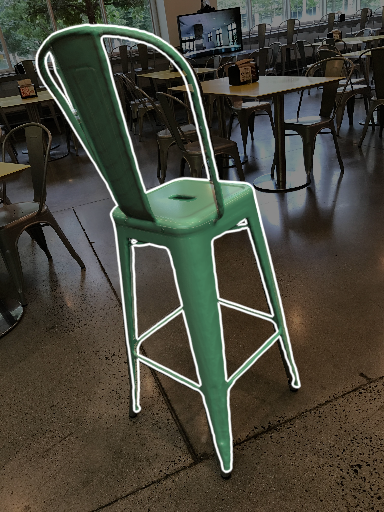}& \colImgN{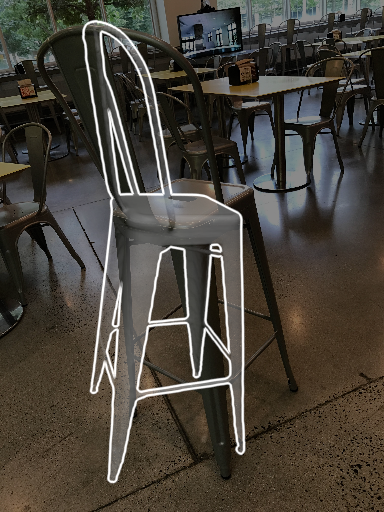}& \colImgN{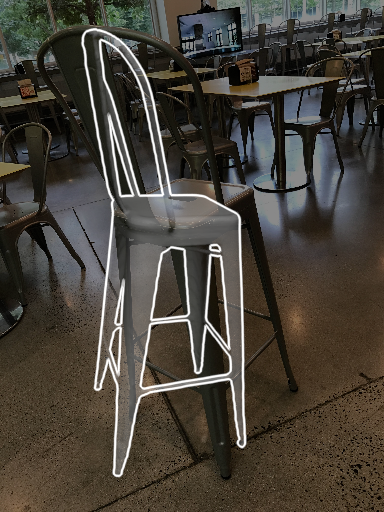}& \colImgN{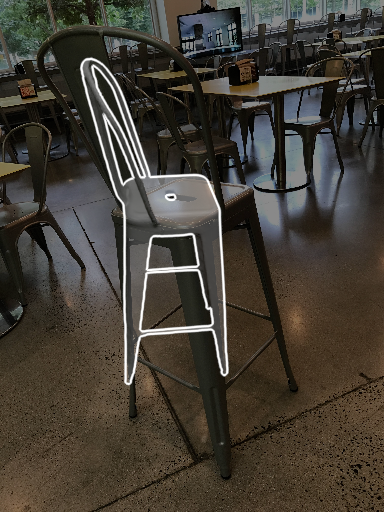}& \colImgN{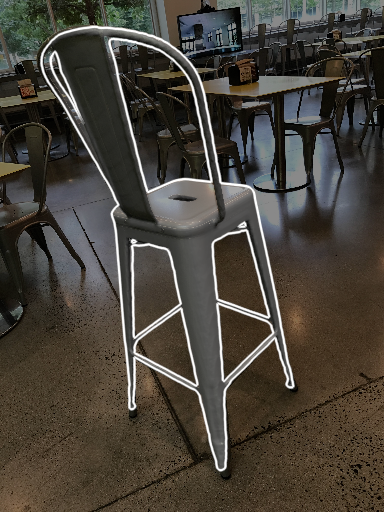}& \colImgN{}& \colImgN{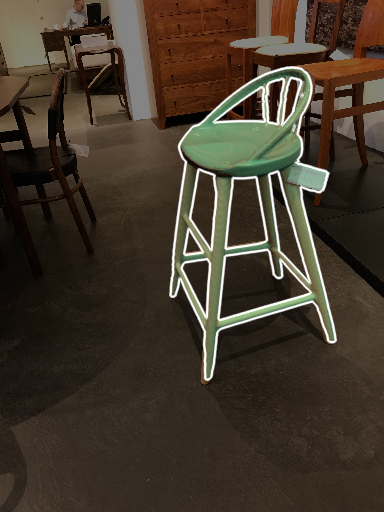}& \colImgN{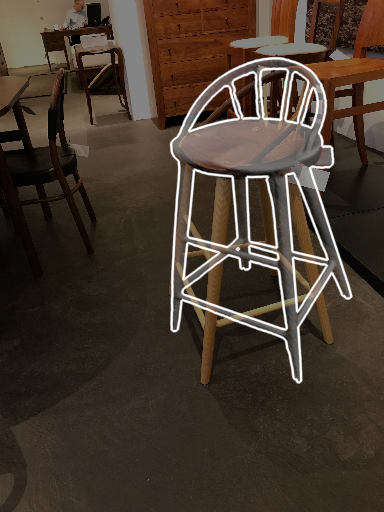}& \colImgN{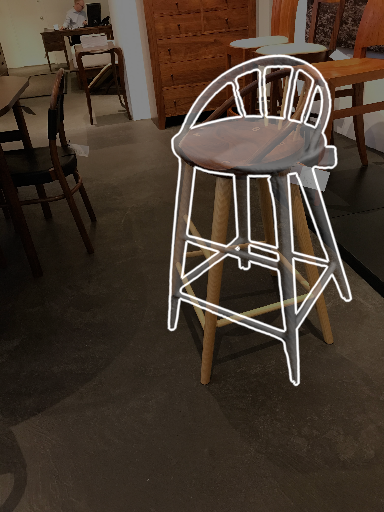}& \colImgN{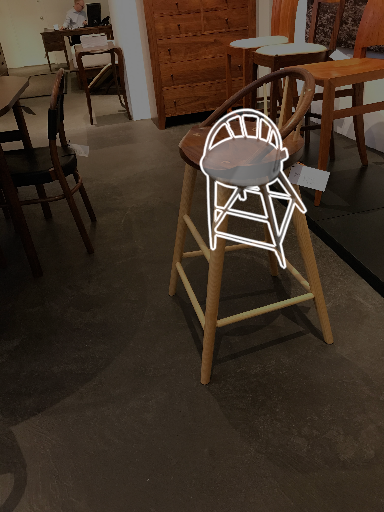}& \colImgN{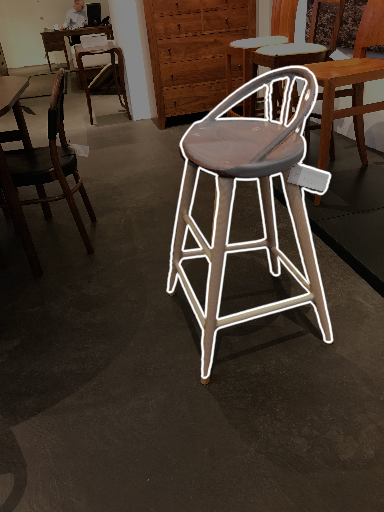}\\[-1pt]
		\colImgN{}& \colImgN{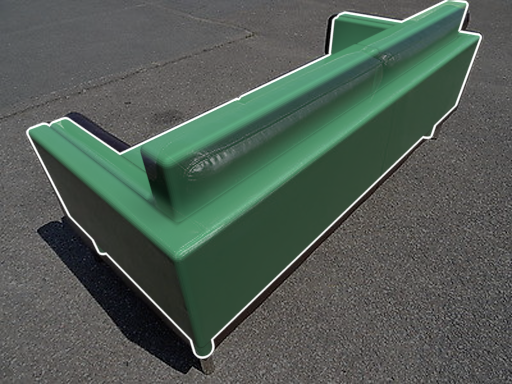}& \colImgN{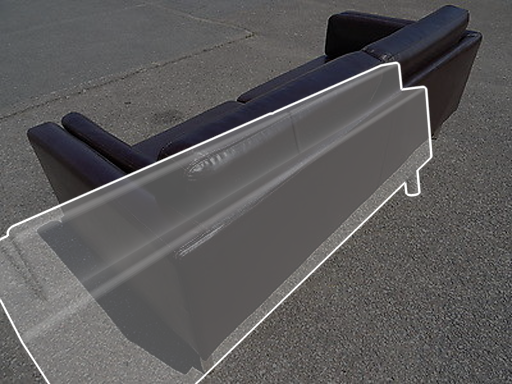}& \colImgN{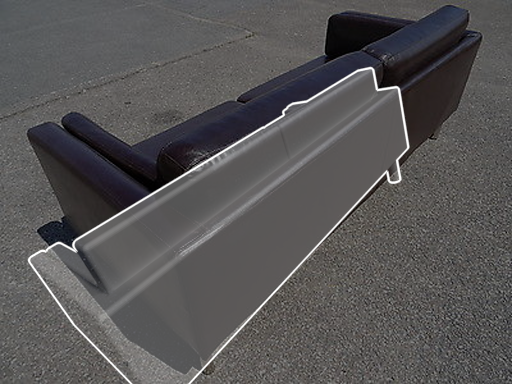}& \colImgN{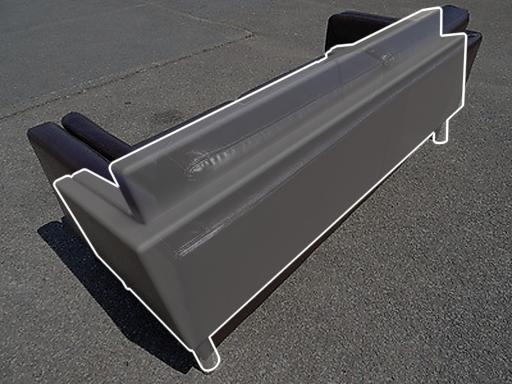}& \colImgN{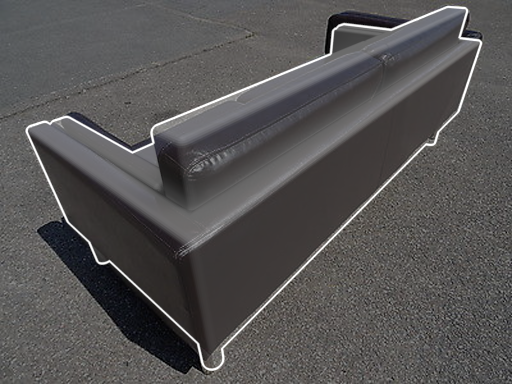}&
		\colImgR{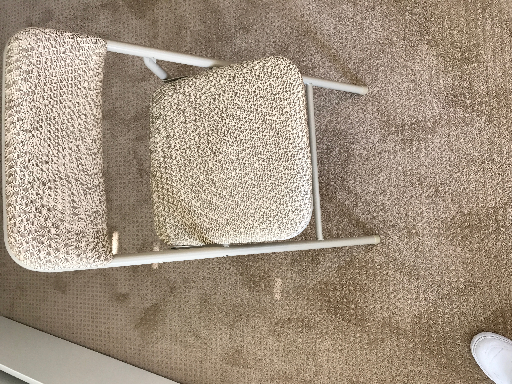}& \colImgR{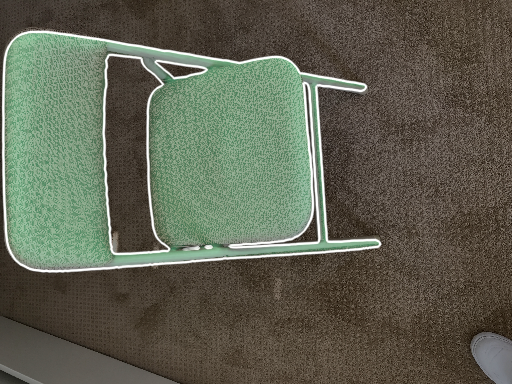}& \colImgR{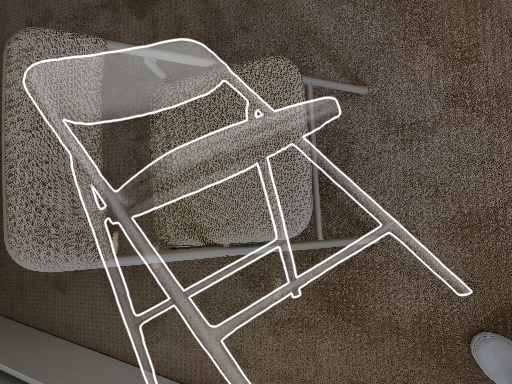}& \colImgR{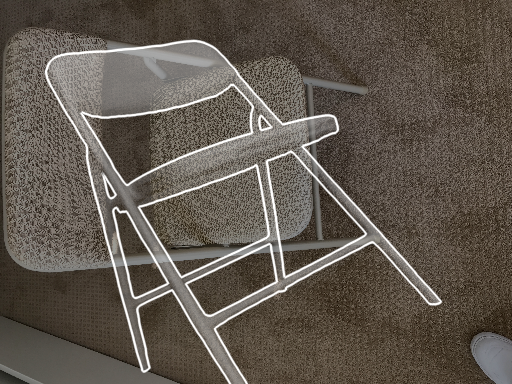}& \colImgR{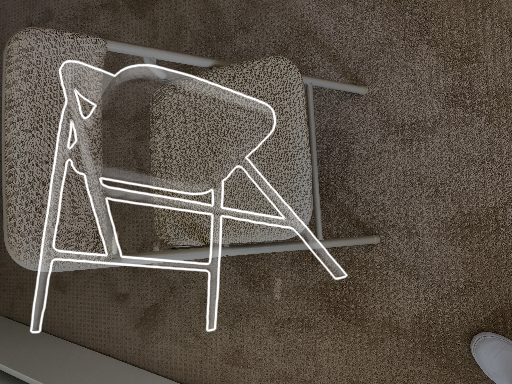}& \colImgR{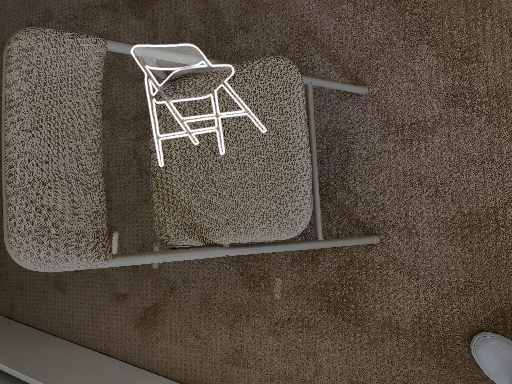}\\[-1pt]
		\scriptsize Image& \scriptsize GT&\scriptsize \cite{Grabner2019a}&\scriptsize \cite{Zakharov2019dpod} &\scriptsize \cite{Kato2018renderer}& \scriptsize Ours&\scriptsize Image& \scriptsize GT&\scriptsize \cite{Grabner2019a}&\scriptsize \cite{Zakharov2019dpod} &\scriptsize \cite{Kato2018renderer}& \scriptsize Ours\\
	\end{tabular}
	\caption{Qualitative 3D pose refinement results for objects of different categories. We project the ground truth 3D model on the image using the 3D pose estimated by different methods. Our approach overcomes the limitations of previous methods and predicts fine-grained 3D poses for objects in the wild. The last example shows a failure case (indicated by the {\color{boxred}red frame}) where the initial 3D pose is too far from the ground truth 3D pose and no refinement method can converge. More qualitative results are presented in our \textbf{supplementary material}. Best viewed in \textbf{digital zoom}.}
	\label{fig:collage}
\end{figure}

Next, \textit{Mask Refinement} improves upon \textit{Image Refinement} by a large margin across all metrics. Due to the 2D-3D alignment with differentiable rendering, this method computes more accurate 3D pose updates and additionally reduces the reprojection error ($MedErr_{P}$). However, we observe that \textit{Mask Refinement} fails in two common situations: First, when the object has holes and the mask is not a single blob the refinement fails (see Figure~\ref{fig:collage}, \eg, 1\textsuperscript{st} row - right example). In the presence of holes, the hand-crafted approximation for the rasterization backward pass accumulates gradients with alternating signs while traversing the image. This results in unreliable per-vertex motion gradients. Second, simply aligning the silhouette of objects is ambiguous as renderings under different 3D poses can have similar masks. The interior structure of the object is completely ignored. As a consequence, the refinement gets stuck in poor local minima. Finally, the performance of \textit{Mask Refinement} is limited by the quality of the target mask predicted from the RGB input image~\cite{He2017mask}.

In contrast, our refinement overcomes these limitations and significantly outperforms the baseline as well as competing refinement methods across all metrics by \textbf{up to 70\%} and \textbf{55\%} relative. Using our geometric correspondence fields, we bridge the domain gap between real-world images and synthetic renderings and align both the object outline as well as interior structures with high accuracy. 

Our approach performs especially well in the fine-grained regime, as shown in Figure~\ref{fig:curves-A}. In this experiment, we plot the 3D pose accuracy $Acc_{R,t}$ which gives the percentage of samples for which the 3D pose distance $e_{R,t}$ is smaller than a varying threshold. For strict thresholds close to zero, our approach outperforms other refinement methods by a large margin. For example, at the threshold $0.015$, we achieve more than 55\% accuracy while the runner-up \textit{Mask Refinement} achieves only 19\% accuracy.

\begin{figure}[t]
	\setlength{\tabcolsep}{1pt}
	\setlength{\fboxsep}{-2pt}
	\setlength{\fboxrule}{2pt}
	\definecolor{boxgreen}{rgb}{0.3, 1.0, 0.3}
	\definecolor{boxred}{rgb}{1.0, 0.3, 0.3}
	\newcommand{\colImgN}[1]{{\includegraphics[width=0.16\linewidth]{#1}}}
	\newcommand{\colImgR}[1]{{\color{boxred}\fbox{\colImgN{#1}}}}
	\newcommand{\colImgG}[1]{{\color{boxgreen}\fbox{\colImgN{#1}}}}
	\centering
	\begin{tabular}{cccccc}
		\colImgN{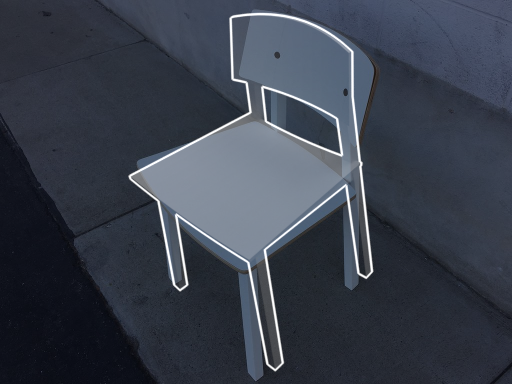}& \colImgN{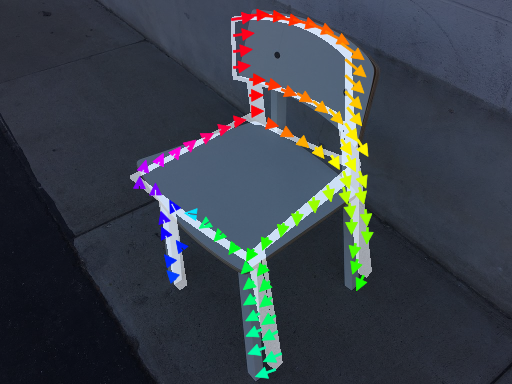}& \colImgN{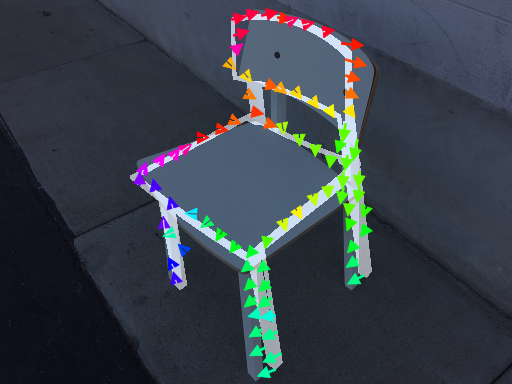}&\colImgN{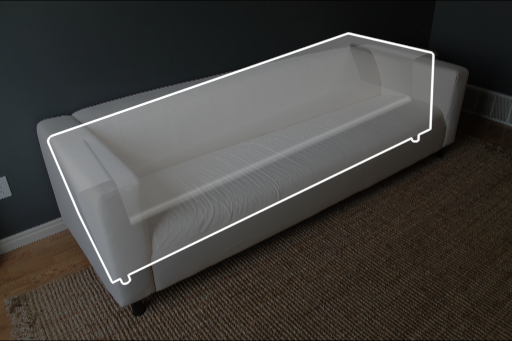}& \colImgN{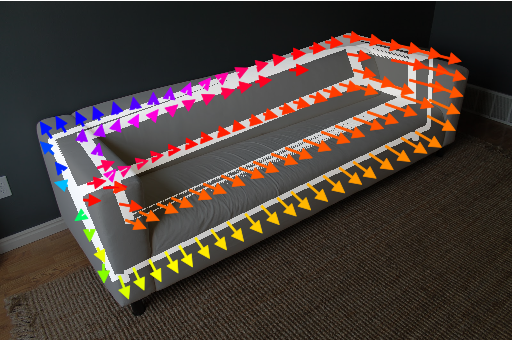}& \colImgN{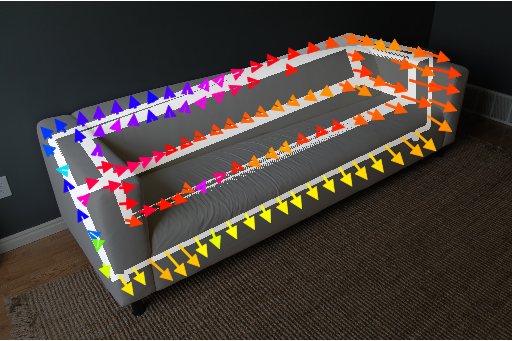}\\[-1pt]
		\scriptsize Initial 3D Pose&\scriptsize Ground Truth&\scriptsize Prediction&\scriptsize Initial 3D Pose&\scriptsize Ground Truth&\scriptsize Prediction\\
	\end{tabular}
	\caption{Qualitative examples of our predicted geometric correspondence fields. Our predicted 2D displacement vectors are highly accurate. Best viewed in \textbf{digital zoom}.}
	\label{fig:gcf}
\end{figure}

This significant performance improvement is also reflected in our qualitative examples presented in Figure~\ref{fig:collage}. Our approach precisely aligns 3D models to objects in RGB images and computes 3D poses which are in many cases visually indistinguishable from the ground truth. Even if the initial 3D pose estimate (\textit{Baseline}) is significantly off, our method can converge towards the correct 3D pose (see Figure~\ref{fig:collage}, \eg, 1\textsuperscript{st} row - left example). Finally, Figure~\ref{fig:gcf} illustrates the high quality of our predicted geometric correspondence fields.

\subsection{Ablation Study}
\label{sec:ablation}

To understand the importance of individual components in our system, we conduct an ablation study in Table~\ref{table:ablation}. For this purpose, we modify a specific system component, retrain our approach, and evaluate the performance impact.

If we use smaller kernels with less receptive field ($3\times3$ vs $7\times7$) or fewer layers (2 vs 4) in our correspondence branch, the performance drops significantly. Also, using shallow mapping branches which only employ a single Conv-BN-ReLU block to simulate simple edge and ridge features results in low accuracy because the computed features are not discriminative enough. If we perform refinement without our geometric attention mechanism, the accuracy degrades due to unreliable correspondence predictions in homogeneous regions.

Next, the choice of the rendered representation is important for the performance of our approach. While using masks only performs poorly, depth, normal, and object coordinate renderings increase the accuracy. Finally, we achieve the best accuracy by exploiting complementary information from multiple different renderings by concatenating depth, normal, and object coordinate renderings.

By inspecting failure cases, we observe that our method does not converge if the initial 3D pose is too far from the ground truth 3D pose (see Figure~\ref{fig:collage}, last example). In this case, we cannot predict accurate correspondences because our computed features are not robust to large viewpoint changes and the receptive field of our correspondence branch is limited. In addition, occlusions cause our refinement to fail because there are no explicit mechanisms to address them. We plan to resolve this issue in the future by predicting occlusion masks and correspondence confidences. However, other refinement methods also fail in these scenarios (see Figure~\ref{fig:collage}, last example).

\begin{table}[t]
	\centering
	\caption{Ablation study of our method. Using components which increase the discriminability of learned features is important for the performance of our approach. Also, our geometric attention mechanism and the chosen type of rendering effect the accuracy.}
	\label{table:ablation}
	\resizebox{0.9\columnwidth}{!}{%
		\setlength{\tabcolsep}{4pt}
		\begin{tabular}{lcccc}
			\toprule
			\multicolumn{1}{c}{}&\multicolumn{1}{c}{\bf Rotation}&\multicolumn{1}{c}{\bf Translation}&\multicolumn{1}{c}{\bf Pose}&\multicolumn{1}{c}{\bf Projection}\\
			\cmidrule(lr){2-2}\cmidrule(lr){3-3}\cmidrule(lr){4-4}\cmidrule(lr){5-5}
			\multicolumn{1}{l}{\multirow{2}{*}{Method}}&\multicolumn{1}{c}{$MedErr_R$}&\multicolumn{1}{c}{$MedErr_{t}$}&\multicolumn{1}{c}{$MedErr_{R,t}$}&\multicolumn{1}{c}{$MedErr_{P}$}\\
			\multicolumn{1}{c}{}&\multicolumn{1}{c}{$\cdot1$}&\multicolumn{1}{c}{$\cdot10^{2}$}&\multicolumn{1}{c}{$\cdot10^{2}$}&\multicolumn{1}{c}{$\cdot10^{2}$}\\
			\midrule
			Ours less receptive&3.01&2.12&1.75&1.56\\
			Ours fewer layers&2.84&1.98&1.58&1.41\\
			Ours shallow features&2.76&2.00&1.55&1.42\\
			Ours without attention&2.70&1.92&1.45&1.37\\
			Ours only MASK&2.98&1.85&1.44&1.41\\
			Ours only OBJ COORD&2.57&1.80&1.39&1.31\\
			Ours only DEPTH&2.60&1.77&1.38&1.29\\
			Ours only NORMAL&2.58&1.76&1.36&1.30\\
			Ours&\bf2.56&\bf1.74&\bf1.34&\bf1.27\\
			\bottomrule
	\end{tabular}}
\end{table}

\subsection{3D Model Retrieval}
\label{sec:retrieval}

So far, we assumed that the ground truth 3D model required for 3D pose refinement is given at runtime. However, we can overcome this limitation by automatically retrieving 3D models from single RGB images. For this purpose, we combine all refinement approaches with the retrieval method presented in~\cite{Grabner2019b}, where the 3D model database essentially becomes a part of the trained model. In this way, we perform initial 3D pose estimation, 3D model retrieval, and 3D pose refinement given only a single RGB image. This setting allows us to benchmark refinement methods against feed-forward baselines in a fair comparison.

The corresponding results are presented in Table~\ref{table:pix3d} ({\setlength\fboxsep{0.5pt}\colorbox{light_blue}{lower part}}) and Figure~\ref{fig:curves-B}. Because the retrieved 3D models often differ from the ground truth 3D models, the refinement performance decreases compared to given ground truth 3D models. Differentiable rendering methods lose more accuracy than traditional refinement methods because they require 3D models with accurate geometry. 

Still, all refinement approaches perform remarkably well with retrieved 3D models. As long as the retrieved 3D model is reasonably close to the ground truth 3D model in terms of geometry, our refinement succeeds. Our method achieves even lower 3D pose error ($MedErr_{R,t}$) with retrieved 3D models than \textit{Mask Refinement} with ground truth 3D models. Finally, using our joint 3D pose estimation-retrieval-refinement pipeline, we reduce the 3D pose error ($MedErr_{R,t}$) compared to the state of the art for single image 3D pose estimation on Pix3D (\textit{Baseline}) \textbf{by 55\%} relative \textbf{without using additional inputs}.

\section{Conclusion}

Aligning 3D models to objects in RGB images is the most accurate way to predict 3D poses. However, there is a domain gap between real-world images and synthetic renderings which makes this alignment challenging in practice. To address this problem, we predict deep cross-domain correspondences in a feature space optimized for 3D pose refinement and combine local 2D displacement vectors into global 3D pose updates using our novel differentiable renderer. Our method outperforms existing refinement approaches by up to 55\% relative and can be combined with feed-forward 3D pose estimation and 3D model retrieval to predict fine-grained 3D poses for objects in the wild given only a single RGB image. Finally, our novel learned differentiable rendering framework can be used for other tasks in the future.


\clearpage
%
%
\bibliographystyle{splncs04}
\bibliography{string,references}

\clearpage


 \begin{center}%
	\let\newline\\
	{\Large \bfseries\boldmath
		\pretolerance=10000
		Geometric Correspondence Fields:\\Learned Differentiable Rendering for\\3D Pose Refinement in the Wild\\--\\Supplementary Material \par}\vskip .8cm
	\if!\@subtitle!\else {\large \bfseries\boldmath
		\vskip -.65cm
		\pretolerance=10000
		\par}\vskip .8cm\fi
	{\lineskip .5em
		\noindent\ignorespaces
		Alexander Grabner\textsuperscript{1,2}, Yaming Wang\textsuperscript{2}, Peizhao Zhang\textsuperscript{2}, Peihong Guo\textsuperscript{2},\\ Tong Xiao\textsuperscript{2}, Peter Vajda\textsuperscript{2}, Peter M. Roth\textsuperscript{1}, and Vincent Lepetit\textsuperscript{1}\vskip.35cm}
	{\small \textsuperscript{1} Graz University of Technology, Austria \qquad \textsuperscript{2} Facebook Inc., USA\\ \email{\{alexander.grabner, pmroth, lepetit\}@icg.tugraz.at}\\ \email{\{wym, stzpz, phg, xiaot, vajdap\}@fb.com}}
\end{center}%


In the following, we provide additional details and experimental results of our novel 3D pose refinement approach. In Sec.~\ref{sec:datasets}, we present our evaluation setup and discuss different datasets. In Sec.~\ref{sec:metrics}, we formally describe our evaluated metrics. In Sec.~\ref{sec:implementation}, we give specific details on the implementation of our approach. In Sec.~\ref{sec:iterative}, we analyze the iterative refinement behavior of different methods. In Sec.~\ref{sec:detailed_results}, we present detailed quantitative 3D pose refinement results for individual object categories. In Sec.~\ref{sec:add-quali}, we show additional qualitative results. Finally, we analyze failure cases of our approach in Sec.~\ref{sec:add-fail}. 

\section{Datasets and Evaluation Setup}
\label{sec:datasets}

We evaluate our proposed 3D pose refinement approach on the challenging Pix3D~\cite{Sun2018pix3d} dataset. The Pix3D dataset provides in-the-wild RGB images with 3D pose, 3D model, and focal length annotations for objects of different categories. We follow the evaluation protocol of previous work~\cite{Grabner2019a} and perform experiments on categories which have more than 300 non-occluded and non-truncated samples (\textit{bed}, \textit{chair}, \textit{sofa}, \textit{table}). Further, we restrict the training and evaluation to samples marked as non-occluded and non-truncated because all evaluated refinement methods lack explicit mechanisms to deal with occlusions. In addition, this dataset does not provide information on which objects parts are occluded nor the extent of the occlusion. For each category, we use 50\% of the samples for training and the other 50\% for testing as in~\cite{Grabner2019a}.

Other category-level datasets do not provide annotations with sufficient accuracy to both train and evaluate fine-grained 3D pose refinement methods. For example, the Comp~\cite{Wang2018fine} and Stanford~\cite{Wang2018fine} datasets only provide coarse 3D pose annotations. In addition to coarse 3D pose annotations, the ScanNet~\cite{Dai2017scannet}, Pascal3D+~\cite{Xiang2014beyond}, and ObjectNet3D~\cite{Xiang2016objectnet3d} datasets also just provide approximate 3D model annotations. Moreover, the latter two datasets assume constant camera intrinsics for images captured with different cameras which further decreases the annotation quality~\cite{Grabner2019a}. As a consequence of this label noise, the training of refinement methods results in models with poor accuracy, while quantitative evaluations are not representative of the true refinement performance due to the lack of accuracy in the annotations.

In contrast, instance-level datasets like LineMOD~\cite{Hinterstoisser2011gradient}, YCB~\cite{Calli2015ycb}, T-LESS~\cite{Hodavn2017tless}, or NOCS~\cite{Wang2019normalized} provide accurate annotations but have many images with strong occlusions. Neither traditional refinement methods~\cite{Li2018deepim,Manhardt2018deep,Zakharov2019dpod}, nor differentiable rendering based refinement methods~\cite{Kato2018renderer,Loper2014opendr,Palazzi2018end}, nor our approach employ explicit mechanisms to deal with occlusions. For example, one issue across all evaluated methods is that they align renderings to real-world images but occlusions are only present in the real-world images while the renderings are always un-occluded. Also, simply training methods based on feed-forward CNNs on occluded objects is in practice not sufficient to handle occlusions~\cite{Oberweger2018making}. However, we specifically plan to address occlusions in the future by predicting occlusion masks and correspondence confidences.

In addition, our approach is specifically designed for category-level 3D pose refinement using untextured 3D models. This task is very different from instance-level 3D pose estimation where exactly matching colored and textured 3D models are available. In this case, methods which leverage color and texture information of 3D models have a clear advantage and should be used instead.

For these two reasons, we did not evaluate our method on instance-level datasets like LineMOD or YCB which have many images with strong occlusions and provide 3D models with colors and textures that exactly match those of the objects in the RGB images.

\section{Metrics}
\label{sec:metrics}

We follow the evaluation protocol of previous work~\cite{Grabner2019a} and
report the median error ($MedErr$) of multiple geometric distances:

\subsubsection*{Rotation:} The 3D rotation distance 
\begin{equation}
e_R = \frac{\Vert \text{log}(\mathbf{R}_\gt^T \mathbf{R}_\pred^{\vphantom{T}} )\Vert_F}{\sqrt{2}}
\end{equation}
represents the minimal angle between the ground truth rotation matrix $\mathbf{R}_\gt$ and the predicted rotation matrix $\mathbf{R}_\pred$~\cite{Tulsiani2015viewpoints}.

\subsubsection*{Translation:} The 3D translation distance
\begin{equation}
e_t = \frac{\Vert \mathbf{t}_\gt - \mathbf{t}_\pred \Vert_2}{\Vert \mathbf{t}_\gt \Vert_2}
\end{equation}
gives the relative error between the ground truth translation $\mathbf{t}_\gt$ and the predicted translation $\mathbf{t}_\pred$~\cite{Hodavn2016evaluation}. 

\subsubsection*{Pose:} The 3D pose distance 
\begin{equation}
e_{R,t} = \underset{\mathbf{M}_i \in \mathcal{M}}{\text{avg}} \frac{d_{\text{bbox}}}{d_{\text{img}}} \cdot \frac{\Vert \text{transf}(\mathbf{M}_i, \mathcal{P}_\gt) - \text{transf}(\mathbf{M}_i, \mathcal{P}_\pred) \Vert_2}{\Vert \mathbf{t}_\gt \Vert_2} 
\end{equation}
represents the average normalized Euclidean distance of all transformed 3D model points in 3D space~\cite{Hinterstoisser2012model,Hodavn2016evaluation}. Each 3D point $\mathbf{M}_i$ of the ground truth 3D model $\mathcal{M}$ is transformed using the ground truth 3D pose $\mathcal{P}_\gt$ and the predicted 3D pose $\mathcal{P}_\pred$. This distance is normalized by the relative size of the object in the image using the ratio between the ground truth 2D bounding box diagonal $d_{\text{bbox}}$ and the image diagonal $d_{\text{img}}$, and the L2-norm of the ground truth translation $\Vert \mathbf{t}_\gt \Vert_2$.

\subsubsection*{Projection:} The 2D projection distance
\begin{equation}
e_{P} = \underset{\mathbf{M}_i \in \mathcal{M}}{\text{avg}} \frac{\Vert \text{proj}(\mathbf{M}_i, \mathcal{P}_\gt) - \text{proj}(\mathbf{M}_i, \mathcal{P}_\pred) \Vert_2}{d_{\text{bbox}}} \>
\end{equation}
is the average reprojection error normalized by the ground truth 2D bounding box diagonal $d_{\text{bbox}}$~\cite{Wang2018fine}. In this case, each 3D point $\mathbf{M}_i$ of the ground truth 3D model $\mathcal{M}$ is projected to the 2D image plane using the ground truth 3D pose $\mathcal{P}_\gt$ and the predicted 3D pose $\mathcal{P}_\pred$ subject to a camera model. In this work, we assume the camera intrinsics to be known.

\section{Implementation and Training Details}
\label{sec:implementation}

To train our refinement network, we resize and pad images to a spatial resolution of $256\times256$ while maintaining the aspect ratio. In this way, we are able to combine images with different aspect ratios in the same training batch. For our mapping branches, we adapt a ResNet-50~\cite{He2016deep,He2016identity} architecture. We utilize all layers up to the end of the first stage but use a stride of 1 for all convolutional layers and discard max pooling layers. For our correspondence branch, we use a channel dimensionality of 64 for all convolutional layers except the output layer.

During training of our network, we additionally employ different forms of data augmentation like mirroring, affine transformations, and independent pixel augmentations like additive or multiplicative noise for the RGB image $I$.

To regularize our network, we use L2 weight decay with a factor of $1e^{-5}$. We train our network $f(\cdot)$ for 1500 epochs using the Adam optimizer~\cite{Kingma2014adam} with an initial learning rate of $\eta=1e^{-3}$. We use a batch size of 8 and decrease the learning rate by a factor of 5 after 1000 and 1400 epochs.

During inference of our system, we compute geometry-level gradients $\nabla \mathbf{m}_i$ from our predicted geometric correspondence fields. In this way, vertices belonging to self-occluded triangles~\cite{Kato2018renderer} or to visible triangles which are masked out by our geometric attention module do not receive gradients. However, since the geometry is fixed, providing gradients for a subset of all vertices is sufficient to perform 3D pose refinement.

\section{Iterative Refinement}
\label{sec:iterative}

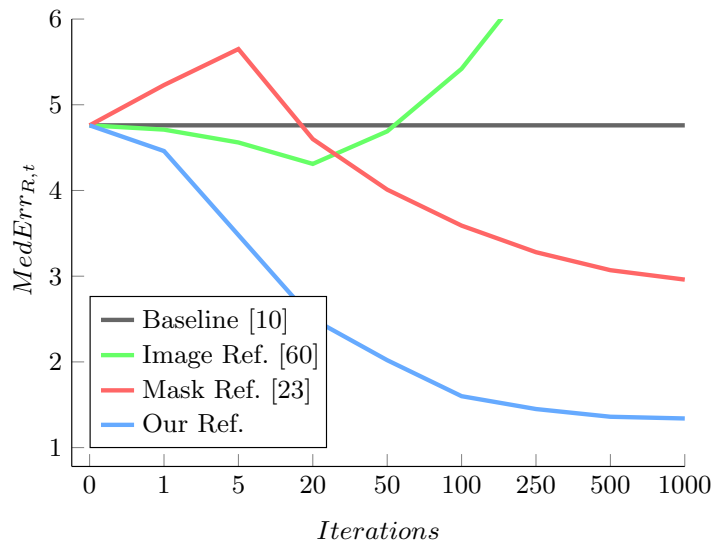
\begin{figure}
	\centering
	\resizebox{0.8\columnwidth}{!}{%
\pgfplotsset{
	compat=1.11,
	legend image code/.code={
		\draw[mark repeat=1,mark phase=1]
		plot coordinates {
			(0cm,0cm)
			(0.4cm,0cm) 
		};%
	}
}
\begin{tikzpicture}[
scale=1,
]
\begin{axis}[
width = 9cm,
height = 7cm,
ytick={1,2,3,4,5,6},
yticklabels={1,2,3,4,5,6},
xtick={0,0.125,0.25,0.375,0.5,0.625,0.75,0.875,1},
xticklabels={0,1,5,20,50,100,250,500,1000},
ymin=0.78,
xmin=-0.03,
ymax=6,
xmax=1,
xlabel=$Iterations$,
ylabel=$MedErr_{R,t}$,
axis x line*=bottom, 
axis y line*=left, 
x label style={at={(axis description cs:0.5,-0.1)},anchor=north},
y label style={at={(axis description cs:-0.04,.53)},anchor=south},
legend style={at={(0.029,0.041)},anchor=south west,font=\footnotesize},
legend cell align={left},
every axis plot/.append style={line width=1.5pt}]

\definecolor{alex_baseline}{HTML}{666666}
\definecolor{alex_image}{HTML}{66FF66}
\definecolor{alex_mask}{HTML}{FF6666}
\definecolor{alex_ours}{HTML}{66AAFF}

\addplot[alex_baseline] plot coordinates {
(0.0,4.76)
(1.0,4.76)
};
\addlegendentry{Baseline~\cite{Grabner2019a}}

\addplot[alex_image] plot coordinates {
	(0.0,4.76) 
	(0.125,4.71) 
	(0.25,4.56) 
	(0.375,4.31) 
	(0.5,4.69) 
	(0.625,5.42) 
	(0.75,6.52) 
	(0.875,7.08) 
	(1.0,7.37) 
};
\addlegendentry{Image Ref.~\cite{Zakharov2019dpod}\hspace*{-1cm}}

\addplot[alex_mask] plot coordinates {
(0.0,4.76) 
(0.125,5.23) 
(0.25,5.65) 
(0.375,4.60) 
(0.5,4.01) 
(0.625,3.59) 
(0.75,3.28) 
(0.875,3.07) 
(1.0,2.96) 
};
\addlegendentry{Mask Ref.~\cite{Kato2018renderer}}

\addplot[alex_ours] plot coordinates {
(0.0,4.76) 
(0.125,4.46) 
(0.25,3.48) 
(0.375,2.5) 
(0.5,2.02) 
(0.625,1.6) 
(0.75,1.45) 
(0.875,1.36) 
(1.0,1.34) 
};
\addlegendentry{Our Ref.}

\end{axis}
\end{tikzpicture}}
	\caption{Evaluation on 3D pose refinement after varying numbers of iterations. In contrast to other methods, our refinement achieves a consistent improvement upon the baseline, increasing with the number of iterations.}
	\label{fig:iterations}
\end{figure}

Figure~\ref{fig:iterations} shows the performance of different refinement methods after varying numbers of iterations. In this experiment, we report the 3D pose metric $MedErr_{R,t}$ as a function of the number of 3D pose updates. The baseline does not perform iterative updates, thus, its $MedErr_{R,t}$ score is constant.

For \textit{Image Refinement}~\cite{Zakharov2019dpod}, the accuracy increases until 20 iterations but then starts to decrease. After the first couple of coarse refinement steps, the predicted updates are not accurate enough to refine the 3D pose but start to jitter without further improving the 3D pose. Moreover, for many objects the prediction fails and the iterative updates cause the 3D pose to drift off which results in high $MedErr_{R,t}$ for large numbers of iterations. 

For \textit{Mask Refinement}~\cite{Kato2018renderer}, we observe an opposite effect. In the beginning, the accuracy decreases but then the performance increases. This is due to degenerated masks predicted from RGB images by Mask R-CNN~\cite{He2017mask}. The mask prediction often fails to capture fine-grained and thin structures, \eg, ornaments and legs of a bed (see Figure~\ref{fig:mask_fail}). These degenerated masks cause large gradients during refinement and quickly pull the 3D pose away from the reasonable initial estimate predicted by the baseline (also see Figure~\ref{fig:collage2}, 2\textsuperscript{nd} row). After five iterations the refinement on samples with correctly predicted masks counteracts this effect and the performance improves.

\begin{table}[t]
	\centering
	\setlength{\tabcolsep}{10pt}
	\caption{Average computation times of different refinement methods for a single iteration using a Titan X GPU. For all evaluated methods, the execution time for computing a single 3D pose update is within the same order of magnitude.}
	\label{table:timings}
	\begin{tabular}{lc}
		\toprule
		Method&Time per Iteration\\
		\midrule
		Image Refinement~\cite{Zakharov2019dpod}&86.5 ms\\
		Mask Refinement~\cite{Kato2018renderer}&29.7 ms\\
		Our Refinement&36.0 ms\\
		\bottomrule
	\end{tabular}
\end{table}

\begin{figure}
	\setlength{\tabcolsep}{1pt}
	\setlength{\fboxsep}{-2pt}
	\setlength{\fboxrule}{2pt}
	\definecolor{boxgreen}{rgb}{0.3, 1.0, 0.3}
	\definecolor{boxred}{rgb}{1.0, 0.3, 0.3}
	\newcommand{\colImgN}[1]{{\includegraphics[width=0.315\linewidth]{#1}}}
	\newcommand{\colImgR}[1]{{\color{boxred}\fbox{\colImgN{#1}}}}
	\newcommand{\colImgG}[1]{{\color{boxgreen}\fbox{\colImgN{#1}}}}
	\centering
	\begin{tabular}{ccc}
		\colImgN{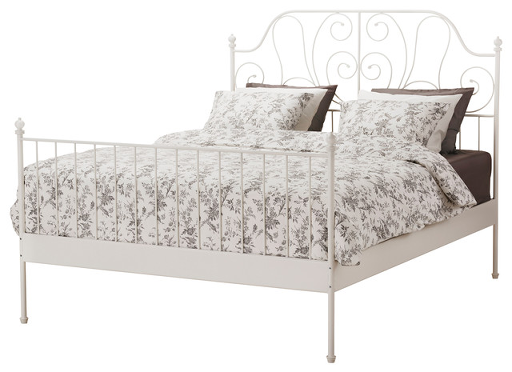}& \colImgN{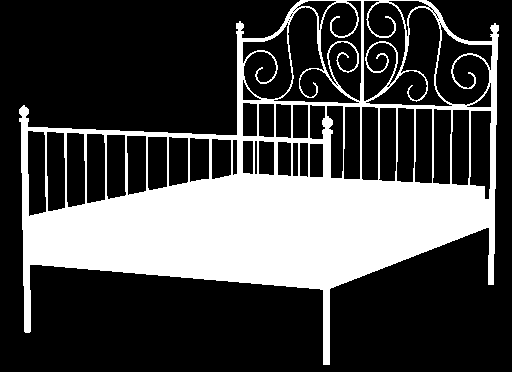}& \colImgN{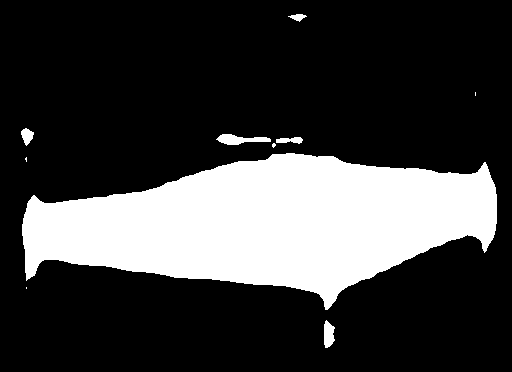}\\
		\footnotesize Image&\footnotesize GT Mask&\footnotesize Predicted Mask\\[3pt]
		\colImgN{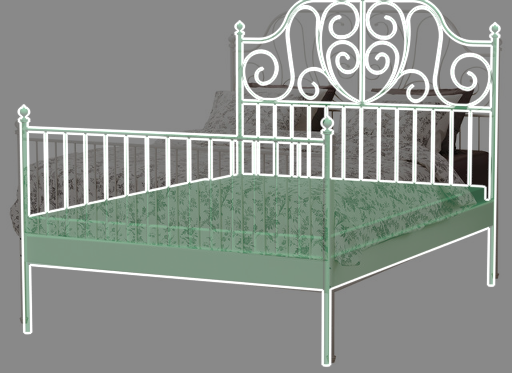}& \colImgN{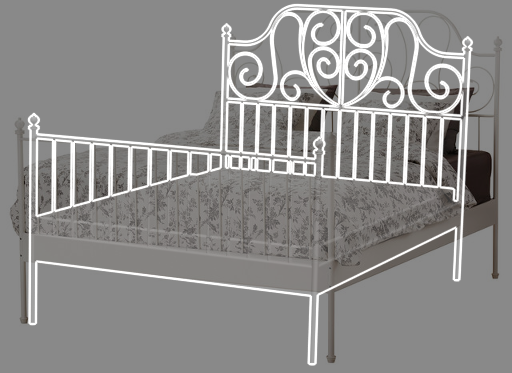}& \colImgN{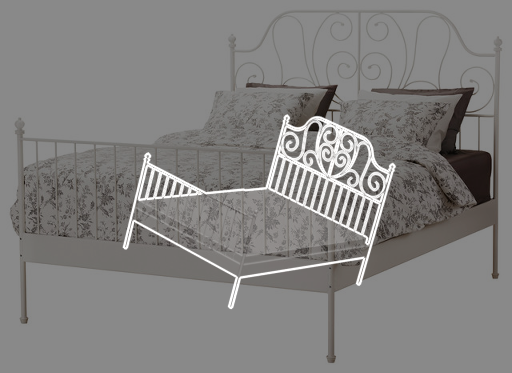}\\
		\footnotesize GT 3D Pose&\footnotesize Baseline~\cite{Grabner2019a} 3D Pose&\footnotesize Mask Ref.~\cite{Kato2018renderer} 3D Pose\\
	\end{tabular}
	\caption{Typical failure case of \textit{Mask Refinement}~\cite{Kato2018renderer}. Degenerated predicted masks (\textit{top right}) cause large gradients during refinement and quickly pull the 3D pose away from reasonable initial estimates by the baseline (\textit{bottom middle}). As a consequence, the 3D pose refinement using \textit{Mask Refinement} fails (\textit{bottom right)}. Also see Figure~\ref{fig:collage2}, 2\textsuperscript{nd} row.}
	\label{fig:mask_fail}
\end{figure}

In contrast to other refinement methods, our approach achieves a consistent improvement upon the baseline, increasing with the number of iterations. As expected, the accuracy saturates for large numbers of iterations. Empirically, we achieve maximum accuracy by performing 1000 3D pose updates using the Adam optimizer~\cite{Kingma2014adam}  with a learning rate of $\eta=0.05$. 


Finally, Table~\ref{table:timings} compares the computation times of different refinement methods. \textit{Image Refinement} evaluates two large ResNet-style networks~\cite{He2016deep} during inference and, thus, this method is the slowest in our evaluation. \textit{Mask Refinement} generates a mask rendering, computes a loss in the mask space, and performs the backward pass of its differentiable renderer in each iteration. This method is the fastest in our evaluation since the target mask predicted from the input RGB image by Mask R-CNN~\cite{He2017mask} does not change during refinement. It is only predicted once before the iterative process and the inference time of Mask R-CNN is not considered in this experiment.

Our refinement is only marginally slower than \textit{Mask Refinement} as we just evaluate our efficient network branches in addition to the forward and backward pass of our differentiable renderer in each iteration. However, all evaluated refinement methods show comparable execution time within the same order of magnitude for computing a single 3D pose update. 

\section{Detailed Quantitative Results}
\label{sec:detailed_results}

Tables~\ref{table:pix3d-gt} (\textit{GT 3D models}) and~\ref{table:pix3d-retrieval} (\textit{retrieved 3D models}) show detailed quantitative results for individual object categories on Pix3D. For completeness, we additionally report the detection accuracy $Acc_{D_{0.5}}$ which gives the percentage of objects for which the intersection over union between the ground truth 2D bounding box and the predicted 2D bounding box is larger than 50\%~\cite{Xiang2014beyond}. We do not make 3D pose predictions for objects which are not detected by the baseline~\cite{Grabner2019a}. However, the reported $MedErr$ metrics are computed over all samples, both detected and not detected.

Our refinement outperforms the baseline as well as competing refinement methods across all metrics. In fact, we do not only increase the \textit{mean} performance over all categories but also achieve state-of-the-art results for each individual category. Using both ground truth (see Table~\ref{table:pix3d-gt}) and retrieved (see Table~\ref{table:pix3d-retrieval}) 3D models, we improve the performance compared to other methods by a large margin for each evaluated category.

\begin{table*}
	\centering
	\setlength{\tabcolsep}{10pt}
	\caption{Detailed 3D pose refinement results for individual categories on Pix3D. In this experiment, we provide the \textbf{ground truth 3D model} for refinement. We outperform existing methods across all categories by a large margin.}
	\label{table:pix3d-gt}
	\resizebox{\columnwidth}{!}{%
		\begin{tabular}{lc|c|c|c|c|c}
			\toprule
			\multicolumn{2}{c}{}&\multicolumn{1}{c}{\bf Detection}&\multicolumn{1}{c}{\bf Rotation}&\multicolumn{1}{c}{\bf Translation}&\multicolumn{1}{c}{\bf Pose}&\multicolumn{1}{c}{\bf Projection}\\
			\cmidrule(lr){3-3}\cmidrule(lr){4-4}\cmidrule(lr){5-5}\cmidrule(lr){6-6}\cmidrule(lr){7-7}
			\multirow{2}{*}{Method}&\multicolumn{1}{c}{\multirow{2}{*}{Category}}&\multicolumn{1}{c}{\multirow{2}{*}{$Acc_{D_{0.5}}$}}&\multicolumn{1}{c}{$MedErr_R$}&\multicolumn{1}{c}{$MedErr_{t}$}&\multicolumn{1}{c}{$MedErr_{R,t}$}&\multicolumn{1}{c}{$MedErr_{P}$}\\
			&\multicolumn{1}{c}{}&\multicolumn{1}{c}{}&\multicolumn{1}{c}{$\cdot1$}&\multicolumn{1}{c}{$\cdot10^{2}$}&\multicolumn{1}{c}{$\cdot10^{2}$}&\multicolumn{1}{c}{$\cdot10^{2}$}\\
			\midrule
			Baseline~\cite{Grabner2019a}&\multirow{4}{*}{bed}&\multirow{4}{*}{99.0}&5.07&6.68&5.18&3.42\\
			Image Refinement~\cite{Zakharov2019dpod}&&&4.65&5.45&4.60&3.38\\
			Mask Refinement~\cite{Kato2018renderer}&&&3.03&4.04&3.07&2.00\\
			Our Refinement&&&\bf2.40&\bf1.84&\bf1.45&\bf1.28\\
			\midrule
			Baseline~\cite{Grabner2019a}&\multirow{4}{*}{chair}&\multirow{4}{*}{95.2}&7.36&5.49&3.90&3.32\\
			Image Refinement~\cite{Zakharov2019dpod}&&&7.10&5.31&3.68&3.34\\
			Mask Refinement~\cite{Kato2018renderer}&&&4.42&4.89&3.17&1.79\\
			Our Refinement&&&\bf2.96&\bf1.77&\bf1.23&\bf1.17\\
			\midrule
			Baseline~\cite{Grabner2019a}&\multirow{4}{*}{sofa}&\multirow{4}{*}{96.5}&4.40&4.96&3.78&2.57\\
			Image Refinement~\cite{Zakharov2019dpod}&&&4.30&3.87&3.15&2.54\\
			Mask Refinement~\cite{Kato2018renderer}&&&2.97&2.89&2.25&1.54\\
			Our Refinement&&&\bf2.28&\bf1.36&\bf1.19&\bf1.08\\
			\midrule
			Baseline~\cite{Grabner2019a}&\multirow{4}{*}{table}&\multirow{4}{*}{94.0}&10.18&7.72&6.17&5.54\\
			Image Refinement~\cite{Zakharov2019dpod}&&&9.81&7.07&5.80&5.40\\
			Mask Refinement~\cite{Kato2018renderer}&&&3.81&4.44&3.34&2.27\\
			Our Refinement&&&\bf2.59&\bf2.00&\bf1.48&\bf1.55\\
			\midrule
			\midrule
			Baseline~\cite{Grabner2019a}&\multirow{4}{*}{\textit{mean}}&\multirow{4}{*}{96.2}&6.75&6.21&4.76&3.71\\
			Image Refinement~\cite{Zakharov2019dpod}&&&6.46&5.43&4.31&3.67\\
			Mask Refinement~\cite{Kato2018renderer}&&&3.56&4.06&2.96&1.90\\
			Our Refinement&&&\bf2.56&\bf1.74&\bf1.34&\bf1.27\\
			\bottomrule
	\end{tabular}}
\end{table*}

\begin{table*}
	\centering
	\setlength{\tabcolsep}{10pt}
	\caption{Detailed 3D pose refinement results for individual categories on Pix3D. In this experiment, we automatically \textbf{retrieve 3D models} for refinement using the method presented in~\cite{Grabner2019b}. We outperform existing methods across all categories by a large margin.}
	\label{table:pix3d-retrieval}
	\resizebox{\columnwidth}{!}{%
		\begin{tabular}{lc|c|c|c|c|c}
			\toprule
			\multicolumn{2}{c}{}&\multicolumn{1}{c}{\bf Detection}&\multicolumn{1}{c}{\bf Rotation}&\multicolumn{1}{c}{\bf Translation}&\multicolumn{1}{c}{\bf Pose}&\multicolumn{1}{c}{\bf Projection}\\
			\cmidrule(lr){3-3}\cmidrule(lr){4-4}\cmidrule(lr){5-5}\cmidrule(lr){6-6}\cmidrule(lr){7-7}
			\multirow{2}{*}{Method}&\multicolumn{1}{c}{\multirow{2}{*}{Category}}&\multicolumn{1}{c}{\multirow{2}{*}{$Acc_{D_{0.5}}$}}&\multicolumn{1}{c}{$MedErr_R$}&\multicolumn{1}{c}{$MedErr_{t}$}&\multicolumn{1}{c}{$MedErr_{R,t}$}&\multicolumn{1}{c}{$MedErr_{P}$}\\
			&\multicolumn{1}{c}{}&\multicolumn{1}{c}{}&\multicolumn{1}{c}{$\cdot1$}&\multicolumn{1}{c}{$\cdot10^{2}$}&\multicolumn{1}{c}{$\cdot10^{2}$}&\multicolumn{1}{c}{$\cdot10^{2}$}\\
			\midrule
			Baseline~\cite{Grabner2019a}&\multirow{4}{*}{bed}&\multirow{4}{*}{99.0}&5.07&6.68&5.18&3.42\\
			Image Refinement~\cite{Zakharov2019dpod}&&&4.65&5.86&4.41&3.38\\
			Mask Refinement~\cite{Kato2018renderer}&&&4.40&5.32&4.21&2.69\\
			Our Refinement&&&\bf2.95&\bf2.75&\bf2.18&\bf1.83\\
			\midrule
			Baseline~\cite{Grabner2019a}&\multirow{4}{*}{chair}&\multirow{4}{*}{95.2}&7.36&5.49&3.90&3.32\\
			Image Refinement~\cite{Zakharov2019dpod}&&&7.15&5.21&3.67&3.35\\
			Mask Refinement~\cite{Kato2018renderer}&&&7.22&6.32&4.53&3.31\\
			Our Refinement&&&\bf4.89&\bf2.87&\bf2.04&\bf2.19\\
			\midrule
			Baseline~\cite{Grabner2019a}&\multirow{4}{*}{sofa}&\multirow{4}{*}{96.5}&4.40&4.96&3.78&2.57\\
			Image Refinement~\cite{Zakharov2019dpod}&&&4.34&3.75&3.02&2.54\\
			Mask Refinement~\cite{Kato2018renderer}&&&3.33&3.00&2.34&1.63\\
			Our Refinement&&&\bf2.60&\bf1.60&\bf1.42&\bf1.19\\
			\midrule
			Baseline~\cite{Grabner2019a}&\multirow{4}{*}{table}&\multirow{4}{*}{94.0}&10.18&7.72&6.17&5.54\\
			Image Refinement~\cite{Zakharov2019dpod}&&&9.73&7.23&6.22&5.68\\
			Mask Refinement~\cite{Kato2018renderer}&&&6.92&6.34&5.52&4.85\\
			Our Refinement&&&\bf4.73&\bf3.38&\bf2.93&\bf3.49\\
			\midrule
			\midrule
			Baseline~\cite{Grabner2019a}&\multirow{4}{*}{\textit{mean}}&\multirow{4}{*}{96.2}&6.75&6.21&4.76&3.71\\
			Image Refinement~\cite{Zakharov2019dpod}&&&6.47&5.51&4.33&3.74\\
			Mask Refinement~\cite{Kato2018renderer}&&&5.47&5.25&4.15&3.12\\
			Our Refinement&&&\bf3.79&\bf2.65&\bf2.14&\bf2.18\\
			\bottomrule
	\end{tabular}}
\end{table*}

\section{Additional Qualitative Results}
\label{sec:add-quali}

\begin{figure*}[h!]
	\setlength{\tabcolsep}{1pt}
	\setlength{\fboxsep}{-2pt}
	\setlength{\fboxrule}{2pt}
	\definecolor{boxgreen}{rgb}{0.3, 1.0, 0.3}
	\definecolor{boxred}{rgb}{1.0, 0.3, 0.3}
	\newcommand{\colImgN}[1]{{\includegraphics[width=0.16\linewidth]{#1}}}
	\newcommand{\colImgR}[1]{{\color{boxred}\fbox{\colImgN{#1}}}}
	\newcommand{\colImgG}[1]{{\color{boxgreen}\fbox{\colImgN{#1}}}}
	\centering
	\begin{tabular}{cccccc}
		\colImgN{}& \colImgN{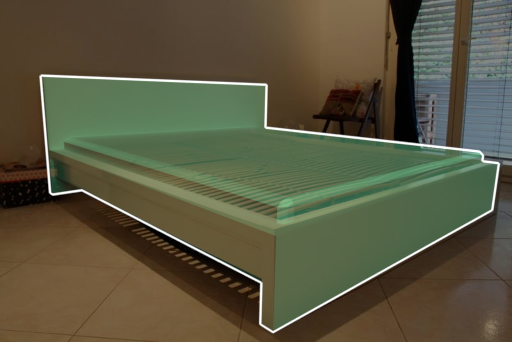}& \colImgN{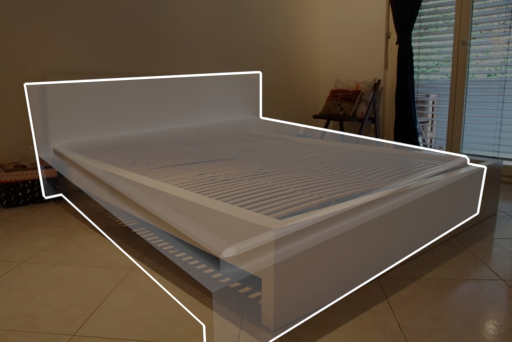}& \colImgN{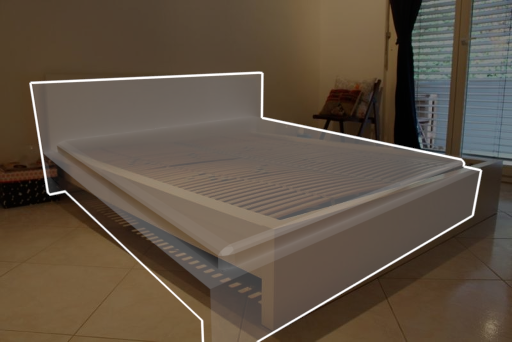}& \colImgN{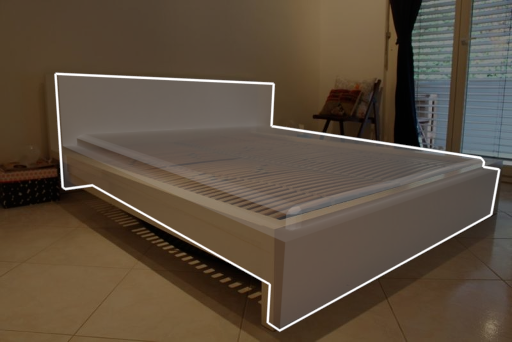}& \colImgN{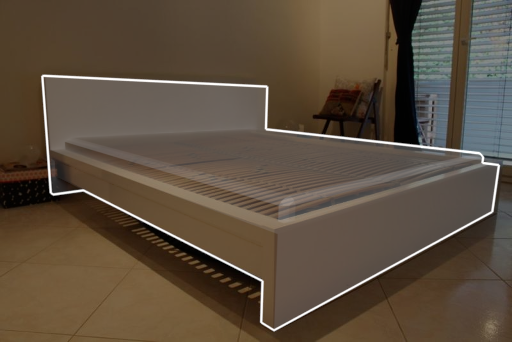}\\[-1pt]
		\colImgN{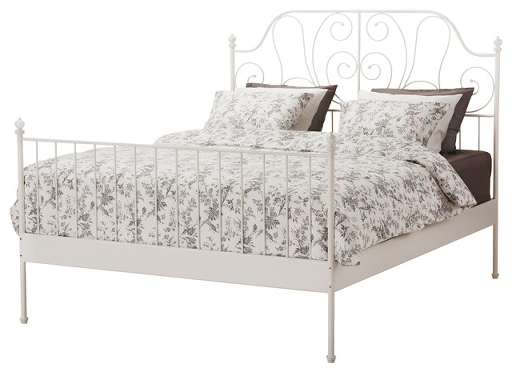}& \colImgN{Images/Collage/0090_gt.png}& \colImgN{Images/Collage/0090_base.png}& \colImgN{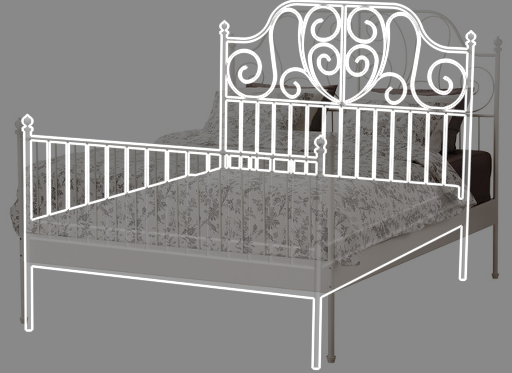}& \colImgN{Images/Collage/0090_mask.png}& \colImgN{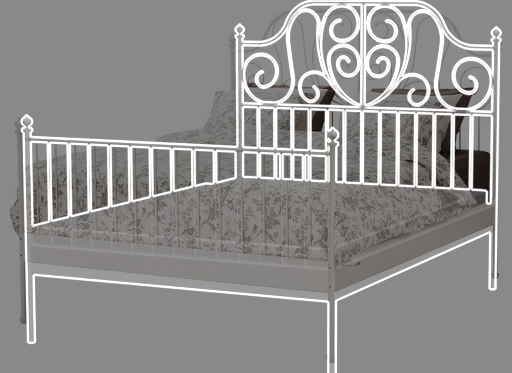}\\[-1pt]
		\colImgN{}& \colImgN{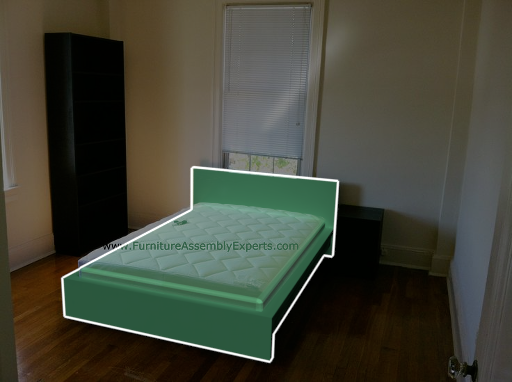}& \colImgN{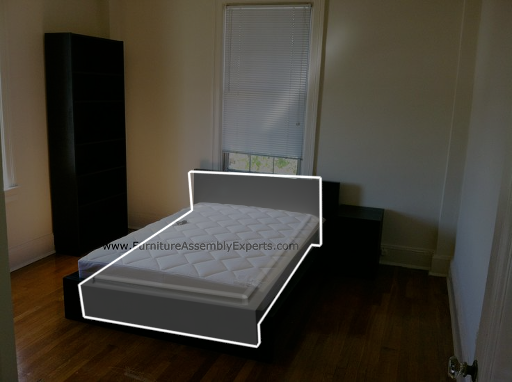}& \colImgN{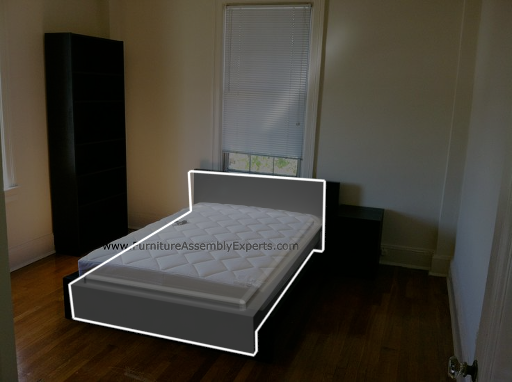}& \colImgN{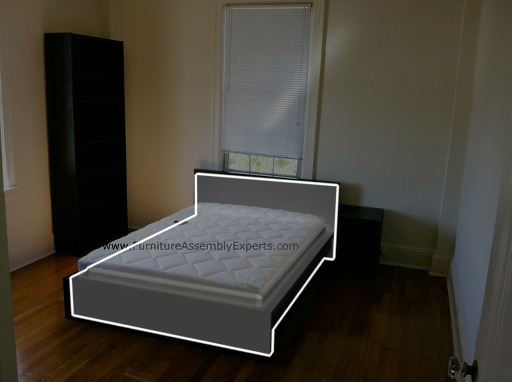}& \colImgN{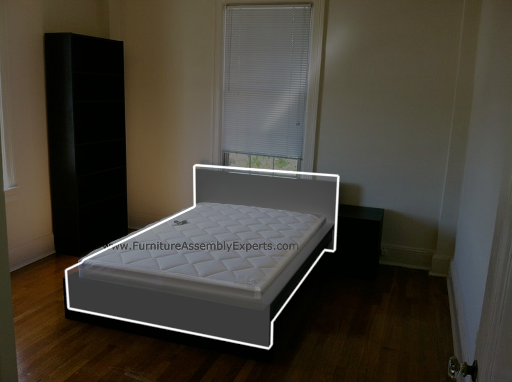}\\[-1pt]
		\colImgN{}& \colImgN{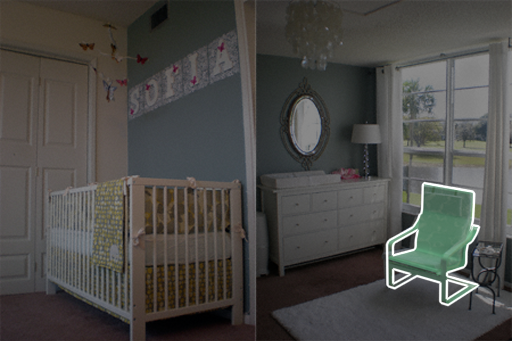}& \colImgN{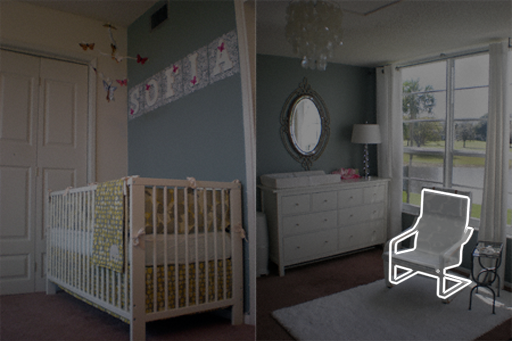}& \colImgN{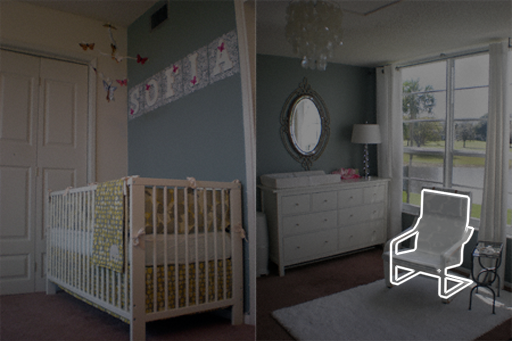}& \colImgN{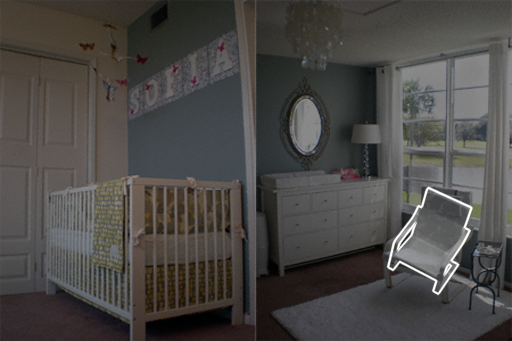}& \colImgN{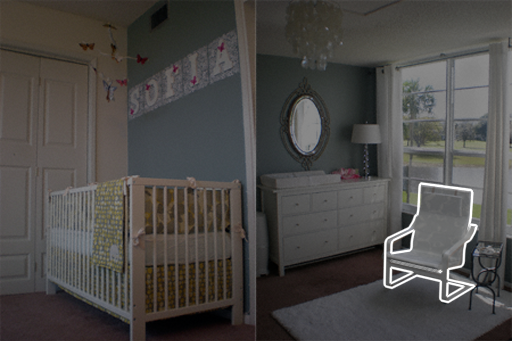}\\[-1pt]
		\colImgN{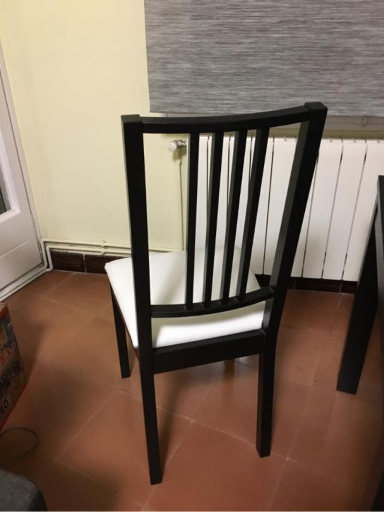}& \colImgN{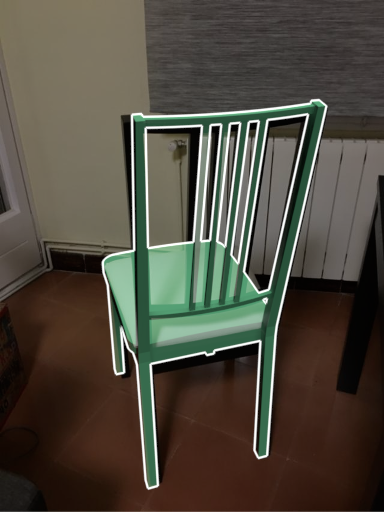}& \colImgN{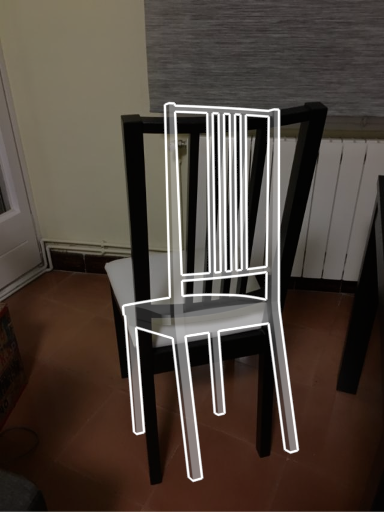}& \colImgN{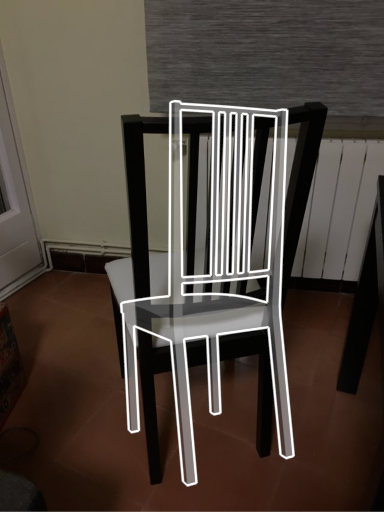}& \colImgN{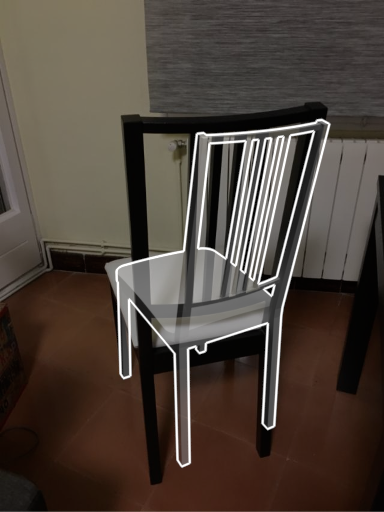}& \colImgN{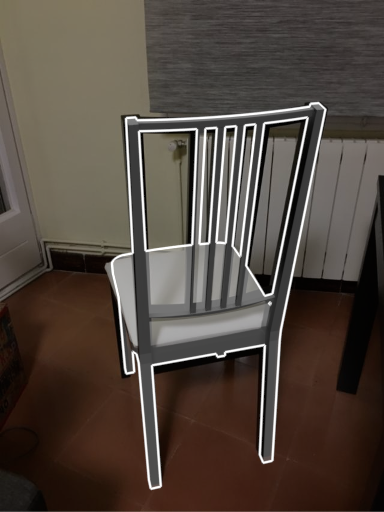}\\[-1pt]
		\colImgN{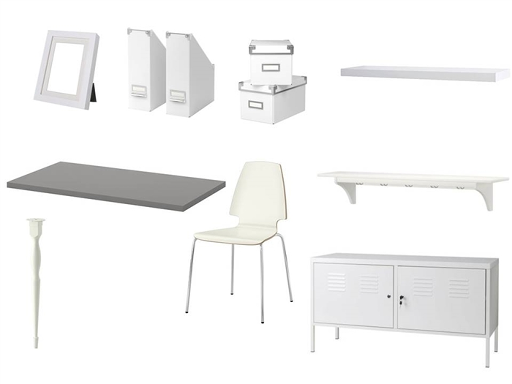}& \colImgN{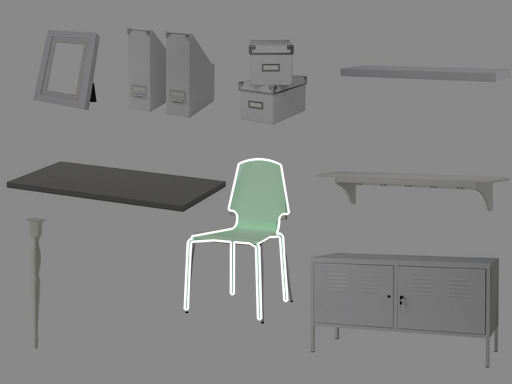}& \colImgN{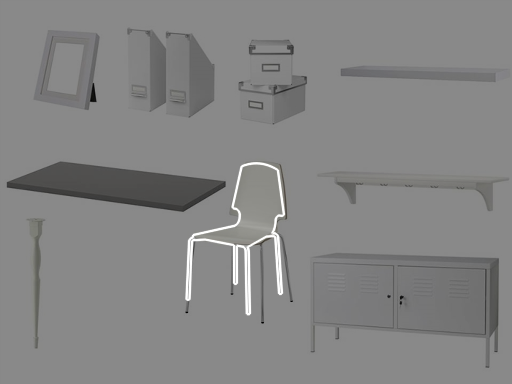}& \colImgN{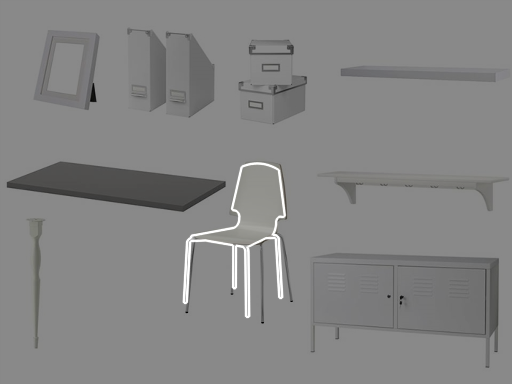}& \colImgN{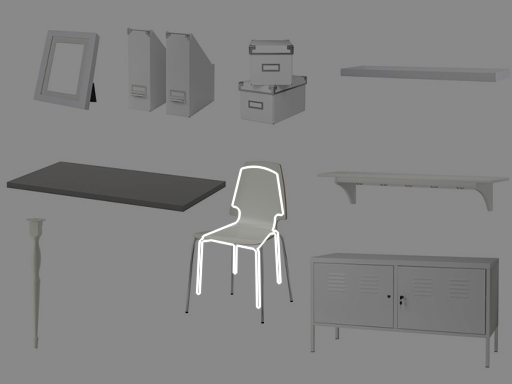}& \colImgN{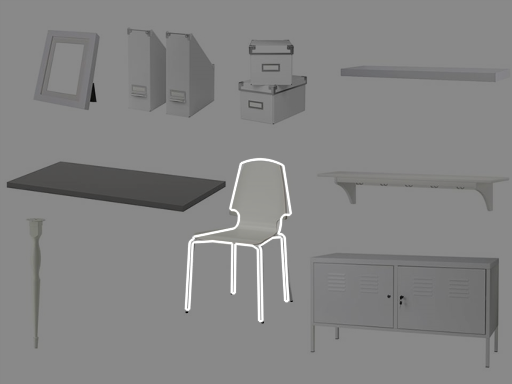}\\[-1pt]
		\colImgN{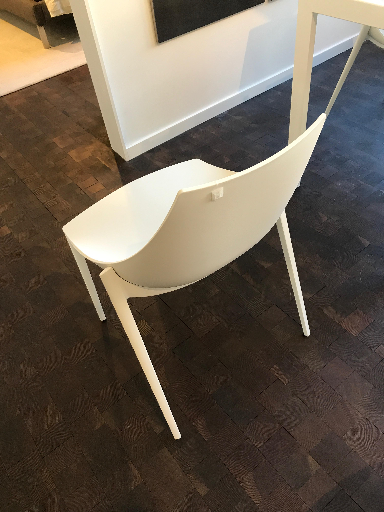}& \colImgN{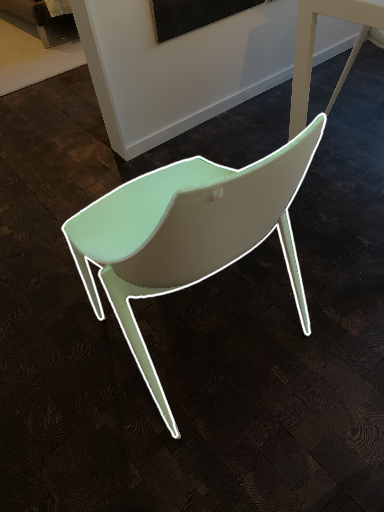}& \colImgN{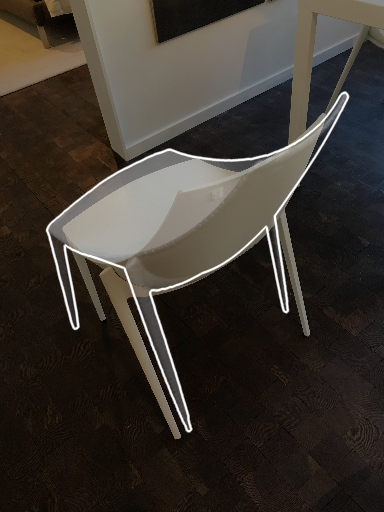}& \colImgN{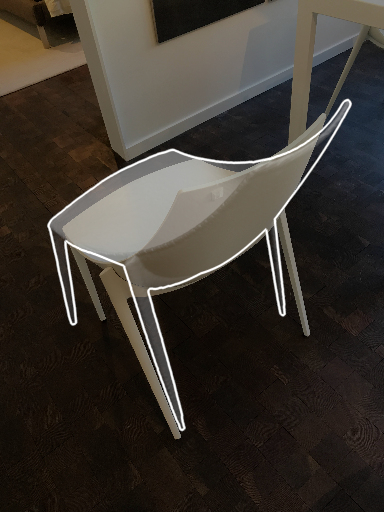}& \colImgN{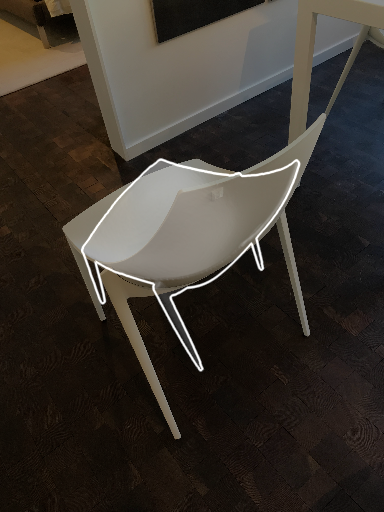}& \colImgN{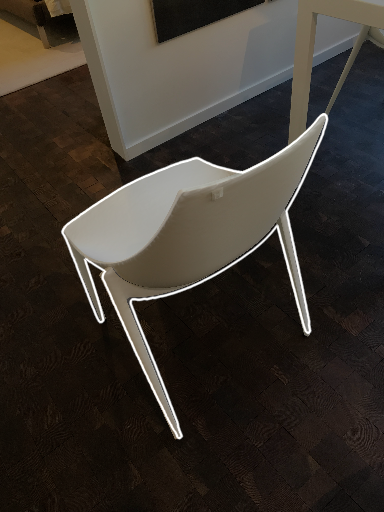}\\[-1pt]
		\colImgN{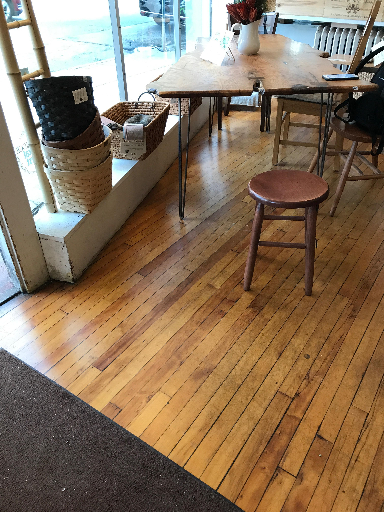}& \colImgN{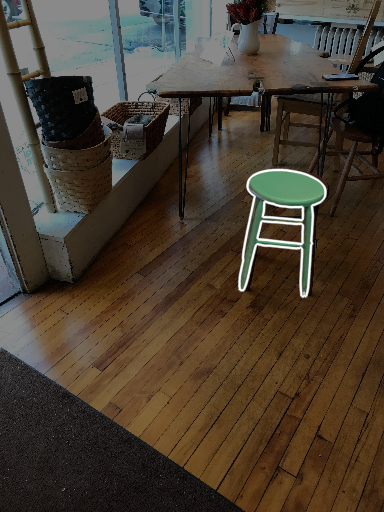}& \colImgN{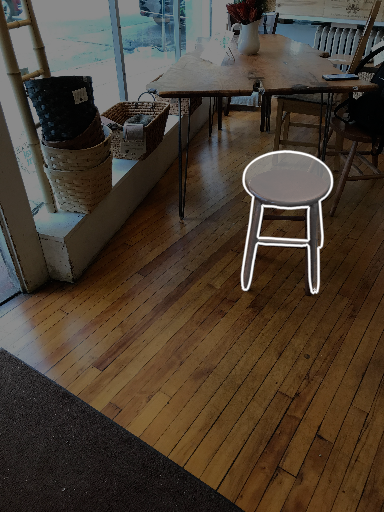}& \colImgN{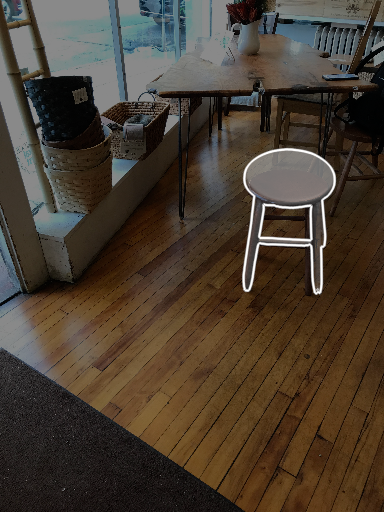}& \colImgN{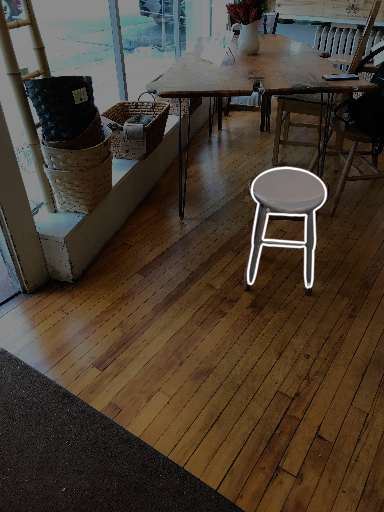}& \colImgN{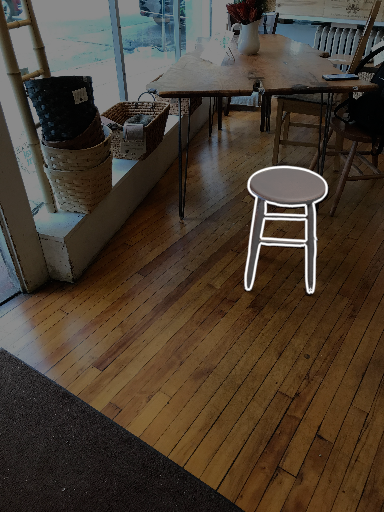}\\[-1pt]
		\footnotesize Image&\footnotesize GT&\footnotesize\cite{Grabner2019a}&\footnotesize\cite{Zakharov2019dpod}&\footnotesize\cite{Kato2018renderer}&\footnotesize Ours\\
	\end{tabular}
	\caption{Additional qualitative 3D pose refinement results for objects of different categories. We project the ground truth 3D model on the image using the predicted 3D pose. Our approach overcomes the limitations of previous methods and predicts fine-grained 3D poses which are in many cases visually indistinguishable from the ground truth. Best viewed in \textbf{digital zoom}.}
	\label{fig:collage2}
\end{figure*}

\begin{figure*}[h!]
	\setlength{\tabcolsep}{1pt}
	\setlength{\fboxsep}{-2pt}
	\setlength{\fboxrule}{2pt}
	\definecolor{boxgreen}{rgb}{0.3, 1.0, 0.3}
	\definecolor{boxred}{rgb}{1.0, 0.3, 0.3}
	\newcommand{\colImgN}[1]{{\includegraphics[width=0.16\linewidth]{#1}}}
	\newcommand{\colImgR}[1]{{\color{boxred}\fbox{\colImgN{#1}}}}
	\newcommand{\colImgG}[1]{{\color{boxgreen}\fbox{\colImgN{#1}}}}
	\centering
	\begin{tabular}{cccccc}
		\colImgN{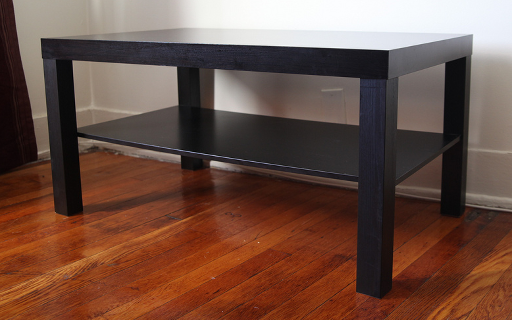}& \colImgN{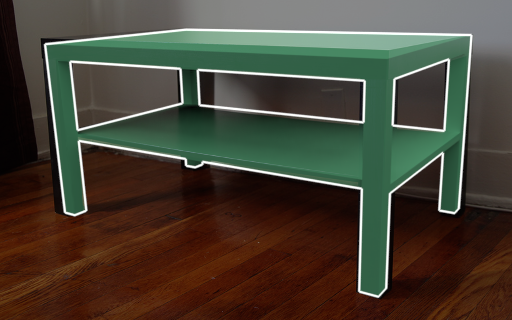}& \colImgN{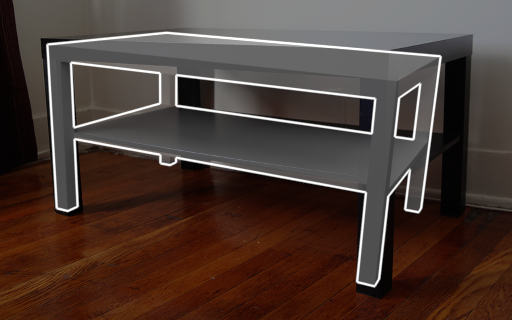}& \colImgN{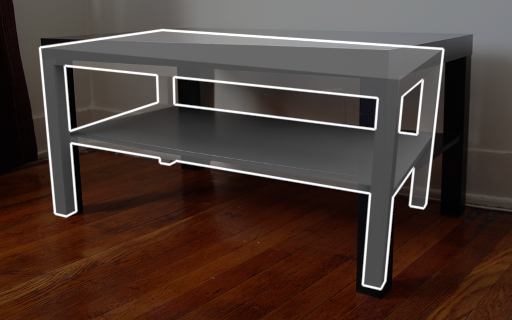}& \colImgN{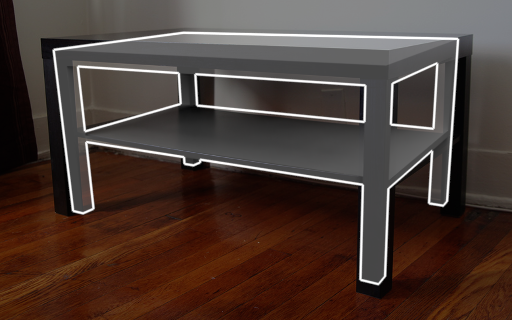}& \colImgN{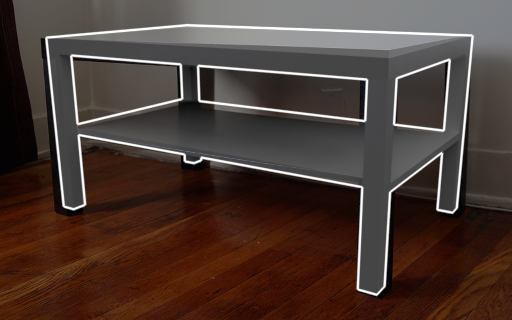}\\[-1pt]
		\colImgN{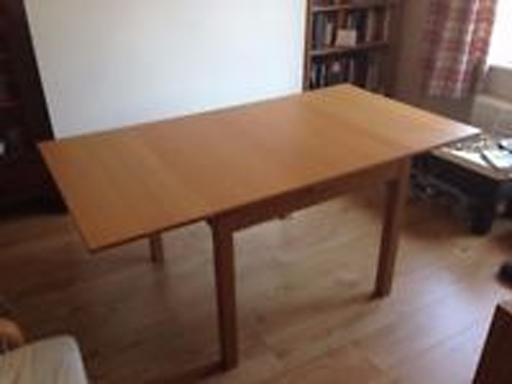}& \colImgN{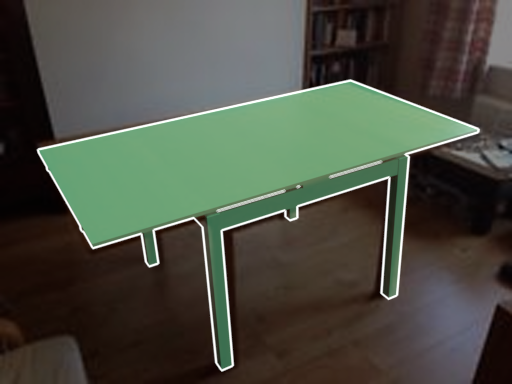}& \colImgN{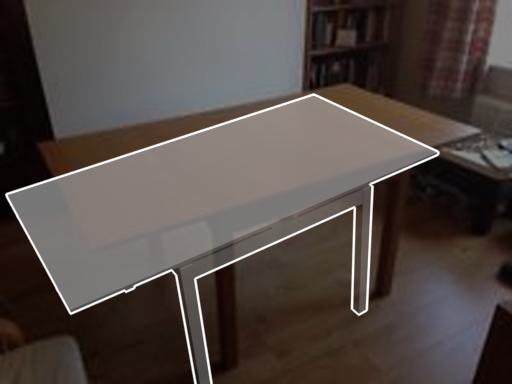}& \colImgN{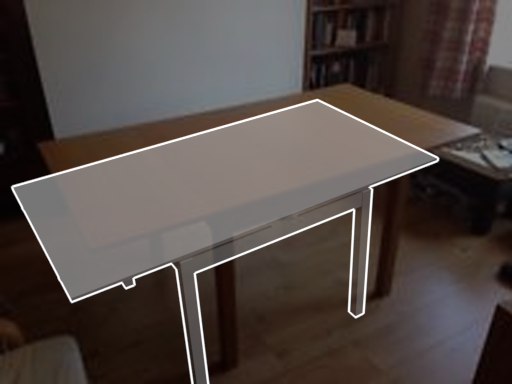}& \colImgN{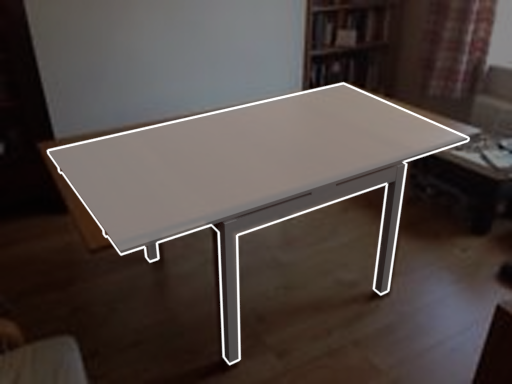}& \colImgN{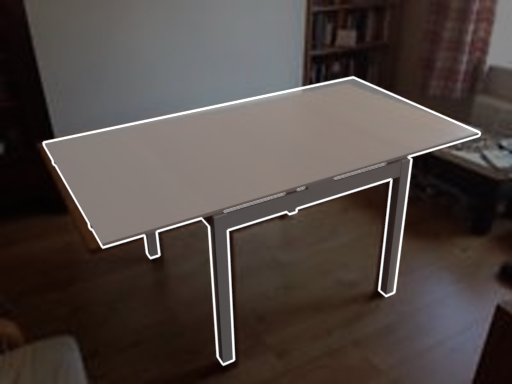}\\[-1pt]
		\colImgN{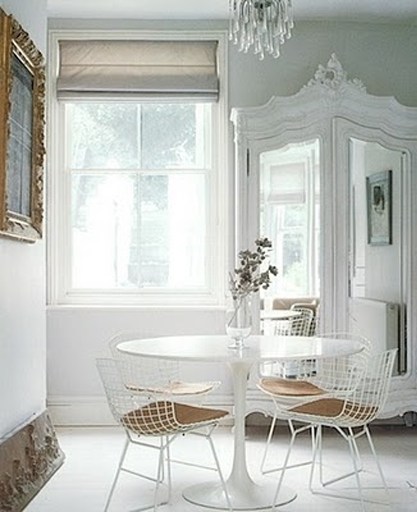}& \colImgN{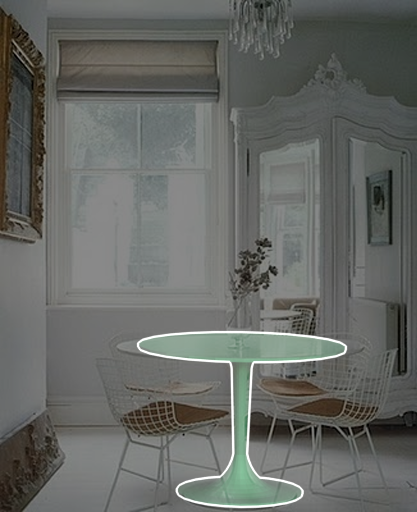}& \colImgN{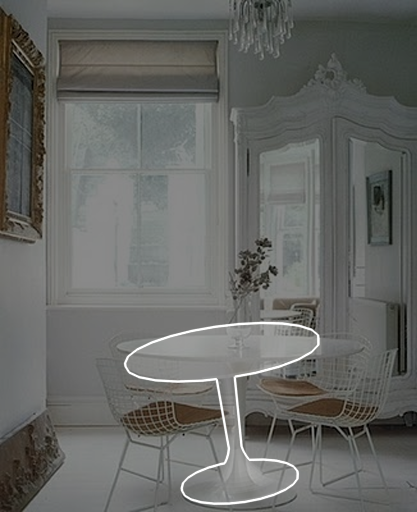}& \colImgN{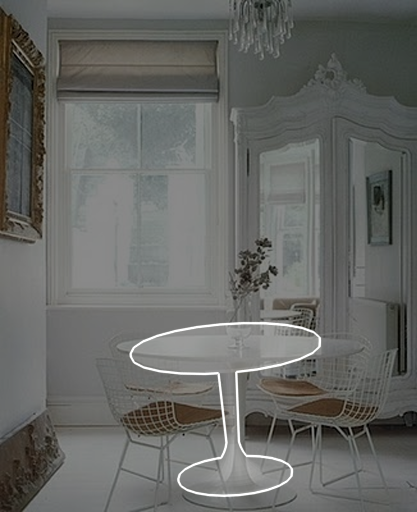}& \colImgN{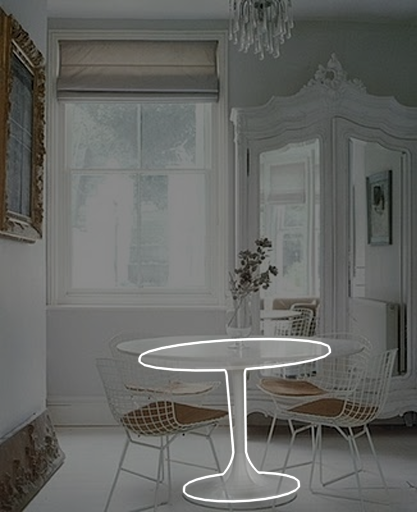}& \colImgN{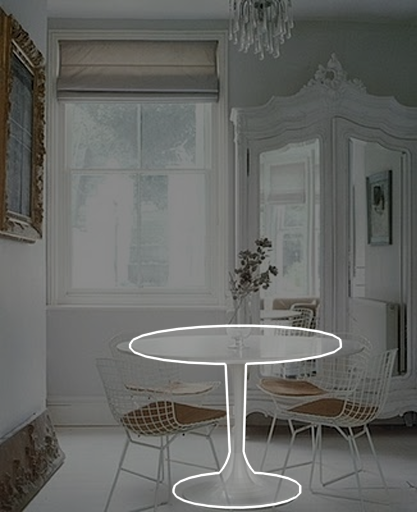}\\[-1pt]
		\colImgN{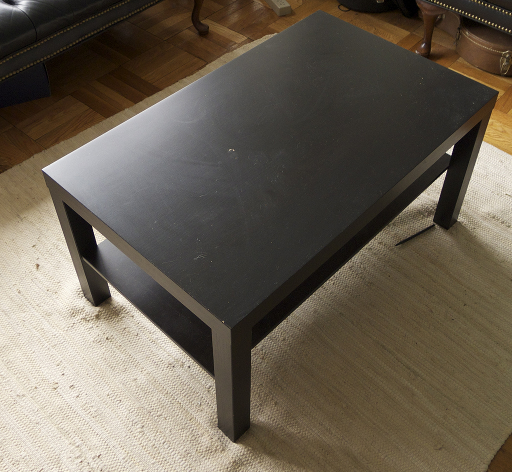}& \colImgN{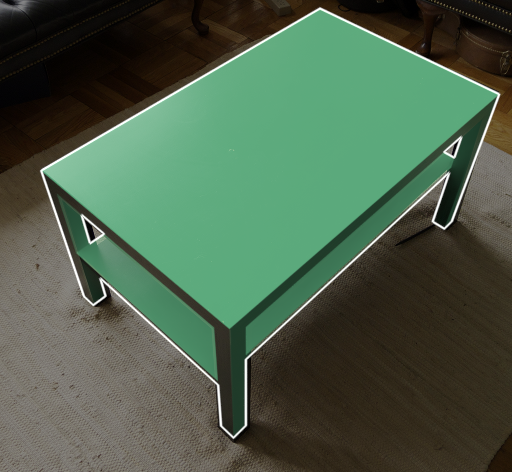}& \colImgN{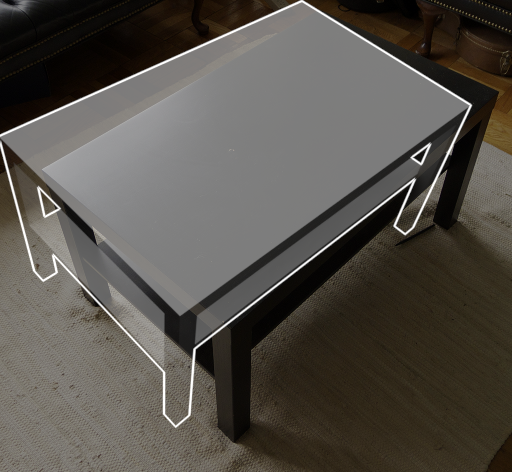}& \colImgN{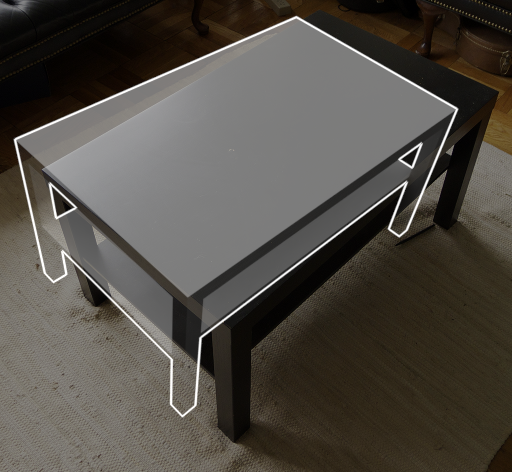}& \colImgN{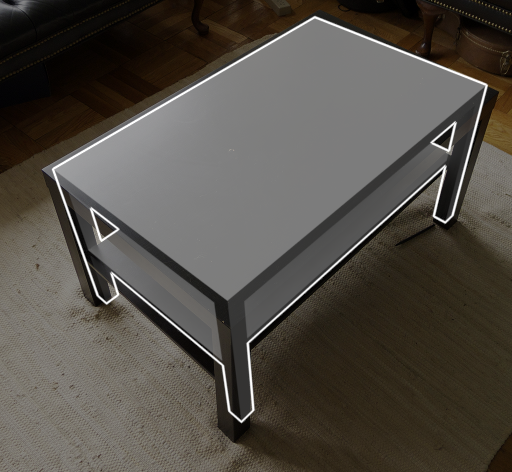}& \colImgN{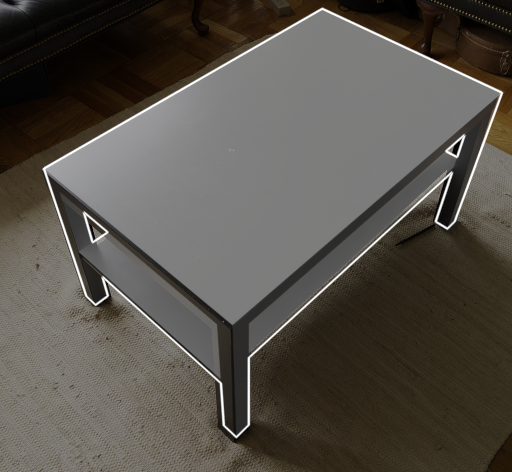}\\[-1pt]
		\colImgN{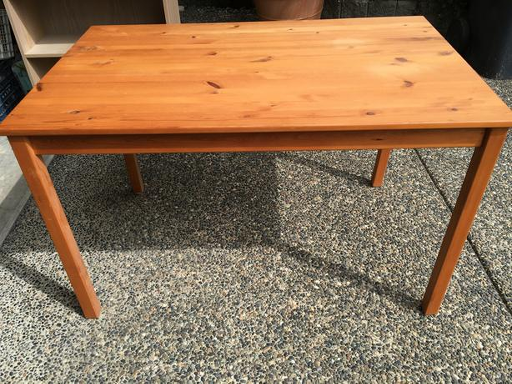}& \colImgN{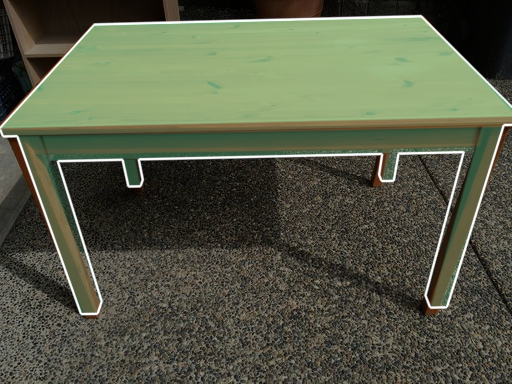}& \colImgN{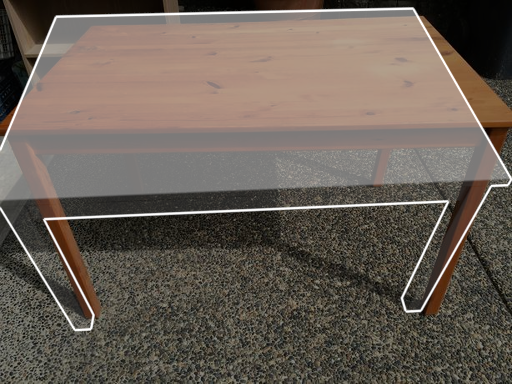}& \colImgN{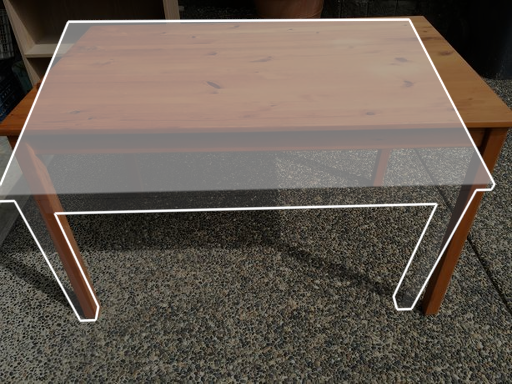}& \colImgN{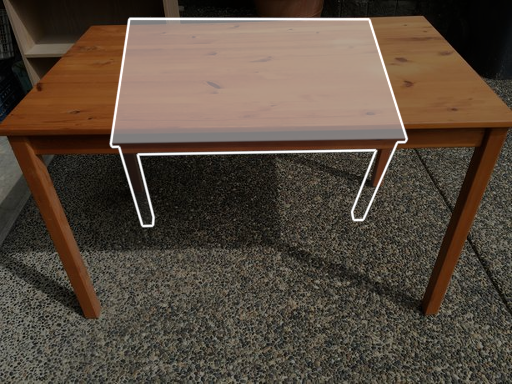}& \colImgN{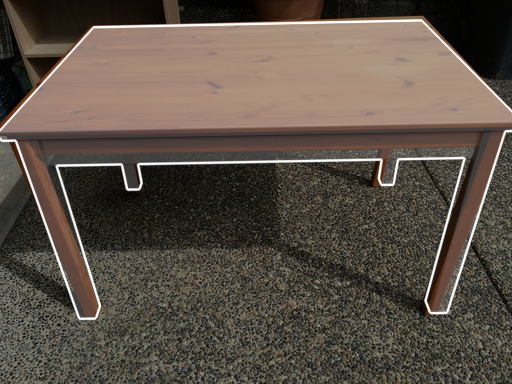}\\[-1pt]
		\colImgN{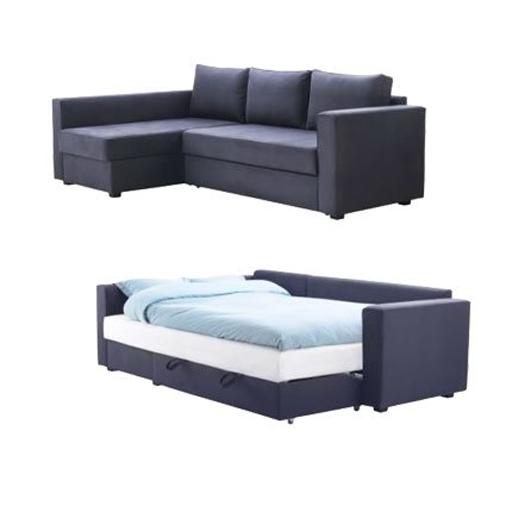}& \colImgN{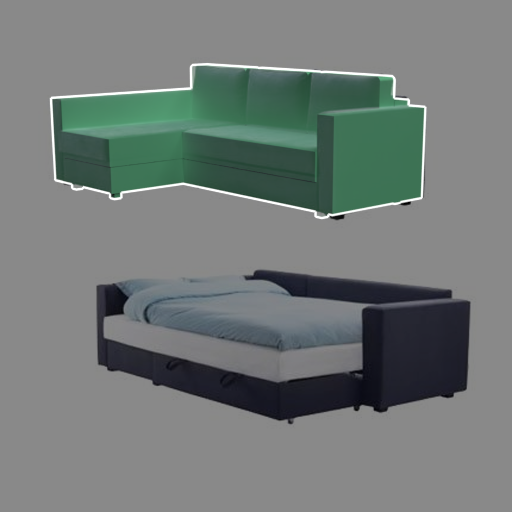}& \colImgN{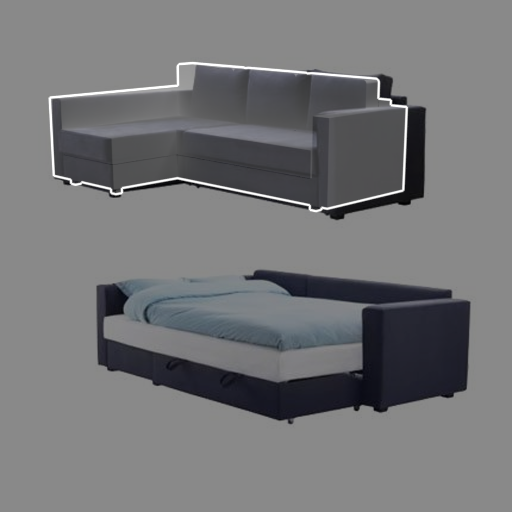}& \colImgN{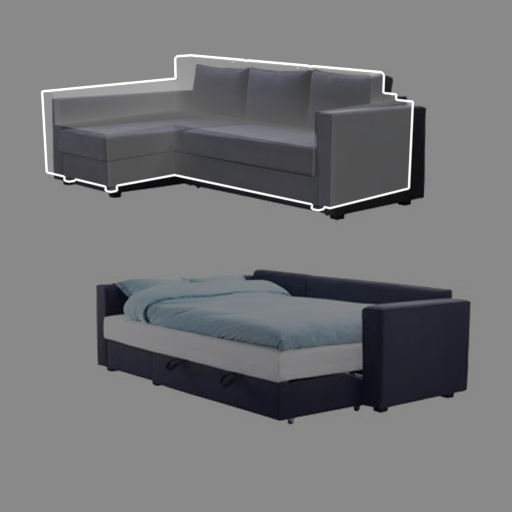}& \colImgN{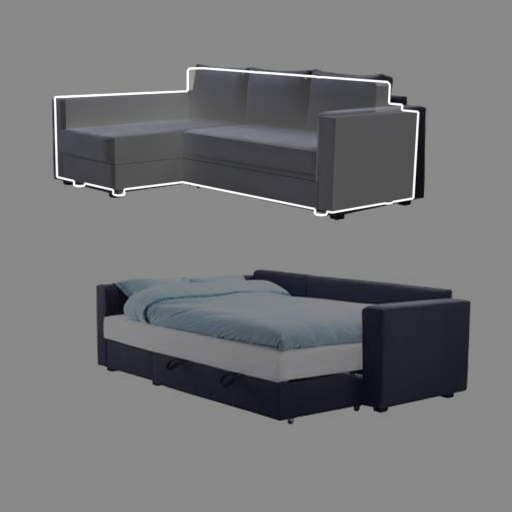}& \colImgN{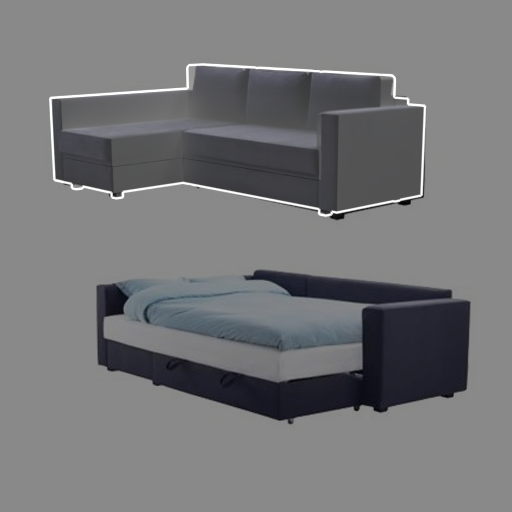}\\[-1pt]
		\colImgN{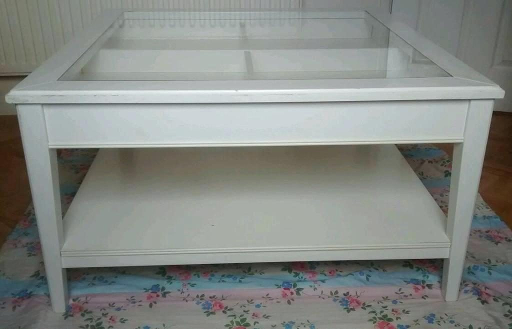}& \colImgN{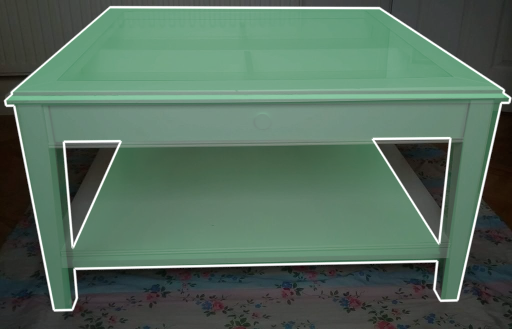}& \colImgN{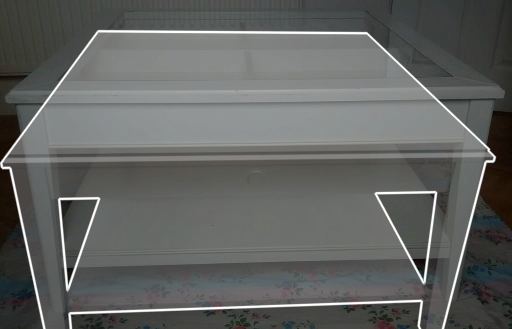}& \colImgN{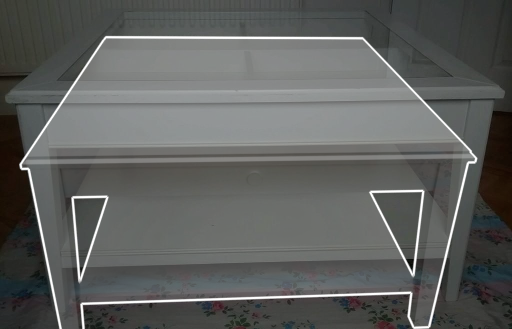}& \colImgN{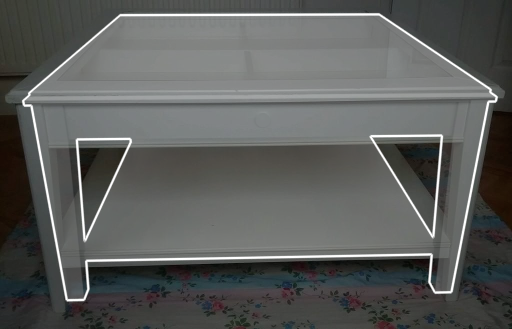}& \colImgN{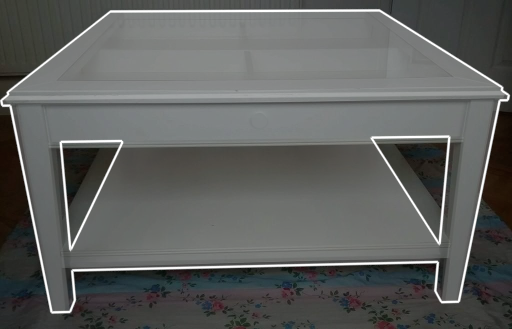}\\[-1pt]
		\colImgN{}& \colImgN{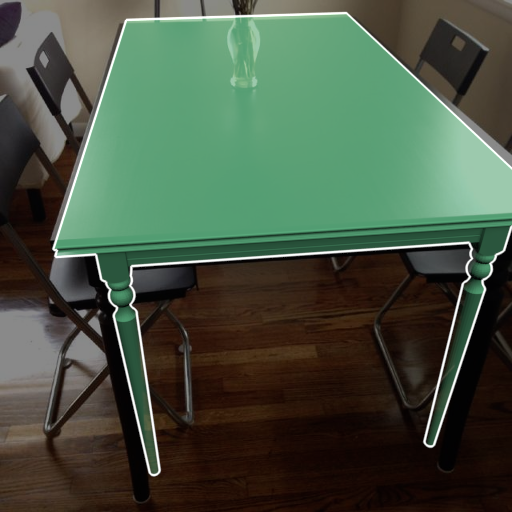}& \colImgN{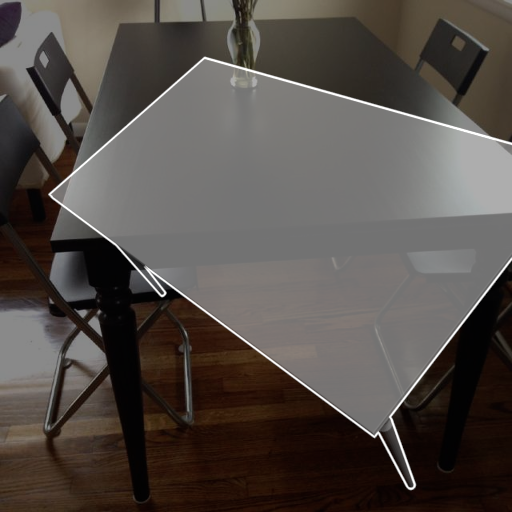}& \colImgN{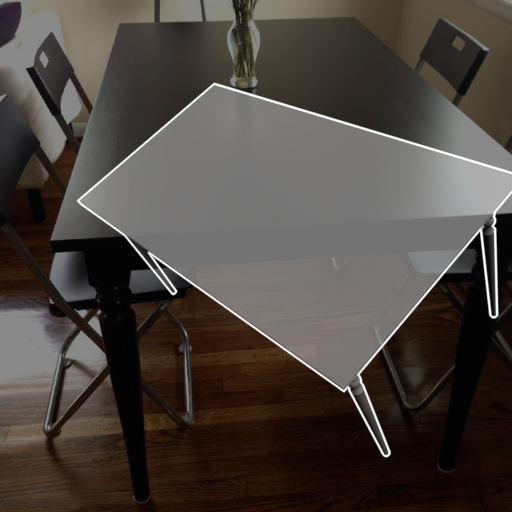}& \colImgN{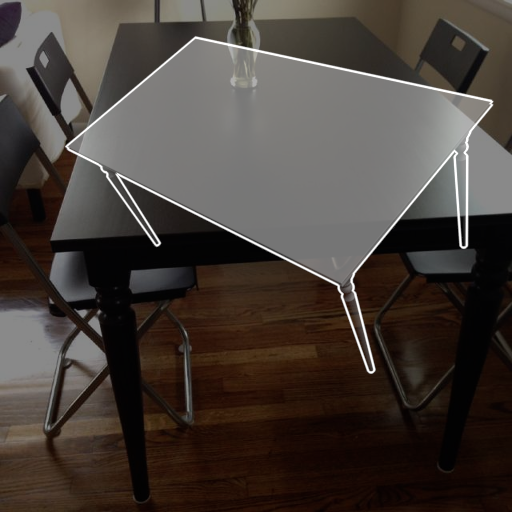}& \colImgN{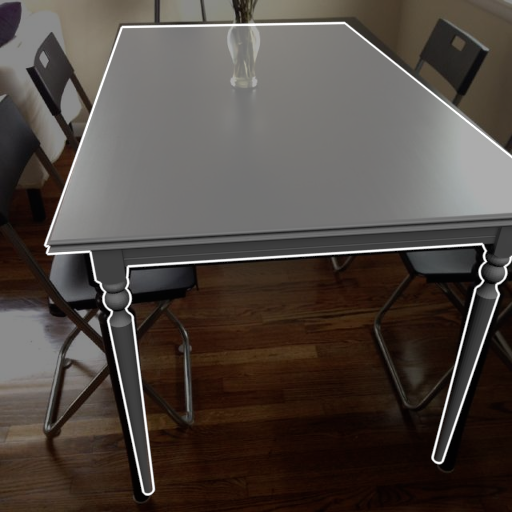}\\[-1pt]
		\colImgN{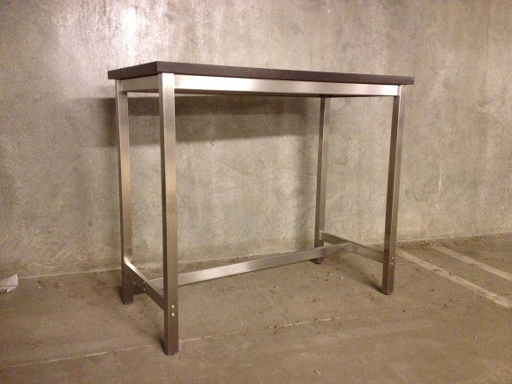}& \colImgN{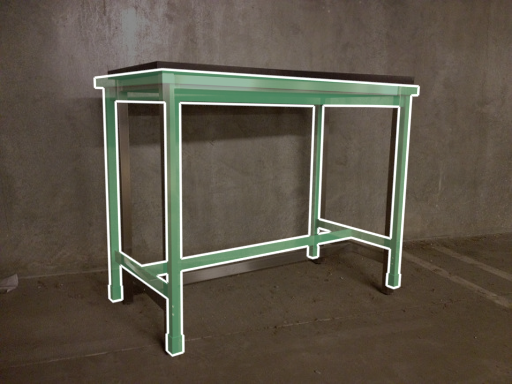}& \colImgN{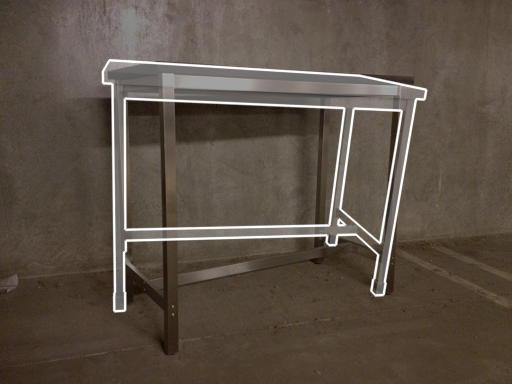}& \colImgN{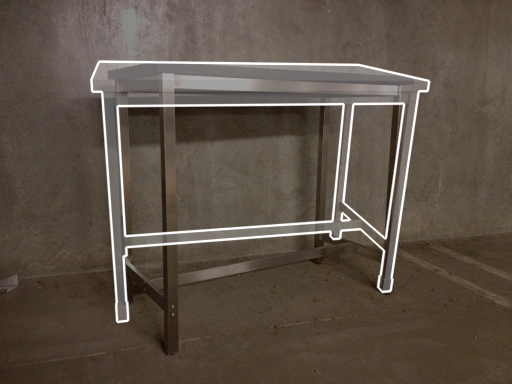}& \colImgN{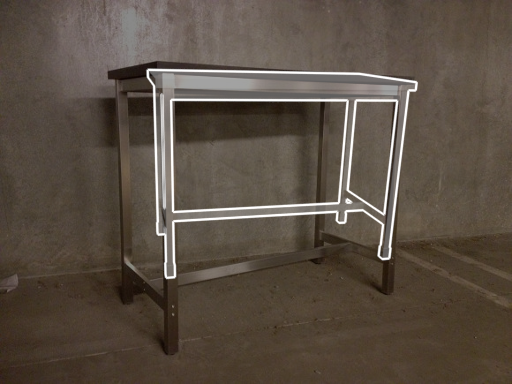}& \colImgN{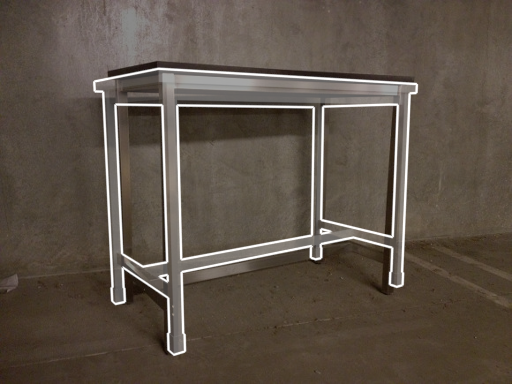}\\[-1pt]
		\footnotesize Image&\footnotesize GT&\footnotesize\cite{Grabner2019a}&\footnotesize\cite{Zakharov2019dpod}&\footnotesize\cite{Kato2018renderer}&\footnotesize Ours\\
	\end{tabular}
	\caption{Additional qualitative 3D pose refinement results for objects of different categories. We project the ground truth 3D model on the image using the predicted 3D pose. Our approach overcomes the limitations of previous methods and predicts fine-grained 3D poses which are in many cases visually indistinguishable from the ground truth. Best viewed in \textbf{digital zoom}.}
	\label{fig:collage3}
\end{figure*}

\begin{figure*}[h!]
	\setlength{\tabcolsep}{1pt}
	\setlength{\fboxsep}{-2pt}
	\setlength{\fboxrule}{2pt}
	\definecolor{boxgreen}{rgb}{0.3, 1.0, 0.3}
	\definecolor{boxred}{rgb}{1.0, 0.3, 0.3}
	\newcommand{\colImgN}[1]{{\includegraphics[width=0.16\linewidth]{#1}}}
	\newcommand{\colImgR}[1]{{\color{boxred}\fbox{\colImgN{#1}}}}
	\newcommand{\colImgG}[1]{{\color{boxgreen}\fbox{\colImgN{#1}}}}
	\centering
	\begin{tabular}{cccccc}
		\colImgN{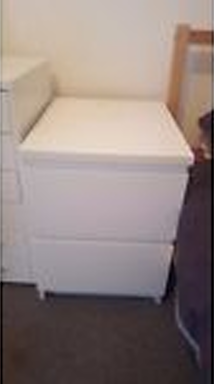}& \colImgN{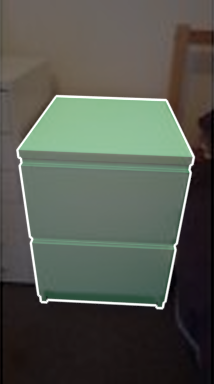}& \colImgN{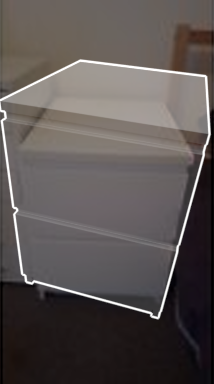}& \colImgN{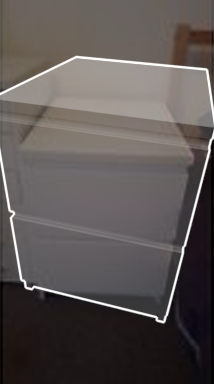}& \colImgN{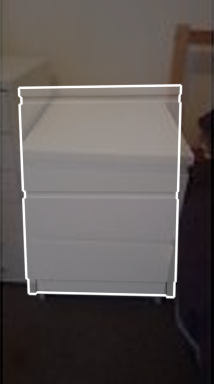}& \colImgN{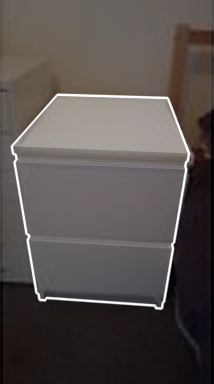}\\[-1pt]
		\colImgN{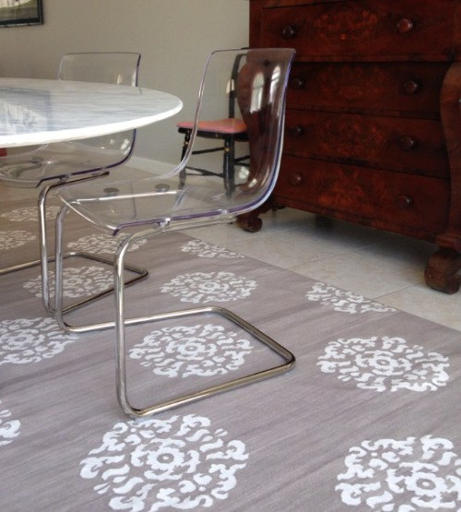}& \colImgN{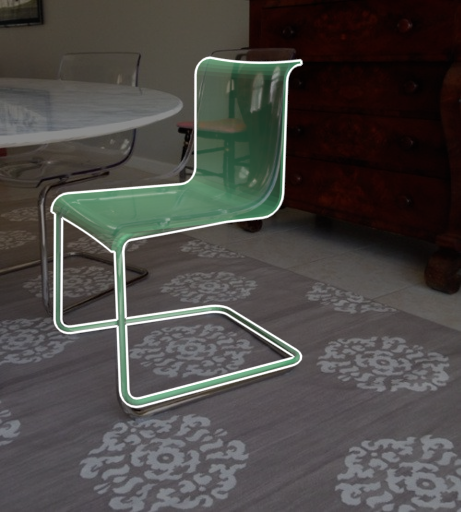}& \colImgN{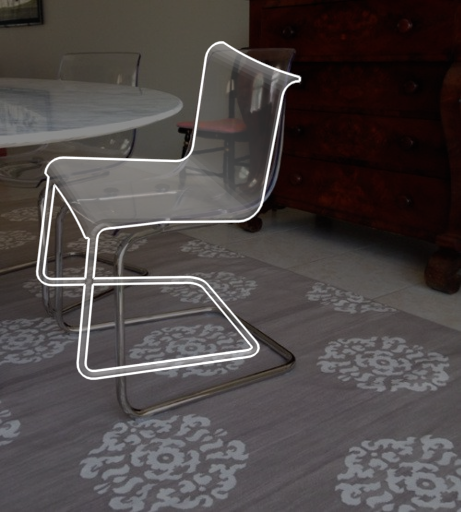}& \colImgN{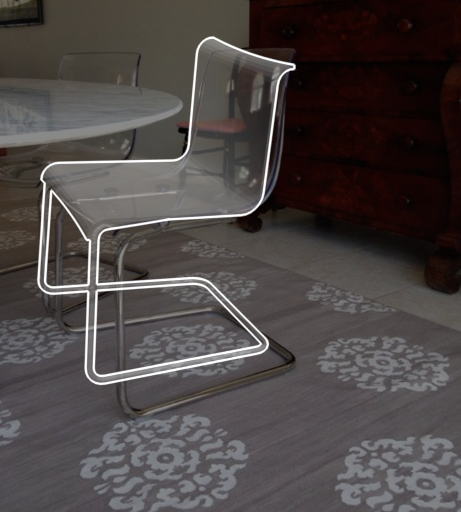}& \colImgN{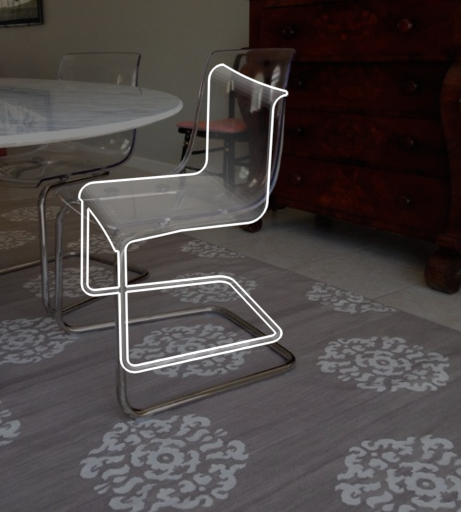}& \colImgN{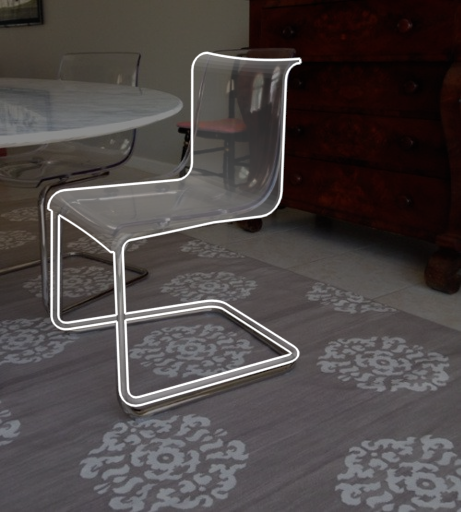}\\[-1pt]
		\colImgN{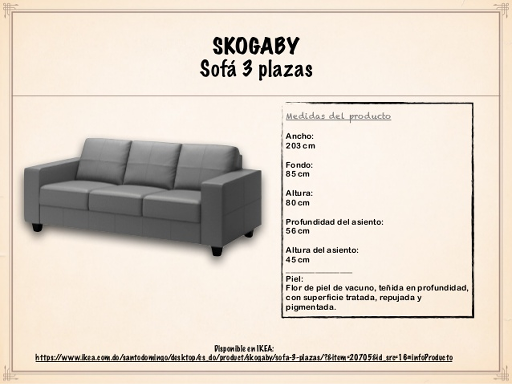}& \colImgN{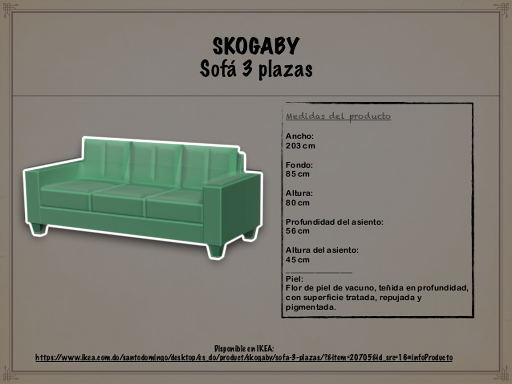}& \colImgN{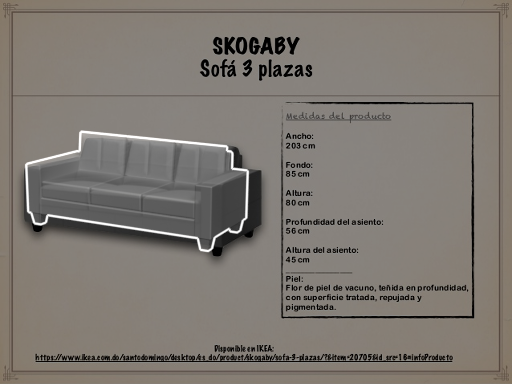}& \colImgN{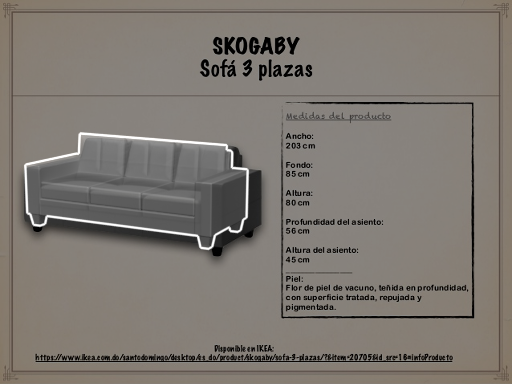}& \colImgN{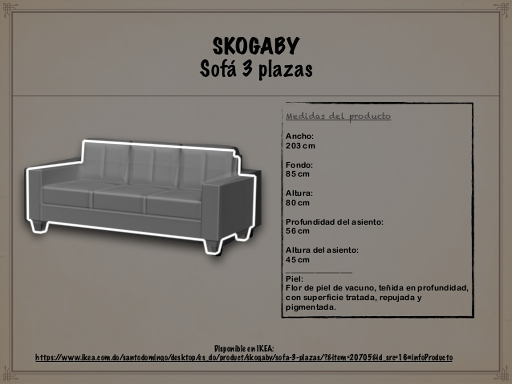}& \colImgN{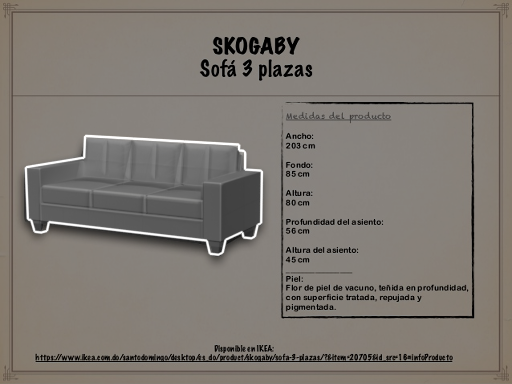}\\[-1pt]
		\colImgN{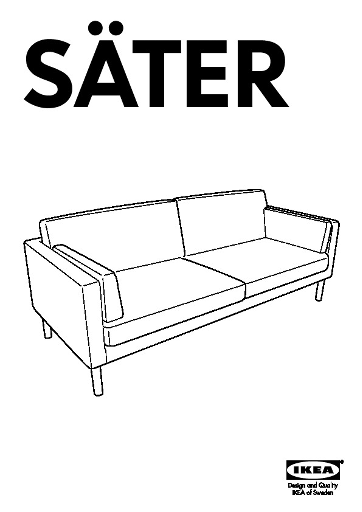}& \colImgN{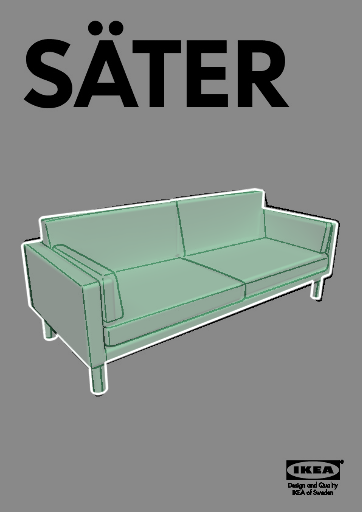}& \colImgN{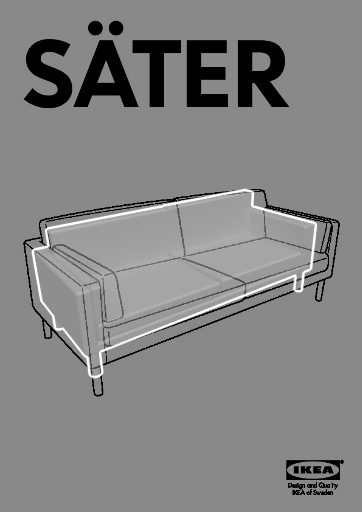}& \colImgN{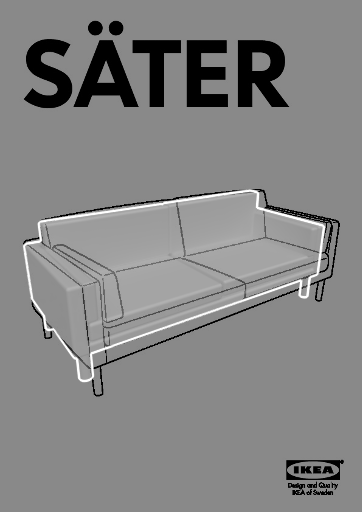}& \colImgN{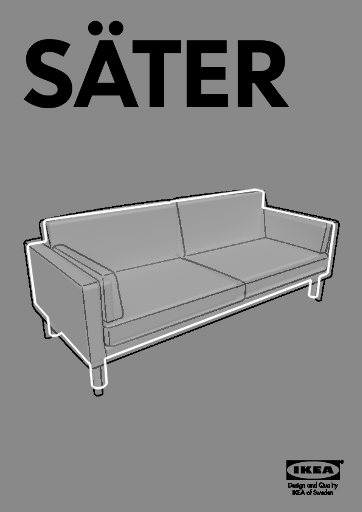}& \colImgN{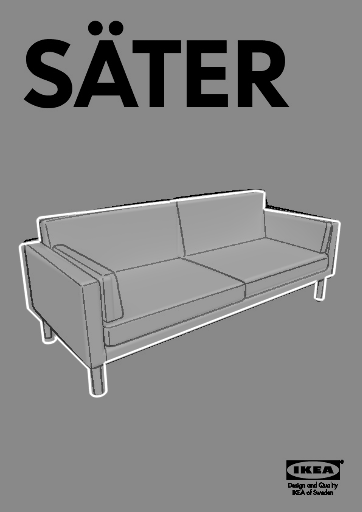}\\[-1pt]
		\colImgN{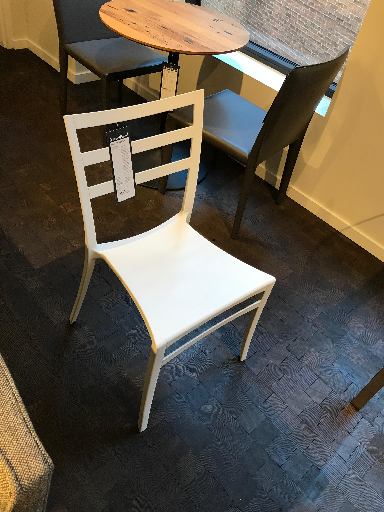}& \colImgN{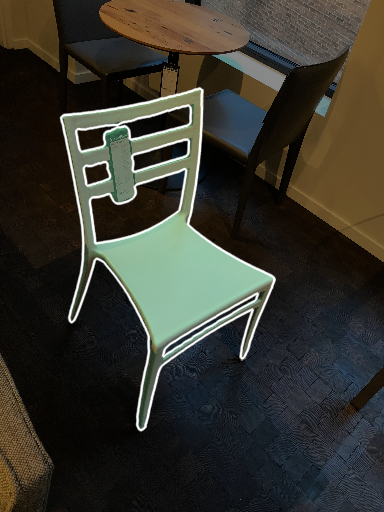}& \colImgN{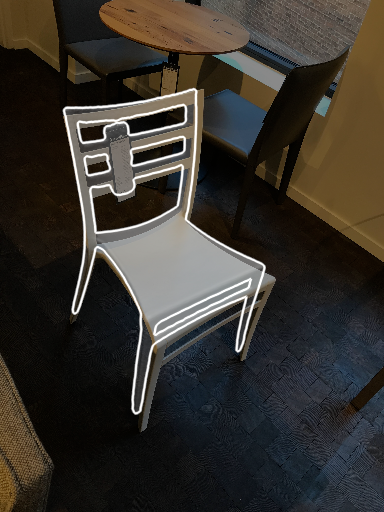}& \colImgN{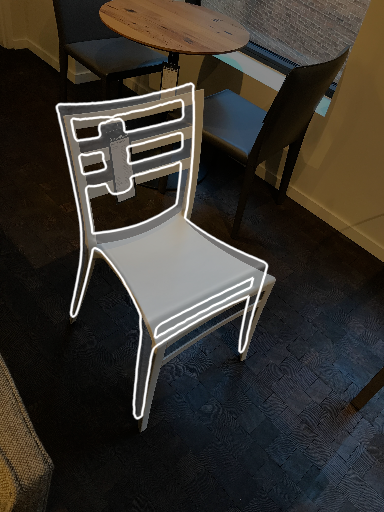}& \colImgN{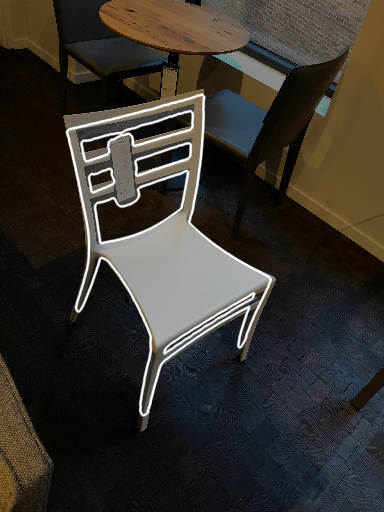}& \colImgN{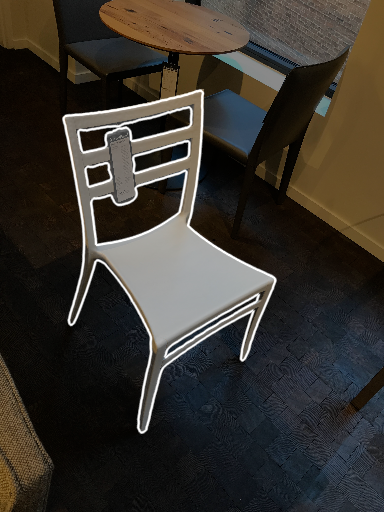}\\[-1pt]
		\colImgN{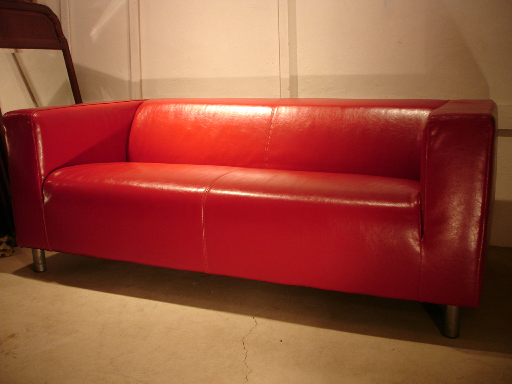}& \colImgN{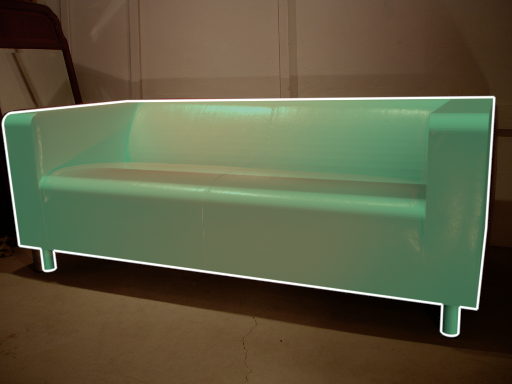}& \colImgN{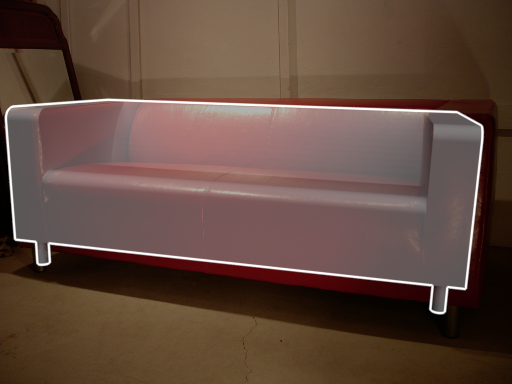}& \colImgN{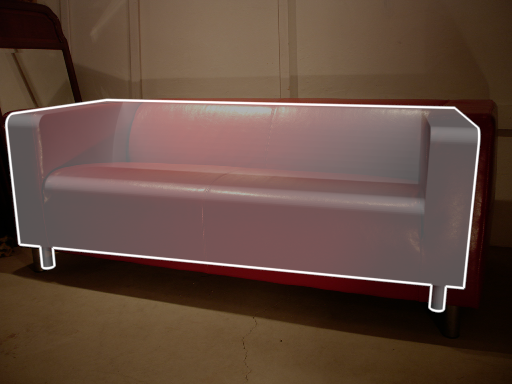}& \colImgN{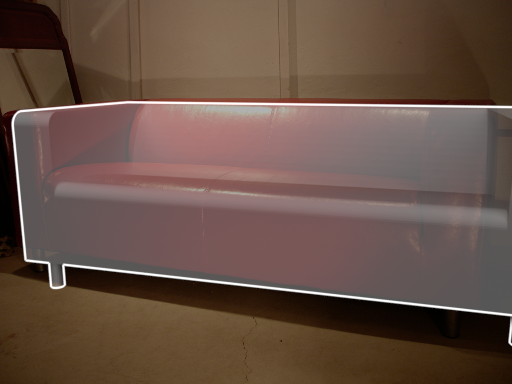}& \colImgN{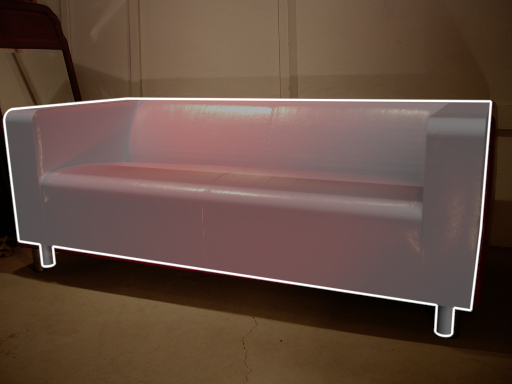}\\[-1pt]
		\colImgN{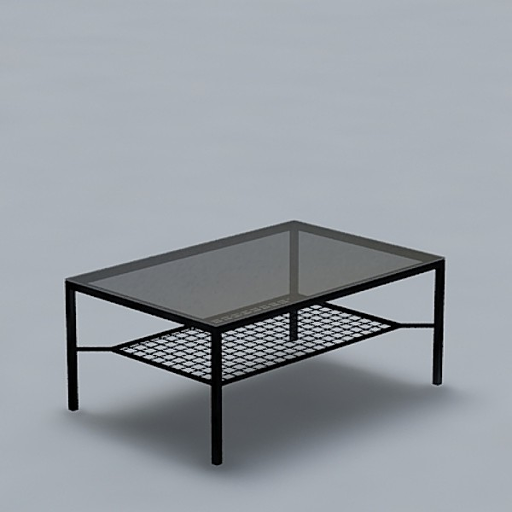}& \colImgN{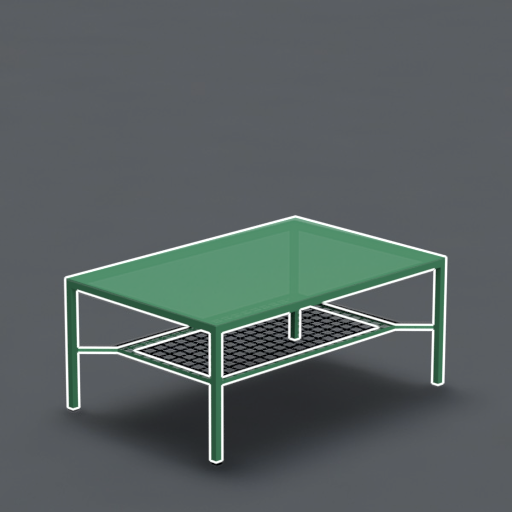}& \colImgN{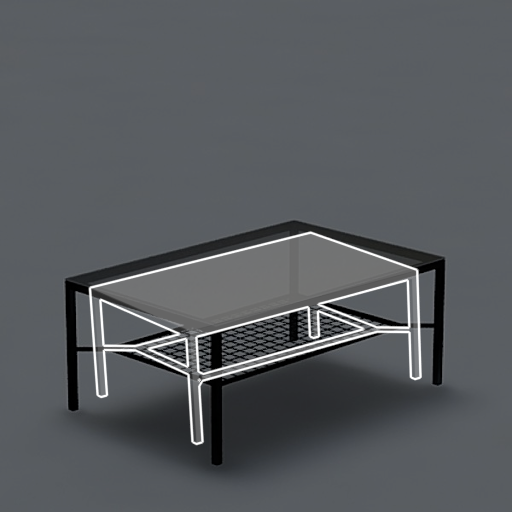}& \colImgN{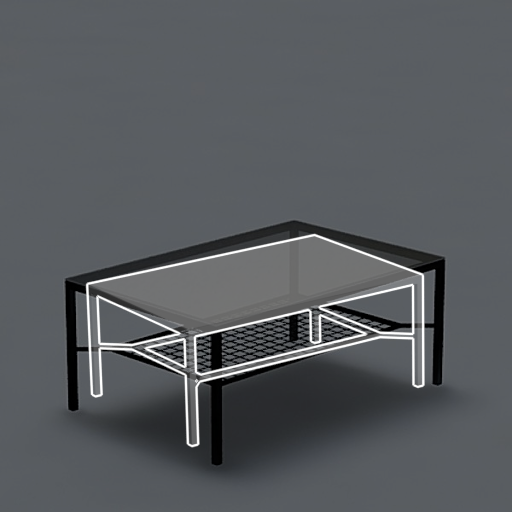}& \colImgN{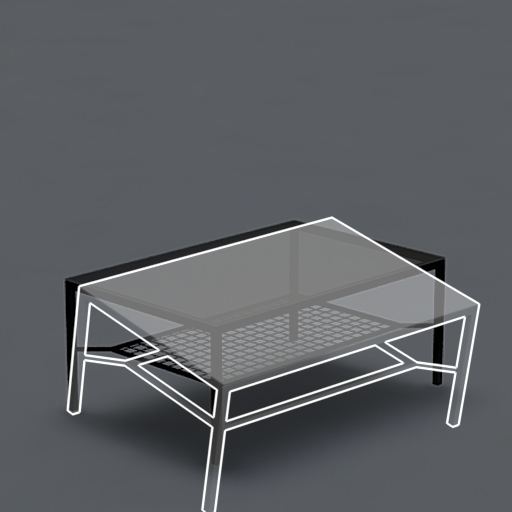}& \colImgN{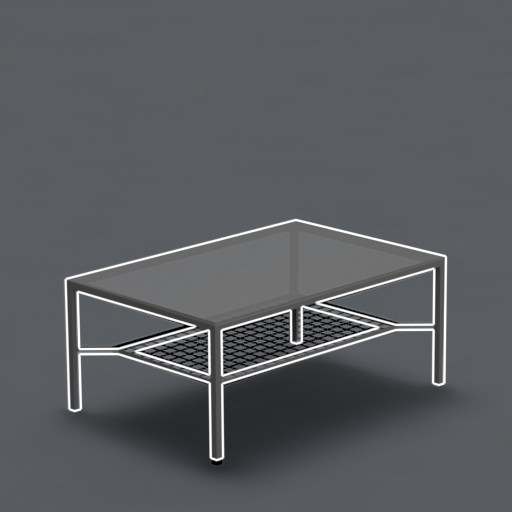}\\[-1pt]
		\footnotesize Image&\footnotesize GT&\footnotesize\cite{Grabner2019a}&\footnotesize\cite{Zakharov2019dpod}&\footnotesize\cite{Kato2018renderer}&\footnotesize Ours\\
	\end{tabular}
	\caption{Additional qualitative 3D pose refinement results for objects of different categories. We project the ground truth 3D model on the image using the predicted 3D pose. Our approach overcomes the limitations of previous methods and predicts fine-grained 3D poses which are in many cases visually indistinguishable from the ground truth. Best viewed in \textbf{digital zoom}.}
	\label{fig:collage4}
\end{figure*}

Figures~\ref{fig:collage2},~\ref{fig:collage3}, and~\ref{fig:collage4} show additional qualitative 3D pose refinement results for different methods complementary to those presented in the main paper. While other methods fail to predict fine-grained 3D poses in the wild, our approach precisely aligns 3D models to objects in RGB images which results in significantly improved 3D poses for objects of different categories. In many cases, our predicted 3D pose is visually indistinguishable from the ground truth 3D pose.

\begin{figure*}
	\setlength{\tabcolsep}{1pt}
	\setlength{\fboxsep}{-2pt}
	\setlength{\fboxrule}{2pt}
	\definecolor{boxgreen}{rgb}{0.3, 1.0, 0.3}
	\definecolor{boxred}{rgb}{1.0, 0.3, 0.3}
	\newcommand{\colImgN}[1]{{\includegraphics[width=0.2\linewidth]{#1}}}
	\newcommand{\colImgR}[1]{{\color{boxred}\fbox{\colImgN{#1}}}}
	\newcommand{\colImgG}[1]{{\color{boxgreen}\fbox{\colImgN{#1}}}}
	\centering
	\begin{tabular}{ccc}
		\colImgN{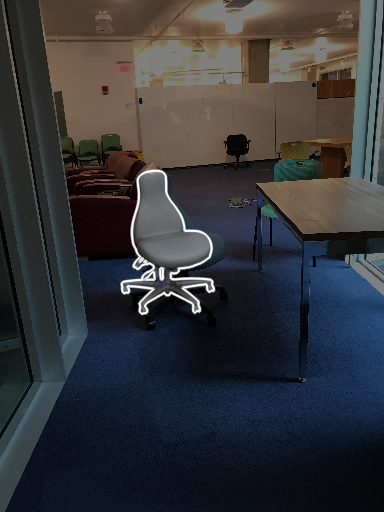}& \colImgN{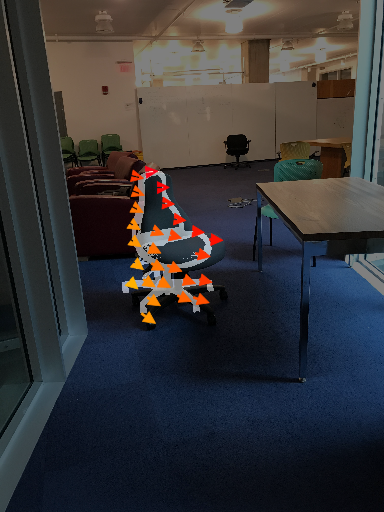}& \colImgN{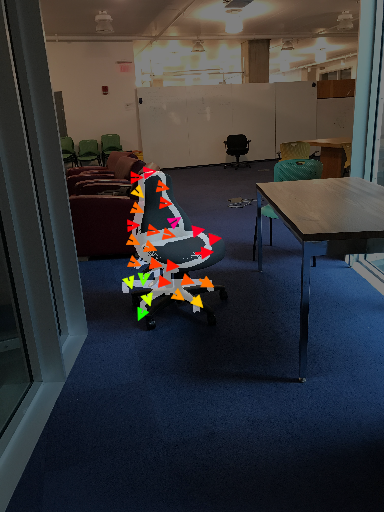}\\[-1pt]
		\colImgN{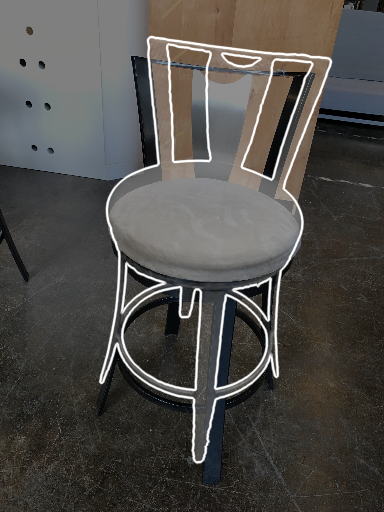}& \colImgN{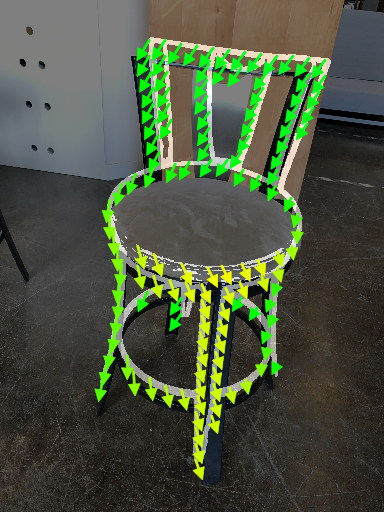}& \colImgN{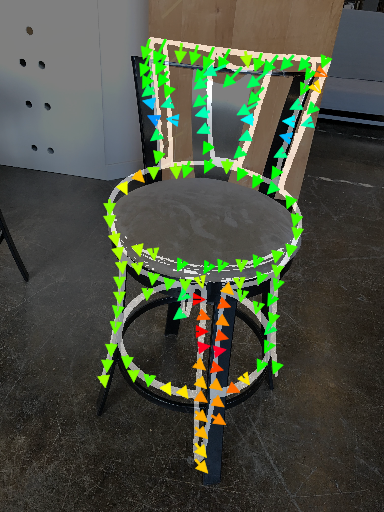}\\[-1pt]
		\colImgN{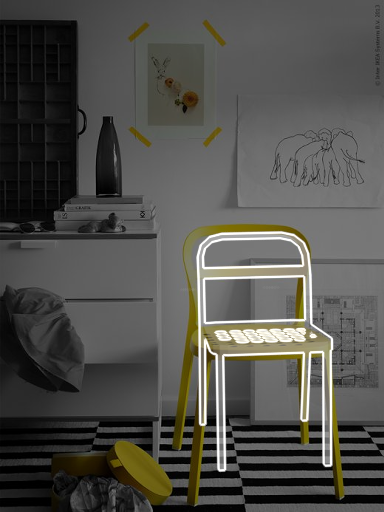}& \colImgN{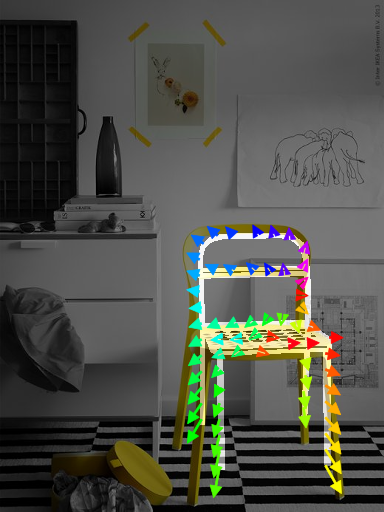}& \colImgN{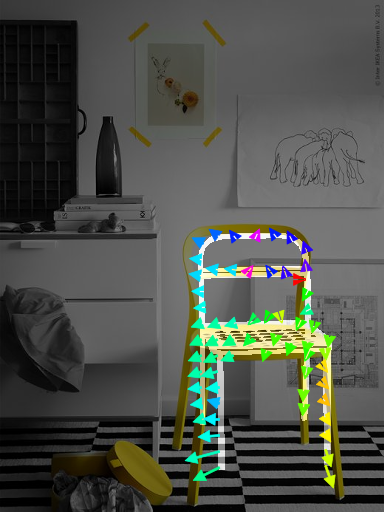}\\[-1pt]
		\colImgN{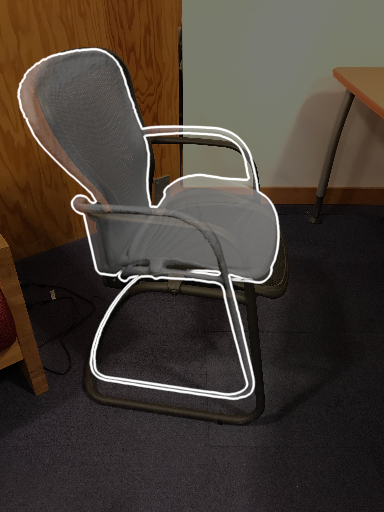}& \colImgN{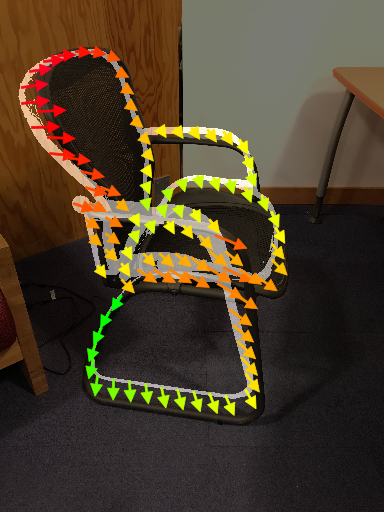}& \colImgN{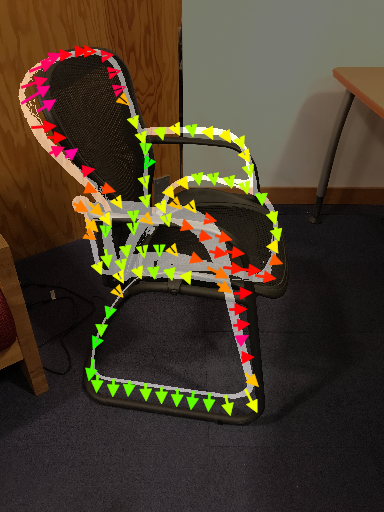}\\[-1pt]
		\colImgN{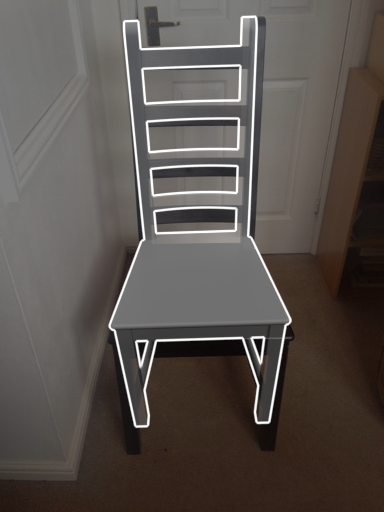}& \colImgN{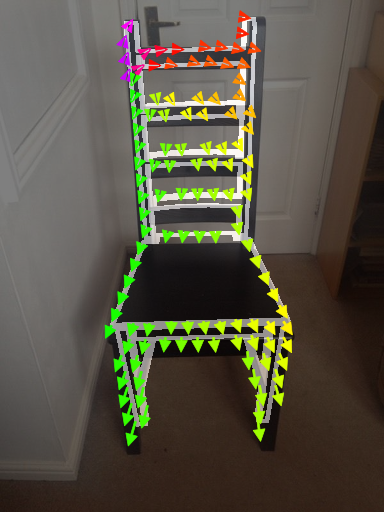}& \colImgN{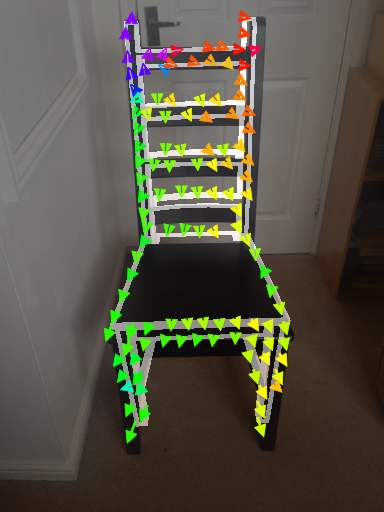}\\[-1pt]
		\footnotesize Initial 3D Pose&\footnotesize Ground Truth&\footnotesize Prediction\\
	\end{tabular}
	\caption{Additional qualitative examples of our predicted geometric correspondence fields. Our predicted 2D displacement vectors are highly accurate for many different objects and scales. Best viewed in \textbf{digital zoom}.}
	\label{fig:gcf2}
\end{figure*}

\begin{figure*}
	\setlength{\tabcolsep}{1pt}
	\setlength{\fboxsep}{-2pt}
	\setlength{\fboxrule}{2pt}
	\definecolor{boxgreen}{rgb}{0.3, 1.0, 0.3}
	\definecolor{boxred}{rgb}{1.0, 0.3, 0.3}
	\newcommand{\colImgN}[1]{{\includegraphics[width=0.2\linewidth]{#1}}}
	\newcommand{\colImgR}[1]{{\color{boxred}\fbox{\colImgN{#1}}}}
	\newcommand{\colImgG}[1]{{\color{boxgreen}\fbox{\colImgN{#1}}}}
	\centering
	\begin{tabular}{ccc}
		\colImgN{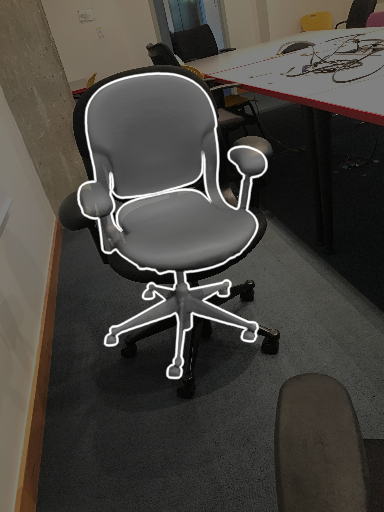}& \colImgN{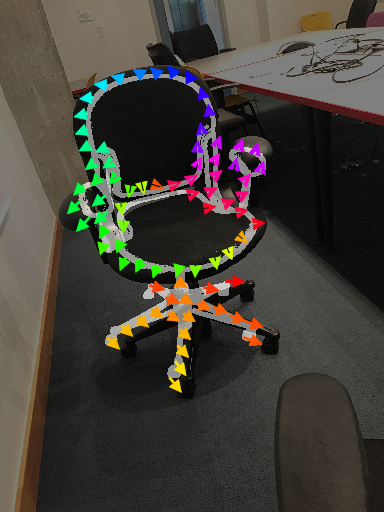}& \colImgN{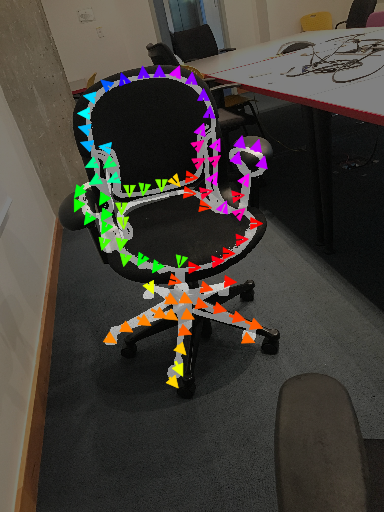}\\[-1pt]
		\colImgN{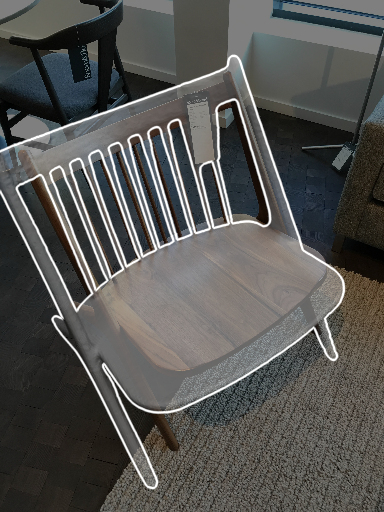}& \colImgN{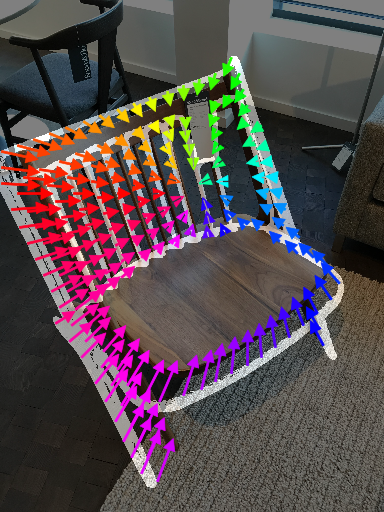}& \colImgN{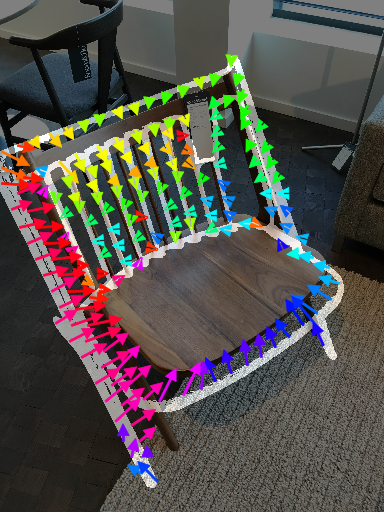}\\[-1pt]
		\colImgN{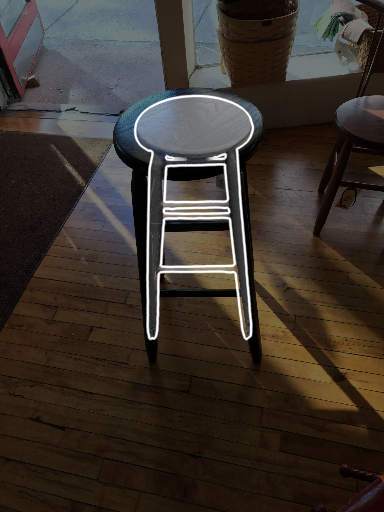}& \colImgN{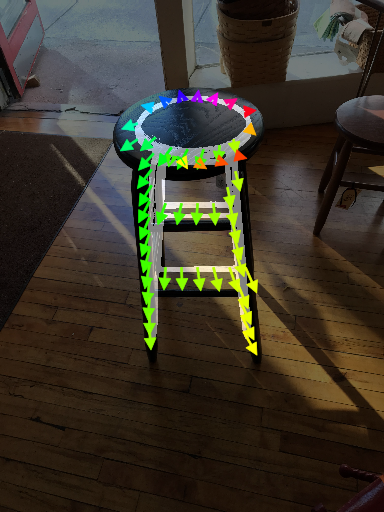}& \colImgN{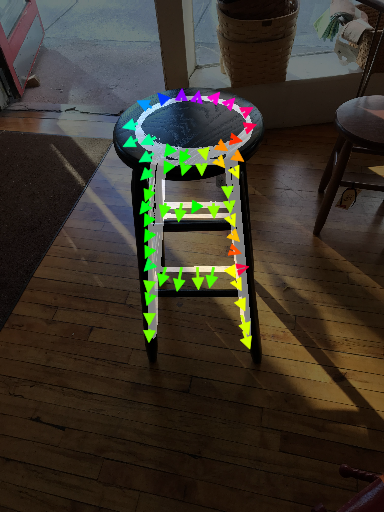}\\[-1pt]
		\colImgN{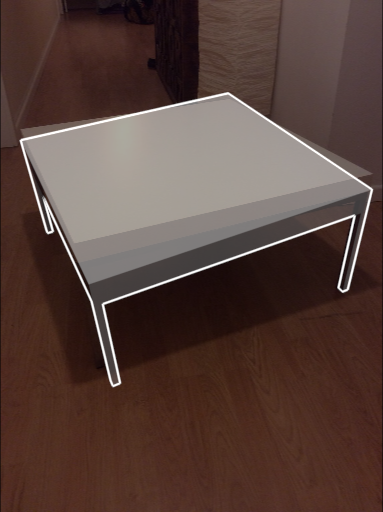}& \colImgN{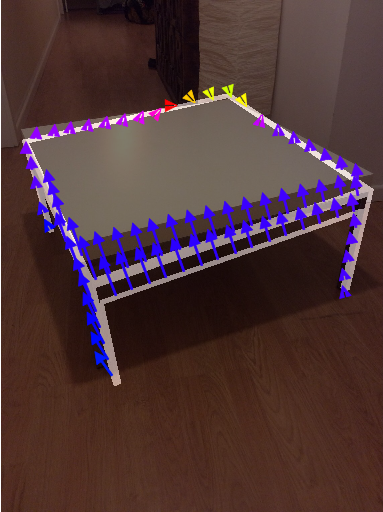}& \colImgN{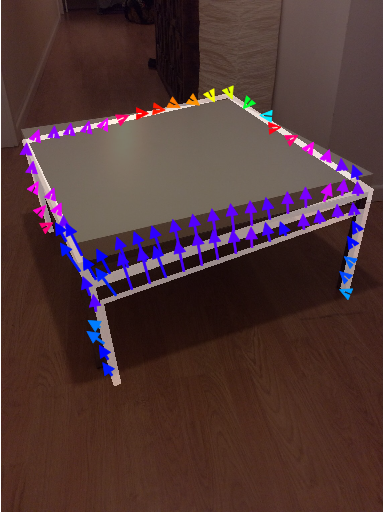}\\[-1pt]
		\colImgN{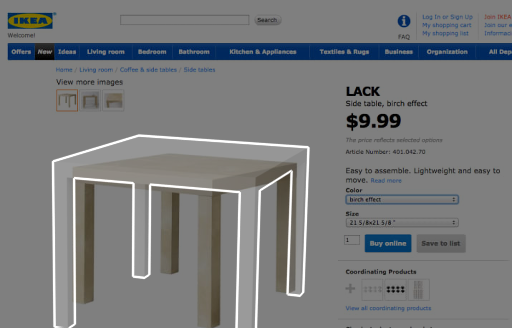}& \colImgN{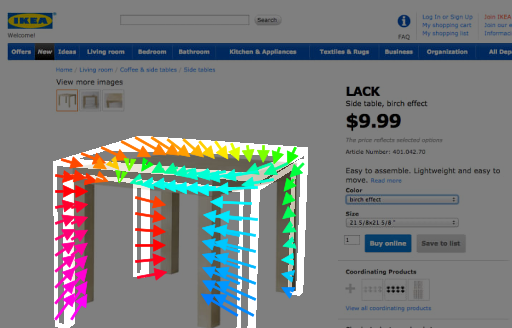}& \colImgN{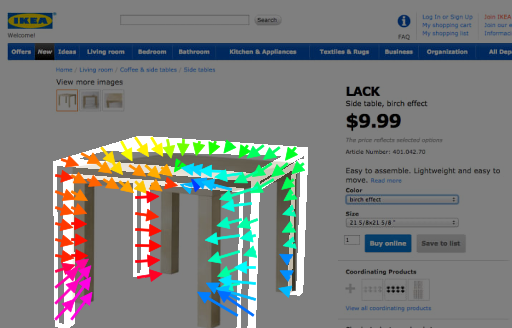}\\[-1pt]
		\colImgN{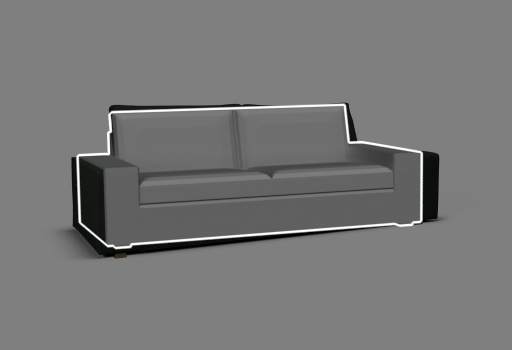}& \colImgN{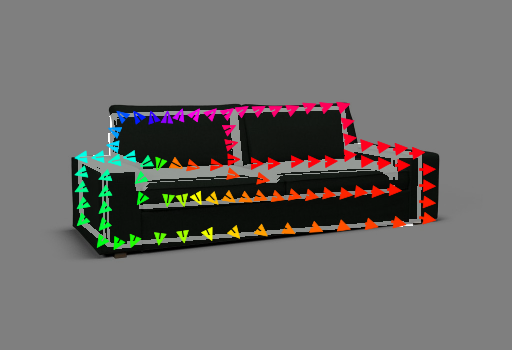}& \colImgN{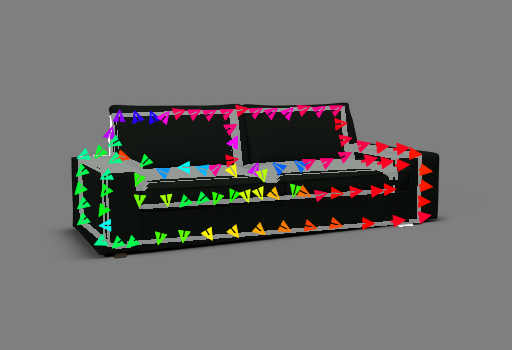}\\[-1pt]
		\footnotesize Initial 3D Pose&\footnotesize Ground Truth&\footnotesize Prediction\\
	\end{tabular}
	\caption{Additional qualitative examples of our predicted geometric correspondence fields. Our predicted 2D displacement vectors are highly accurate for many different objects and scales. Best viewed in \textbf{digital zoom}.}
	\label{fig:gcf3}
\end{figure*}

Figures~\ref{fig:gcf2} and~\ref{fig:gcf3} show additional qualitative results of our predicted geometric correspondence fields. Our predicted 2D displacement vectors are highly accurate for many different objects and scales in the wild. The illustrations also show our computed geometric attention masks, outlined in white underneath the 2D displacement vectors.

\section{Additional Failure Cases}
\label{sec:add-fail}

\begin{figure*}
	\setlength{\tabcolsep}{1pt}
	\setlength{\fboxsep}{-2pt}
	\setlength{\fboxrule}{2pt}
	\definecolor{boxgreen}{rgb}{0.3, 1.0, 0.3}
	\definecolor{boxred}{rgb}{1.0, 0.3, 0.3}
	\newcommand{\colImgN}[1]{{\includegraphics[width=0.24\linewidth]{#1}}}
	\newcommand{\colImgR}[1]{{\color{boxred}\fbox{\colImgN{#1}}}}
	\newcommand{\colImgG}[1]{{\color{boxgreen}\fbox{\colImgN{#1}}}}
	\centering
	\begin{tabular}{ccc}
		\colImgN{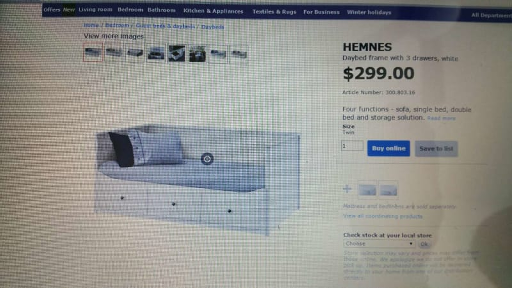}& \colImgN{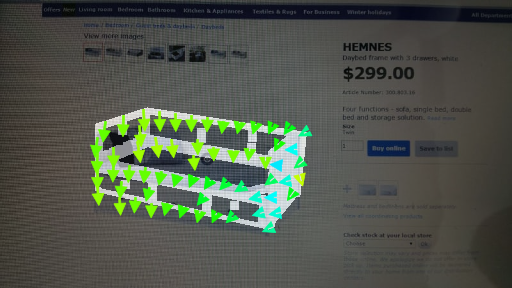}& \colImgN{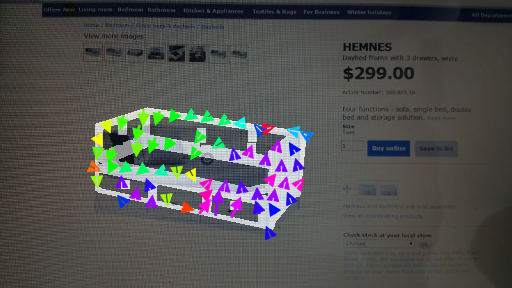}\\[-1pt]
		\footnotesize Image&\footnotesize GT GCF&\footnotesize Predicted GCF\\[3pt]
		\colImgN{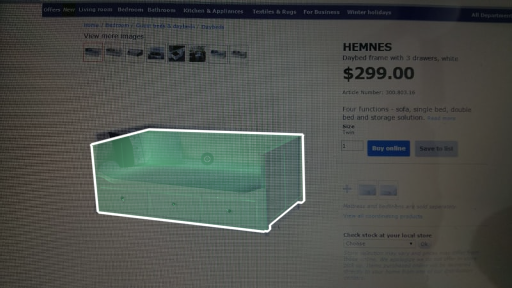}& \colImgN{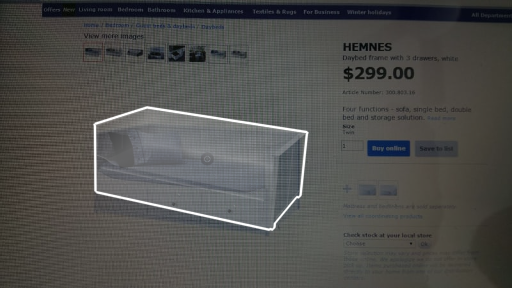}& \colImgN{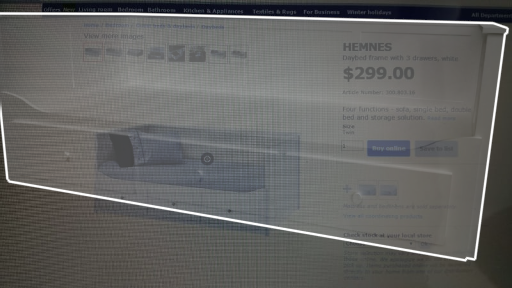}\\[-1pt]
		\footnotesize GT 3D Pose&\footnotesize Baseline~\cite{Grabner2019a} 3D Pose&\footnotesize Our 3D Pose\\[6pt]
		\midrule[1.5pt]
		\\[-6pt]
		\colImgN{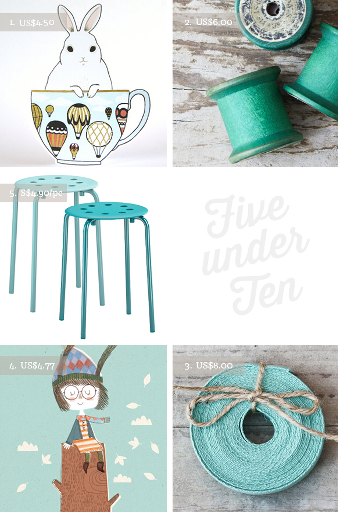}& \colImgN{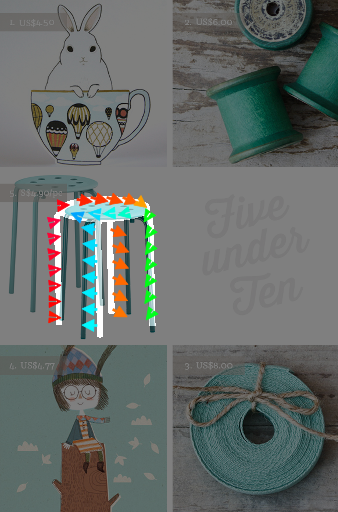}& \colImgN{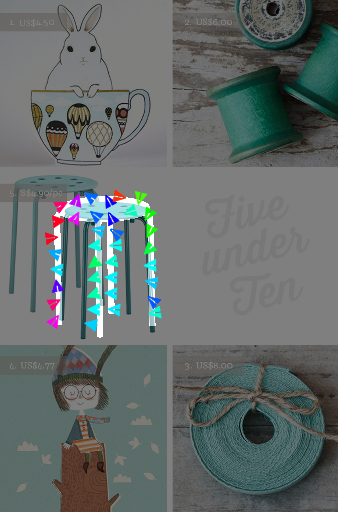}\\[-1pt]
		\footnotesize Image&\footnotesize GT GCF&\footnotesize Predicted GCF\\[3pt]
		\colImgN{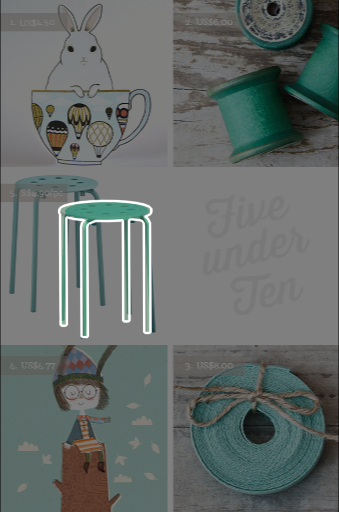}& \colImgN{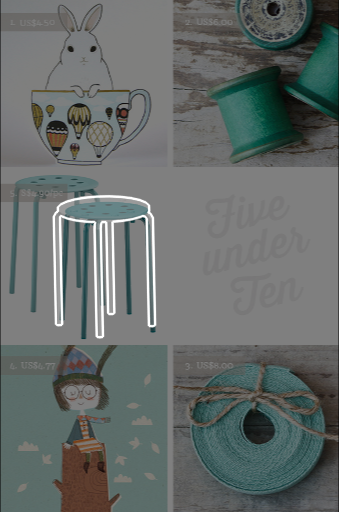}& \colImgN{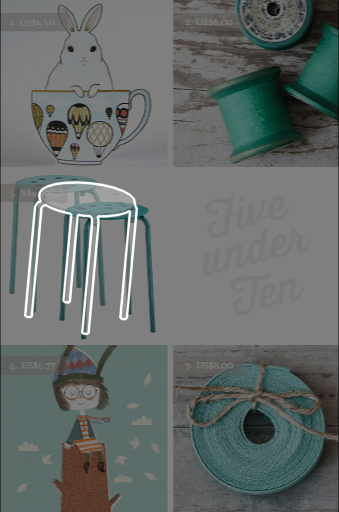}\\[-1pt]
		\footnotesize GT 3D Pose&\footnotesize Baseline~\cite{Grabner2019a} 3D Pose&\footnotesize Our 3D Pose\\[3pt]
	\end{tabular}
	\caption{Failure cases of our approach. In the presence of strong image noise (\textit{top example}), we cannot predict accurate geometric correspondence fields (GCF) and, thus, our refinement fails. Also, if there are duplicate or ambiguous structures in the image our method sometimes predicts wrong correspondences and aligns the 3D model to unintended image parts (\textit{bottom example}). Best viewed in \textbf{digital zoom}.}
	\label{fig:ours_fail}
\end{figure*}

Figure~\ref{fig:ours_fail} shows additional failure cases of our approach. In the presence of strong image noise, we cannot predict accurate geometric correspondence fields and, thus, our refinement fails. Also, if there are duplicate or ambiguous structures in the image our method sometimes predicts wrong correspondences and aligns the 3D model to unintended image parts.

\end{document}